\documentclass[10pt]{article}
\usepackage[preprint]{tmlr}
\usepackage{amsmath,amsfonts,bm}

\def\eqref#1{equation~\ref{#1}}

\def\1{\bm{1}}

\DeclareMathAlphabet{\mathsfit}{\encodingdefault}{\sfdefault}{m}{sl}
\SetMathAlphabet{\mathsfit}{bold}{\encodingdefault}{\sfdefault}{bx}{n}

\usepackage{hyperref}
\usepackage{caption}
\usepackage{placeins}  
\usepackage{url}
\usepackage{booktabs}
\usepackage{array}
\usepackage{makecell}
\usepackage{graphicx}
\usepackage{amsmath}
\usepackage{amssymb}
\usepackage{xcolor}
\usepackage{colortbl}       
\definecolor{rowhl}{HTML}{E6EEF8}

\definecolor{tdmblue}{HTML}{2A78D6}
\definecolor{shiftorange}{HTML}{EB6834}
\definecolor{inkgray}{HTML}{4A4A4A}
\definecolor{citeblue}{HTML}{1F5FAD}

\hypersetup{colorlinks=true, citecolor=citeblue, linkcolor=citeblue, urlcolor=citeblue}
\newcommand{\doi}[1]{doi: \href{https://doi.org/#1}{\nolinkurl{#1}}}

\title{Typed Decision Models under Prevalence Shift and Label Budgets: Decision Training, Calibration, and Implicit Priors}

\author{\name Abdul Rahman Iqbal \email abdul.rahmanmezzi@gmail.com \\
      \addr Faculty of Computer Science and Information Technology\\
      The Superior University, Lahore, Pakistan
      \AND
      \name Muhammad Fawad Nasim \email fawad.nasim@superior.edu.pk \\
      \addr Faculty of Computer Science and Information Technology\\
      The Superior University, Lahore, Pakistan
      \AND
      \name Amir Saleem \email mramir401@gmail.com \\
      \addr Independent Researcher}

\def\month{MM}
\def\year{YYYY}
\def\openreview{\url{https://openreview.net/forum?id=XXXX}}

\begin{document}

\maketitle

\begin{abstract}
Typed decision models (TDMs) return probabilities over caller-defined options, and Jev's vendor describes it as
calibrated. We ask whether vendor decision training improves a model over its backbone, whether a TDM's calibration
and thresholds withstand prevalence (label) shift like a fine-tuned classifier's, and how many labels a trained model
needs to overtake a TDM. We compare three TDMs (Jev, Tev1-4B, Nimble-9B), their base models and trained classifiers on
identical candidate-set decisions from five public datasets (on intent datasets, within the gold label's intent group)
under simulated shift, with 35 paired tests frozen before the confirmatory analysis (earlier test-data checks
disclosed), and estimate label-budget curves. First, decision training is not always beneficial: Nimble-9B, which
differs from Qwen3.5-9B only by a LoRA adapter, is better calibrated on intent tasks but 12 Macro-F1 points worse on
financial sentiment, a loss that persists across prompt formats; Tev1-4B's largest gain over its backbone is a
prompt-format effect. Second, on 50 fixed moderate shifts, Jev's top-label calibration error and threshold risk grow
significantly more than a fine-tuned DeBERTa's only on financial sentiment, sensitive to the prior set; averaged over
2{,}000 shifts the gap is at most one point. Third, label-free EM correction with the model's implicit prior lowers
Jev's shifted calibration error where it is highest, but labelled recalibration lowers it far more. Fourth, DeBERTa
needs 100--250 labels to overtake Jev on hate speech and does not clearly overtake it on CLINC150 at any budget; a
QLoRA adapter on Qwen3.5-4B overtakes Jev with at most 2{,}500 labels on every dataset, on CLINC150 only marginally
and not under in-budget selection. The TDMs tested are neither uniformly better than their backbones nor uniquely
fragile under shift; their lead over trained models depends on the task.
\end{abstract}

\section{Introduction}
\label{sec:intro}

Consider a bank that routes incoming customer messages to one of a fixed set of intents and sends a message to a
human whenever the routing model is less than 90\% confident. The threshold was chosen on validation data in which,
say, one message in ten was a complaint. After a service outage, four messages in ten are complaints. The model,
the messages of each kind and the threshold are unchanged; only the class frequencies have moved. We study this case
by resampling labelled test sets, which isolates the change in class frequencies; real deployments may change more.
Does the model's
confidence still mean what it meant at validation time, and does the threshold still send the intended share of
messages to a human?

This question has become practical for a new class of models. Typed decision models (TDMs), also called
``System One'' models, do not generate text: given an input and a caller-defined set of options, they return a
probability distribution over the options \citep{deusser2026jev,tang2026audit}. Jev is offered as a hosted API and
is described by its vendor as calibrated; Tev1-4B and Nimble-9B are open models fine-tuned from Qwen3.5 backbones
for the same interface. A trained classifier has a training prior that label-shift corrections can use
\citep{saerens2002adjusting,lipton2018detecting}. A zero-shot model has none. The closest substitute is an
\emph{implicit prior}: the average distribution the model predicts on a development set, which requires no labels.

Existing evaluations leave two gaps. First, the benefit of vendor ``decision training'' has not been separated from the
backbone: an audit of early TDM papers calls for same-backbone ablations \citep{tang2026audit}, existing comparisons of
a vendor TDM with its backbone read the two models out differently \citep{rafe2026systemone}, and the study that does
compare identical weights fine-tunes its own models and finds gains only on some tasks \citep{li2026llmasjev}. Second, TDMs have been compared with language models and trained classifiers at a
single class distribution \citep{deusser2026jev,rafe2026systemone}, so it is not known whether their calibration and
thresholds transfer when the class frequencies change; a recent audit of 28 TDM papers lists calibration under
deployment shift as an open problem \citep{tang2026audit}. Prior adaptation without labels has been studied for language
models and vision--language models \citep{estienne2023unsupervised,han2026labelshift}, but not for TDMs against trained
classifiers under controlled shift. We are not aware of a study that combines a commercial TDM, open TDMs paired with
their own backbones, trained classifiers and controlled prevalence shift with label-free prior correction
(\hyperref[tab:coverage]{Table~\ref*{tab:coverage}}); this combination is what allows us to ask what decision training adds and whether it
survives a change in class frequencies. A third question is practical: a TDM needs no labelled data, but it is not
known how many labels a trained classifier, or a task-tuned version of the same open backbone, needs to do better
\citep{lescao2021howmany,edwards2024language}.

We study four research questions on five public, human-labelled datasets, using identical test items and
candidate sets for every system:
\textbf{RQ1} (decision training): does vendor decision training change accuracy and calibration relative to the same
backbone?
\textbf{RQ2} (prevalence shift): under controlled prevalence shift, do a TDM's calibration error and threshold risk
change differently from those of a fine-tuned classifier?
\textbf{RQ3} (implicit-prior correction): does expectation--maximisation (EM) prior correction with the implicit
prior reduce a TDM's calibration error under shift?
\textbf{RQ4} (label budgets): how many training labels does a fine-tuned classifier, or a task QLoRA on a TDM's
backbone, need before it is more accurate than Jev?
Our results concern \emph{candidate-set} decisions, in which the gold label is always among 3--22 offered options, and
\emph{simulated} label shift, in which only the class frequencies change; they do not cover full-label classification
over all 150 or 64 intents, or deployment shifts that also change the text of each class (\hyperref[sec:limitations]{Section~\ref*{sec:limitations}}).
The tests and the comparison family were fixed and hashed before the final runs, and the analysis code before any
confirmatory analysis was run (\hyperref[sec:protocol]{Section~\ref*{sec:protocol}}). This does not mean that no test
data were used before: pipeline checks before the freeze used test splits, mainly FPB's, including early versions of
the shift tests on FPB, and the one amendment followed some final test predictions; both are listed in
\hyperref[app:deviations]{Appendix~\ref*{app:deviations}}. The label-budget study (RQ4) was frozen in the same way
before any of its final test predictions; it estimates learning curves and adds no test to the family.

Our central finding is that the advantages of typed decision models are conditional: on the task, on the prompt
format used for the comparison, on the class distribution at deployment, and on how many labels are available. A
TDM should therefore be tested against its own base model, under prevalence shift, and against a cheaply tuned model
before it is deployed.

\begin{figure}[t]
  \centering
  \includegraphics[width=\linewidth]{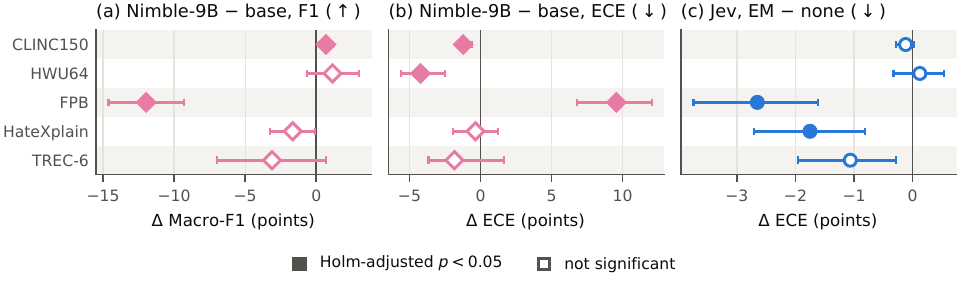}
  \caption{\textbf{Decision training changes a backbone in a task-dependent direction, and label-free prior correction
  helps where calibration is poor.}
  (a, b) Paired differences between Nimble-9B and its own base model Qwen3.5-9B (same weights and 8-bit precision,
  differing only by a LoRA adapter) in Macro-F1 (higher favours Nimble) and expected calibration error (ECE; lower
  favours Nimble), in percentage points.
  (c) Change in Jev's ECE under prevalence shift ($\alpha=1$, 50 fixed target priors) when EM prior correction with
  its implicit prior is applied (lower is better).
  Bars are 95\% paired bootstrap intervals ($B=10{,}000$, unadjusted); filled markers are significant after Holm
  correction over all 35 primary tests.}
  \label{fig:teaser}
\end{figure}

Our findings are as follows (\hyperref[fig:teaser]{Figure~\ref*{fig:teaser}}; all 35 tests in \hyperref[tab:alltests]{Table~\ref*{tab:alltests}}).
\begin{itemize}
  \item \textbf{Decision training is task-dependent.} Nimble-9B, which differs from Qwen3.5-9B only by a LoRA
  adapter, is better calibrated on both intent datasets but 12.0 Macro-F1~points worse on financial sentiment, where
  its calibration error is also higher (\hyperref[sec:rq1]{Section~\ref*{sec:rq1}}); this holds in two generic prompt
  formats and in bf16 (exploratory). In the Tev1 format, Tev1-4B is never significantly worse than Qwen3.5-4B and is better on five of
  ten tests; these gains may depend on that format. Its largest gain, 10.0 Macro-F1~points on FPB, is a prompt-format effect: in generic formats Qwen3.5-4B
  scores 9.6~points higher and the gain disappears (exploratory). Its HWU64 gain is not resolved without the test
  texts that also occur in MASSIVE (exploratory).
  \item \textbf{Moderate prevalence shift separates Jev from a fine-tuned classifier only a little.} On the
  pre-specified set of 50 target priors at a Dirichlet concentration of $\alpha=1$, Jev's ECE and the excess risk of
  its confidence threshold grow significantly more than DeBERTa-v3-large's (after Holm correction) only on FPB, by 1.7
  and 2.5 percentage points, a result conditional on that prior set and on the dataset whose test split was exposed in
  pipeline checks; averaged over
  2{,}000 shift scenarios (exploratory), the difference is at most one point on every dataset and largest on TREC-6 (\hyperref[sec:rq2]{Section~\ref*{sec:rq2}}).
  \item \textbf{Implicit-prior correction helps where calibration is poor.} EM with Jev's implicit prior, which needs
  no labels, lowers Jev's ECE under shift by 2.7~points on FPB and 1.8~points on HateXplain, the two datasets where its
  unshifted ECE is highest (0.095 and 0.279), with the same sign for every prior set; on the other three datasets the
  effect is not significant after Holm correction. Recalibration on labelled development data lowers the error much
  further (\hyperref[sec:rq3]{Section~\ref*{sec:rq3}}).
  \item \textbf{The labels needed to overtake Jev depend on the task.} A fine-tuned
  DeBERTa-v3-large overtakes Jev with 100--250 labels on hate speech, but needs 2{,}500--5{,}000 on HWU64 (also with
  model selection inside the budget) and does not clearly overtake it on CLINC150 even with the full training set. A
  QLoRA adapter on Qwen3.5-4B overtakes Jev with at most 100 labels on two datasets and with at most 2{,}500 on every
  dataset, on CLINC150 only marginally and not under in-budget selection; with 500 labels it is above both its frozen base and Tev1-4B on every dataset
  (fixed schedule; \hyperref[sec:rq4]{Section~\ref*{sec:rq4}}).
  \item \textbf{Artifacts.} We release the frozen label texts, candidate sets, prompts, evaluation code and every
  per-example probability vector at \url{https://github.com/AbdulRahmanIqbal/tdm-prevalence-shift}.
\end{itemize}
In short, decision training helps a backbone on some tasks and hurts it on others, prevalence shift matters less
for the TDMs we tested than the calibration error they already have without shift, and the number of labels at which a
TDM stops being the better choice ranges from about a hundred to more than the full training set.

\section{Related work}
\label{sec:related}

\begin{table}[t]
  \centering
  \caption{\textbf{Our contribution is the combination of these design elements; each has been studied before.}
  Coverage of vendor TDMs, matched backbones, trained classifiers, controlled prevalence shift and label-free prior
  correction in related studies. \checkmark: covered; $\circ$: partly covered (Pair: different readouts for the
  two models, or the authors' own fine-tunes rather than a vendor TDM; Tests: intervals or tests without multiplicity
  control); --: not covered.
  Columns: TDM = commercial or vendor TDMs evaluated; Pair = TDM compared with its own backbone under the same readout;
  Trained = task-trained classifier baseline; Shift = controlled class-prior shift; Prior = label-free prior correction
  for zero-shot models; Tests = paired tests with multiplicity control.}
  \label{tab:coverage}
  \scriptsize
  \begin{tabular}{lcccccc}
    \toprule
    Study & TDM & Pair & Trained & Shift & Prior & Tests \\
    \midrule
    \citet{deusser2026jev} & \checkmark & -- & -- & -- & -- & $\circ$ \\
    \citet{rafe2026systemone} & \checkmark & $\circ$ & \checkmark & -- & -- & \checkmark \\
    \citet{li2026llmasjev} & \checkmark & $\circ$ & -- & -- & -- & $\circ$ \\
    \citet{css2026decision} & \checkmark & -- & -- & -- & -- & -- \\
    \citet{estienne2023unsupervised} & -- & -- & -- & -- & \checkmark & -- \\
    \citet{lipton2018detecting,alexandari2020maximum,garg2020unified} & -- & -- & \checkmark & \checkmark & -- & $\circ$ \\
    \midrule
    \rowcolor{rowhl} Our study & \checkmark & \checkmark & \checkmark & \checkmark & \checkmark & \checkmark \\
    \bottomrule
  \end{tabular}
\end{table}

\hyperref[tab:coverage]{Table~\ref*{tab:coverage}} summarises how our study relates to the closest prior work. Individually, the ingredients are
known; the question that is new is how a zero-shot decision model's calibration, thresholds and label-free prior
correction behave under prevalence shift compared with a trained classifier, and what decision training changes relative
to the identical backbone.

\paragraph{Evaluations of typed decision models.}
Jev was released in September 2026, and evaluations followed within weeks. \citet{deusser2026jev} evaluate Jev
zero-shot on 37 public datasets against two open models and report well-calibrated choice probabilities overall, with
miscalibration, mostly overconfidence, concentrated on the datasets where accuracy is low.
\citet{rafe2026systemone} compare several TDMs with trained classifiers and generative models under matched conditions
and find that the ranking depends on whether task labels exist; they also compare each decoder-based TDM, including Nimble,
with its Qwen3.5 backbone, but read the backbone out differently (by option-key likelihood), on their typed-decision
benchmark, a sample of CLINC150 and several social-science tasks. \citet{li2026llmasjev} argue that general-purpose
language models already behave like Jev-style decision models when read out by scoring the candidate labels, and that
fine-tuning helps only on some tasks. Other early studies cover annotation tasks in computational social science
\citep{css2026decision}, optimised natural-looking context that flips decisions \citep{jevout2026}, and
confidence-based escalation of LLM judges
\citep{jevjudge2026}. An audit of 28 early TDM papers finds no independent accuracy advantage of the typed readout over
comparable label-probability readouts, and lists calibration under deployment shift and the attribution of gains as
open problems \citep{tang2026audit}. Complementary to these evaluations, \citet{li2026jeva} take an architectural
view: their decision-native model encodes the request and each option separately, which makes typed decisions
null-aware, invariant to the order of the options, scalable to large candidate sets and reusable across requests. We
evaluate existing TDMs instead, and control option order by fixing and counterbalancing it
(\hyperref[sec:protocol]{Section~\ref*{sec:protocol}}). All of these studies evaluate at the class distribution of the
test set. Our
study varies the class distribution in a controlled way, and pairs each open TDM with its base model under the same
prompt, readout and precision, on five datasets; for Nimble-9B, which differs from its base only by an adapter, this
isolates the effect of decision training.

\paragraph{Label shift and prior correction.}
Under label (prior) shift, $p(y)$ changes while $p(x\mid y)$ stays fixed. \citet{saerens2002adjusting} re-estimate the
target prior with EM, using the classifier's posteriors and the training prior. Black-box shift estimation
\citep{lipton2018detecting} and its regularised variant \citep{azizzadenesheli2019regularized} estimate the target
prior from a confusion matrix. \citet{alexandari2020maximum} show that EM is hard to beat once the classifier is
calibrated with class-specific biases (BCTS), and \citet{garg2020unified} unify these estimators and attribute the
advantage of maximum likelihood to calibration. All of these methods assume a known source prior. For zero-shot
models, which have no training prior, prior adaptation without labels has been proposed for language models
\citep{estienne2023unsupervised}, and zero-shot CLIP models have been adapted to a given label-distribution prior
\citep{kahana2022improving} or to an unknown target distribution estimated from unlabelled test data
\citep{han2026labelshift}. We use the simplest
version of this idea, the mean predicted distribution on a development set as an implicit source prior, and test it
on a commercial TDM against a trained classifier whose source prior is known.

\paragraph{Calibration of pretrained and prompted models.}
Modern networks are often miscalibrated, and temperature scaling on held-out data is a strong fix
\citep{guo2017calibration}. Fine-tuned transformers are relatively well calibrated in-domain \citep{desai2020calibration},
whereas prompted language models carry label biases that a content-free calibration step can reduce
\citep{zhao2021calibrate}. We measure calibration with top-label ECE and selective risk \citep{geifman2017selective}, and
ask how both change when the class frequencies move.

\paragraph{Zero-shot models versus trained classifiers.}
With enough labelled data, fine-tuned encoders tend to outperform in-context or zero-shot use of larger language models
on text classification \citep{edwards2024language,bucher2024finetuned}. Fine-tuning on few examples is, however, unstable \citep{mosbach2021stability,zhang2021revisiting}, which our
training protocol takes into account. \citet{lescao2021howmany} measure the value of a prompt in labelled examples by
comparing learning curves over several runs; we use the same idea to ask how many labels a trained model needs to
overtake a TDM. Results from few-shot model selection are optimistic when a large labelled development set is used
\citep{perez2021true}, which applies to our small budgets as well. For intent detection, work on CLINC150 and HWU64
reports accuracy at 5 to 30 examples per intent \citep{casanueva2020efficient,zhang2021fewshot}; we translate our budgets
into examples per class for comparison. Parameter-efficient fine-tuning with low-rank adapters, also on quantised
models, makes task-specific tuning of the same open backbones cheap \citep{hu2022lora,dettmers2023qlora}; we use it to
compare task fine-tuning with vendor decision training under an identical readout. Our curves end with classifiers
trained on the full training sets, the setting in which a trained classifier is strongest.

\section{Setting}
\label{sec:setting}

\begin{figure}[t]
  \centering
  \includegraphics[width=\linewidth]{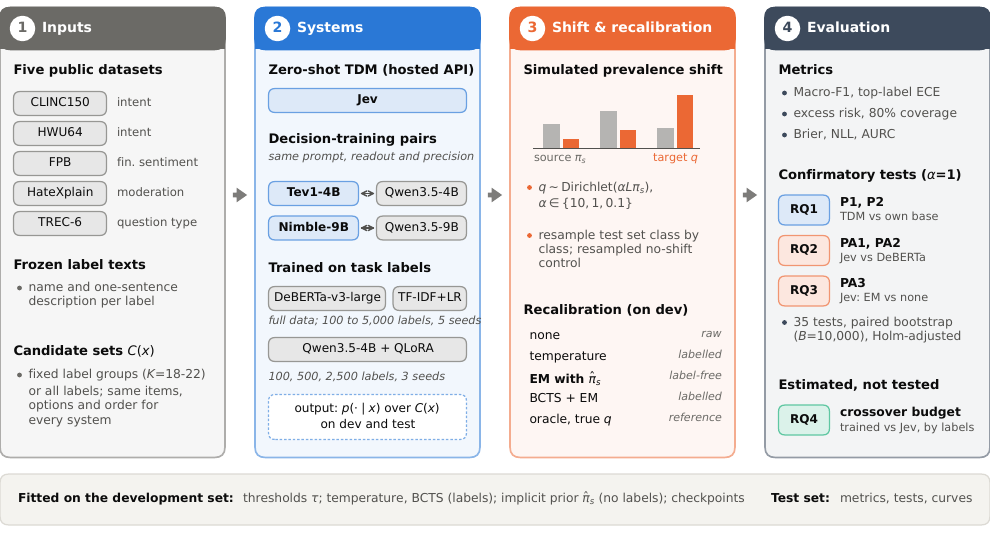}
  \caption{\textbf{All systems answer identical candidate-set decisions, and shift, recalibration and tests operate on
  their saved probabilities, so every comparison is paired.}
  (1) Five public datasets with frozen label texts and fixed candidate sets. (2) A hosted TDM, two open TDMs each paired
  with its own backbone, and models trained on task labels (two classifiers on the full data and on label budgets, and a
  QLoRA adapter on Qwen3.5-4B) return a probability vector over the candidates. (3) The test set
  is resampled to simulated target priors (\hyperref[eq:shift]{Eq.~\ref*{eq:shift}}), and five recalibration arms are fitted on the unshifted
  development set (\hyperref[eq:implicit]{Eqs.~\ref*{eq:implicit}}--\nobreak\ref{eq:em}). (4) Metrics (\hyperref[eq:ece]{Eqs.~\ref*{eq:ece}} and~\ref{eq:risk}), the 35
  confirmatory tests (\hyperref[eq:pa1]{Eqs.~\ref*{eq:pa1}}--\nobreak\ref{eq:pa3}) and the label-budget curves, which are
  estimated, not tested; everything else is descriptive.}
  \label{fig:pipeline}
\end{figure}

\hyperref[fig:pipeline]{Figure~\ref*{fig:pipeline}} summarises the evaluation. Every system answers the same candidate-set decisions; prevalence
shift and recalibration are then applied to the saved probability vectors, so that all comparisons are paired.

\paragraph{Decision models and candidate sets.}
Let $\mathcal{Y}$ be the label set of a dataset and $L=|\mathcal{Y}|$ its number of classes. Each test example $x$ comes
with a candidate set $C(x)\subseteq\mathcal{Y}$ of $K$ labels that always contains the gold label. A system returns a
probability vector $p(\cdot\mid x)$ over $C(x)$. The prediction is $\hat y=\arg\max_{c\in C(x)} p(c\mid x)$, with ties
broken by candidate position, and the confidence is $\max_{c\in C(x)} p(c\mid x)$. All systems receive the same candidate
set, in the same order, for every example (\hyperref[sec:protocol]{Section~\ref*{sec:protocol}}).

\paragraph{Prevalence shift.}
Let $\pi_s$ be the class distribution of the source data and $q$ that of the deployment data. Under prevalence (label)
shift, $q\neq\pi_s$ while $p(x\mid y)$ is unchanged. We simulate it with a source-centred variant of the Dirichlet shift
of \citet{lipton2018detecting} and \citet{alexandari2020maximum}, who use a symmetric concentration, by drawing a target
prior and class counts for a sample of the test size $n$,
\begin{equation}
  q\sim\mathrm{Dirichlet}(\alpha L\,\pi_s),\qquad (n_y)_{y\in\mathcal{Y}}\mid q\sim\mathrm{Multinomial}(n,q),
  \label{eq:shift}
\end{equation}
where $\pi_s$ is the class distribution of the training split (the same for all systems), and then drawing $n_y$ test
examples of class $y$ with replacement; we call the result the shifted sample $S_q$. The concentration $\alpha L\pi_s$
centres the draws on the source prior and reduces to the usual $\mathrm{Dirichlet}(\alpha\mathbf{1})$ when $\pi_s$ is
uniform. Smaller $\alpha$ means stronger shift; we use $\alpha\in\{10,1,0.1\}$, and the confirmatory tests use $\alpha=1$.

\paragraph{Source prior and implicit prior.}
Prior correction needs the source prior $\pi_s$. For a trained classifier it is the class distribution of its training
data. A zero-shot model has no training data for the task, so we use its \emph{implicit prior}, the smoothed mean
predicted distribution over the $n_{\mathrm{dev}}$ development examples,
\begin{equation}
  \hat\pi_s(y)=\frac{\frac{1}{n_{\mathrm{dev}}}\sum_{i=1}^{n_{\mathrm{dev}}}p(y\mid x_i)+\epsilon}{1+L\epsilon},
  \qquad \epsilon=10^{-3},
  \label{eq:implicit}
\end{equation}
where $p(y\mid x_i)=0$ for $y\notin C(x_i)$; the smoothing keeps classes that the model never predicts from receiving a
prior of zero. This estimate needs no labels. It is motivated by the law of total probability: if the model's posteriors
were calibrated on the source distribution, their mean would equal the source prior, $\mathbb{E}_{x}[p(y\mid
x)]=\pi_s(y)$, and EM started from $\hat\pi_s$ would leave an unshifted sample unchanged. The implicit prior is thus the
source prior with which the model's own posteriors are consistent, a proxy in the spirit of black-box shift estimation
\citep{lipton2018detecting}. The converse does not hold. Agreement of the mean prediction with $\hat\pi_s$ is a property
of the marginal only; it does not make $p(y\mid x)$ the source posterior, which is what the likelihood argument for EM
requires \citep{saerens2002adjusting}. For a miscalibrated model, EM with $\hat\pi_s$ is therefore not a consistent
estimator of the target prior, and we treat it as an empirical, label-free correction whose effect is tested (PA3)
rather than guaranteed. A simulation with deliberately distorted posteriors (exploratory;
\hyperref[tab:emmiscal]{Table~\ref*{tab:emmiscal}}) shows the estimate degrading as calibration worsens, most for
underconfident posteriors, while staying far closer to the target prior than no correction, and with distorted
posteriors the implicit prior serves EM at least as well as the true source prior. EM also only rescales probabilities: a candidate to which the model assigns
probability zero keeps it after correction, so EM cannot repair the examples on which Jev gives the gold label
probability zero (below). Empirically, it is close to the true class distribution on the intent datasets and far from it
on FPB and HateXplain, where the models have a label bias: its total-variation distance to the development set's class distribution is
at most 0.033 for every zero-shot system on CLINC150, but 0.17--0.38 on FPB and 0.23--0.27 on HateXplain
(\hyperref[tab:implicitprior]{Table~\ref*{tab:implicitprior}}; \hyperref[fig:implicitprior]{Figure~\ref*{fig:implicitprior}}). We do not evaluate BBSE or RLLS themselves, because they need a confusion
matrix estimated on labelled data, whereas our question concerns correction without labels; labelled recalibration is
evaluated separately (BCTS+EM).

\paragraph{Recalibration arms.}
For every system we compare five ways of producing probabilities on the shifted test set; the four deployable ones are
fitted on the unshifted development set without target labels, and the oracle uses the true target prior:
\textbf{none} (raw probabilities);
\textbf{temperature} scaling \citep{guo2017calibration};
\textbf{EM} prior correction \citep{saerens2002adjusting};
\textbf{BCTS+EM}, bias-corrected temperature scaling followed by EM \citep{alexandari2020maximum}; and
\textbf{oracle}, prior correction with the true target prior $q$ (an upper reference, not a deployable method).
EM starts from $\hat q^{(0)}=\hat\pi_s$ and alternates, for the $n$ examples of the shifted sample and $y\in C(x)$,
\begin{equation}
  \tilde p^{(t)}(y\mid x)=\frac{p(y\mid x)\,\hat q^{(t)}(y)/\hat\pi_s(y)}
  {\sum_{y'\in C(x)}p(y'\mid x)\,\hat q^{(t)}(y')/\hat\pi_s(y')},\qquad
  \hat q^{(t+1)}(y)=\frac{1}{n}\sum_{i=1}^{n}\tilde p^{(t)}(y\mid x_i),
  \label{eq:em}
\end{equation}
until $\max_y|\hat q^{(t+1)}(y)-\hat q^{(t)}(y)|<10^{-6}$ (at most 100 iterations). The oracle arm applies the E-step
once with $q$ in place of $\hat q^{(t)}$.

\paragraph{EM on candidate sets.}
For CLINC150 and HWU64 the prior lives on all $L$ classes while each posterior is restricted to its candidate set, which
is why the E-step normalises over $C(x)$ only. This is the EM of \citet{saerens2002adjusting} with the pair
$z=(x,C(x))$ as the observation in place of $x$, and it inherits that method's justification if $p(z\mid y)$ is the
same in source and target. That holds in our design: on CLINC150 and HWU64 the candidate set is the fixed group $G(y)$
of the gold label, and on the other datasets it contains every label, so in both cases $C$ is determined by $y$ and
$p(z\mid y)=p(x\mid y)\,\mathbf{1}[C=G(y)]$ is unchanged by a prior shift. The order of the options carries no
information about $y$, because the gold position is counterbalanced (\hyperref[sec:protocol]{Section~\ref*{sec:protocol}}).
Bayes' rule then gives $p_t(y\mid z)\propto p_s(y\mid z)\,q(y)/\pi_s(y)$ for $y\in C(x)$ and zero outside, the E-step of
\hyperref[eq:em]{Eq.~\ref*{eq:em}}, and the M-step is the fixed point of the likelihood of the observed $z$. Two
consequences follow. Because $C(x)$ reveals the group, EM only redistributes probability among the classes of a group:
the estimated mass of each group equals its share of the shifted sample. And the source prior must be the one with which
the posteriors $p_s(y\mid z)$ are consistent, which for a zero-shot model is its implicit prior rather than the training
prior. Our shift simulation resamples whole examples with their candidate sets, so $p(z\mid y)$ is unchanged by
construction. In a deployment, an upstream step would choose the group. If that step is a fixed function of the input,
its choices alone do not break the invariance of $p(z\mid y)$ under label shift; what fails is support: when the chosen
group excludes the gold label, the restricted posterior gives that label probability zero, and no prior correction can
recover it. A simulation with exactly calibrated posteriors (exploratory; \hyperref[tab:emsim]{Table~\ref*{tab:emsim}})
bears this out: with the gold label's group as candidate set, the EM estimate of the target prior is almost as accurate
as the sample's own class shares, and it degrades steadily as a group selector misses the gold label's group more
often.

\paragraph{Metrics.}
Our main calibration metric is top-label expected calibration error with 15 equal-width confidence bins
$B_b=[(b-1)/15,\,b/15)$, the last bin closed \citep{guo2017calibration},
\begin{equation}
  \mathrm{ECE}(S)=\sum_{b=1}^{15}\frac{|B_b|}{|S|}\,\bigl|\mathrm{acc}(B_b)-\mathrm{conf}(B_b)\bigr|,
  \label{eq:ece}
\end{equation}
where $\mathrm{acc}$ and $\mathrm{conf}$ are the accuracy and mean confidence of the examples of $S$ in a bin. Selective
prediction accepts an example if its confidence $c_i$ reaches a threshold $\tau$, chosen on the development set as the
$\lceil 0.8\,n_{\mathrm{dev}}\rceil$-th largest confidence, so that (with ties, at least) 80\% of development examples are
accepted; each recalibration arm has its own threshold. Its selective risk \citep{geifman2017selective} on a sample $S$ and, under
shift, its \emph{excess risk} relative to the unshifted test sample are
\begin{equation}
  R(S;\tau)=\frac{\sum_{i\in S}\mathbf{1}[c_i\ge\tau]\,\mathbf{1}[\hat y_i\neq y_i]}{\sum_{i\in S}\mathbf{1}[c_i\ge\tau]},
  \qquad
  \Delta R(S_q)=R(S_q;\tau)-R(T;\tau),
  \label{eq:risk}
\end{equation}
where $T$ is the unshifted test sample. We also report Macro-F1, accuracy, MCC, Brier score, negative log-likelihood
(NLL, with probabilities clipped at $10^{-12}$, since Jev returns exact zeros) and the area under the risk--coverage
curve (AURC). Jev assigns probability zero to the gold label on 0.4--7\% of test examples (most on HateXplain), so its
NLL is dominated by the clip and is reported for completeness only. Because top-label ECE is biased upward on resampled data, we also report the change in Brier score and NLL
under shift (\hyperref[tab:brier]{Tables~\ref*{tab:brier}} and~\ref{tab:nll}).

\paragraph{Test statistics.}
P1 and P2 compare a TDM with its own base model on the same test examples, $\Delta=m(\mathrm{TDM})-m(\mathrm{base})$ for
$m\in\{\text{Macro-F1},\mathrm{ECE}\}$. PA1--PA3 are defined on a fixed set of $M=50$ target priors $q_1,\dots,q_M$,
drawn once from \hyperref[eq:shift]{Eq.~\ref*{eq:shift}} at $\alpha=1$. For a test sample $T$, let $S_m$ be its shifted sample under $q_m$ and
$N_m$ a no-shift control drawn by the same resampler at $T$'s own class distribution. With $J$ for Jev and $D$ for
DeBERTa-v3-large,
\begin{align}
  \mathrm{PA1}&=\frac{1}{M}\sum_{m=1}^{M}\Bigl(\bigl[\mathrm{ECE}_J(S_m)-\mathrm{ECE}_J(N_m)\bigr]
    -\bigl[\mathrm{ECE}_D(S_m)-\mathrm{ECE}_D(N_m)\bigr]\Bigr),\label{eq:pa1}\\
  \mathrm{PA2}&=\frac{1}{M}\sum_{m=1}^{M}\Bigl(\Delta R_J(S_m)-\Delta R_D(S_m)\Bigr),\label{eq:pa2}\\
  \mathrm{PA3}&=\frac{1}{M}\sum_{m=1}^{M}\Bigl(\mathrm{ECE}^{\mathrm{EM}}_J(S_m)-\mathrm{ECE}_J(S_m)\Bigr),\label{eq:pa3}
\end{align}
where each system uses its own threshold in $\Delta R$ and $\mathrm{ECE}^{\mathrm{EM}}_J$ is computed after EM with
Jev's implicit prior. Positive PA1 and PA2 mean that Jev degrades more than DeBERTa-v3-large under shift; negative PA3
means that EM helps. PA1 subtracts a resampled control because ECE is biased upward on resampled data, whereas PA2 does
not need one because selective risk, a mean of per-example losses over the accepted examples, is nearly unbiased under
resampling. \hyperref[sec:protocol]{Section~\ref*{sec:protocol}} describes how $T$ is
resampled to obtain intervals.

\paragraph{Hypotheses.}
RQ1--RQ3 of \hyperref[sec:intro]{Section~\ref*{sec:intro}} map to the following tests, all specified before the final
test-set runs.
\textbf{H1} (RQ1): vendor decision training changes Macro-F1 or ECE relative to the base model (two-sided). The
controlled test is P2 (Nimble-9B vs Qwen3.5-9B: same base weights and precision, differing only by an adapter). P1
(Tev1-4B vs Qwen3.5-4B) is the closest available comparison, since Tev1 is a full fine-tune whose exact recipe is not
published.
\textbf{H2} (RQ2): under shift, the change in ECE (PA1, \hyperref[eq:pa1]{Eq.~\ref*{eq:pa1}}) and the excess threshold risk (PA2,
\hyperref[eq:pa2]{Eq.~\ref*{eq:pa2}}) differ between Jev and DeBERTa-v3-large (two-sided).
\textbf{H3} (RQ3): EM prior correction with Jev's implicit prior lowers Jev's ECE under shift (PA3, \hyperref[eq:pa3]{Eq.~\ref*{eq:pa3}},
tested two-sided, which is conservative for this directional hypothesis). This confirmatory test concerns plain EM on
Jev only; the other arms and systems are reported descriptively.

\paragraph{Label budgets and crossover.}
RQ4 is answered by estimation, not by a test. For a trained system $s$ with $b$ training labels and seeds
$k=1,\dots,K$, let $\Delta(b)=\frac{1}{K}\sum_{k}m(s_{b,k})-m(J)$ be the paired Macro-F1 difference to Jev on the same
test examples. Its 95\% interval comes from a bootstrap that resamples both the $K$ seeds and the test examples
($B=10{,}000$), so it covers training variation as well as test-set noise. The \emph{crossover budget} is the smallest
evaluated budget $b$ whose interval lies above zero; we report it as the interval between that budget and the one before
it, or as ``not reached'' if no budget up to the full training set qualifies. As a rough guide, we also give the budget
at which the linearly interpolated point estimate first crosses zero \citep{lescao2021howmany}.

\section{Experimental protocol}
\label{sec:protocol}

\paragraph{Datasets.}
We use five public, human-labelled English datasets that represent common decision tasks
(\hyperref[tab:datasets]{Table~\ref*{tab:datasets}}): intent classification for assistants, CLINC150 in-scope \citep{larson2019evaluation} and
HWU64 \citep{liu2019benchmarking}; financial sentiment, FPB \citep{malo2014good}; hate and offensive speech,
HateXplain \citep{mathew2021hatexplain}; and question type, TREC-6 \citep{li2002learning}. CLINC150, HateXplain and
TREC-6 use their official splits. FPB has no official split; we use a public 64/16/20 split of the 50\%-agreement subset,
following \citet{araci2019finbert}, whose 970-example test set is the one used by \citet{deusser2026jev}. For HWU64 and TREC-6, which have no official development
split, the development set is a stratified sample of the training set that follows the training class distribution.
Train and development texts that duplicate a test text after normalisation (lower-casing and removing punctuation),
and texts that occur with conflicting labels, are removed; the test sets are never changed. All raw files and processed
splits are pinned by SHA-256 hashes.

\begin{table}[t]
  \centering
  \caption{\textbf{The five datasets span intent, sentiment, moderation and question-type decisions with 3 to 150
  classes.} CLINC150 and HWU64 use fixed label groups as candidate sets ($K$ = 18--22 options); the other three use all
  labels. Train/dev/test sizes after cleaning.}
  \label{tab:datasets}
  \scriptsize
  \begin{tabular}{llrrrrr}
\toprule
Dataset & Domain & Classes & Candidates $K$ & Train & Dev & Test \\
\midrule
CLINC150 & intent (assistant) & 150 & 18--19 & 14,939 & 2,994 & 4,500 \\
HWU64 & intent (home assistant) & 64 & 21--22 & 7,871 & 1,076 & 1,076 \\
FPB & financial sentiment & 3 & 3 & 3,090 & 775 & 970 \\
HateXplain & hate / offensive speech & 3 & 3 & 15,330 & 1,921 & 1,924 \\
TREC-6 & question type & 6 & 6 & 4,869 & 498 & 500 \\
\bottomrule
\end{tabular}

\end{table}

\paragraph{Label texts and candidate sets.}
Every label has a short name and a one-sentence description, written once for all systems and frozen before any model
was run. For datasets with at most 20 labels, the candidate set contains all labels. For CLINC150 and HWU64 we
partition the labels once into fixed groups (8 groups of 18--19 and 3 groups of 21--22) by balanced $k$-means on
embeddings of the label texts, and the candidate set of an example is the group of its gold label. We chose fixed groups
because a leakage audit showed that candidate sets built around the gold label (the gold label plus its nearest
neighbours) reveal the gold label from the option list alone: a classifier that never sees the text identifies it for
97.6\% of CLINC150 and 90.2\% of HWU64 examples, against 5\% chance. With fixed groups, every member of a group receives
the same set of options. The order of the options is shuffled per example with a fixed seed, with the gold position
counterbalanced across examples, and all systems read the same candidate file.

\paragraph{Systems and inputs.}
\hyperref[tab:systems]{Table~\ref*{tab:systems}} lists the systems. Jev is queried through its hosted API, with the label names as option keys
and the label descriptions as criteria; the response's version field was \texttt{jev-1.13.0} for every call (16{,}234 in the full run and 7{,}500 in
three repeats); 465 of the 16{,}234 full-run calls (464 rate-limit responses and one timeout) succeeded after one
retry, and no call failed. Jev's API returns its probabilities rounded to two decimals; we use them as returned and renormalise over the candidates. Each open TDM and its base model receive \emph{token-identical} prompts
in the TDM's own training format: the Tev1 format for Tev1-4B and Qwen3.5-4B, and the Nimble format for Nimble-9B and
Qwen3.5-9B (checked on all 16{,}234 development and test prompts). Thinking mode is disabled. The probability vector is
the softmax ($T=1$) of the candidate letters' logits at the first answer position, computed in fp32 from the final
hidden state. The two 9B models run in 8-bit precision so that they fit on a 12\,GB GPU, and the Nimble adapter is
checked to be fully loaded before every run; the two 4B models run in bf16. Local scoring was deterministic in our check:
scoring 50 CLINC150 development examples twice gave identical probabilities for all four local models. Tev1 was trained mostly on decisions with at most
eight options, so the 18--22 options of CLINC150 and HWU64 lie outside its training distribution, and the licence of its
weights was still being finalised at the time of writing. Nimble-9B is the \texttt{original-2676} release, trained on
2{,}676 synthetic examples; we do not use the later release whose training data include public TREC questions
(\hyperref[app:runs]{Appendix~\ref*{app:runs}}).

\begin{table}[t]
  \centering
  \caption{\textbf{Each open TDM is paired with its own backbone under identical prompts, readout and precision.}
  Tev1-4B with Qwen3.5-4B (Tev1 prompt format) and Nimble-9B with Qwen3.5-9B (Nimble prompt format, identical 8-bit base
  weights). Revisions are Hugging Face commit hashes. Readout: ``letters'' is the softmax ($T=1$) over the logits of the
  candidate letters at the first answer position; the trained classifiers use a softmax over each example's candidates;
  Jev returns probabilities directly.}
  \label{tab:systems}
  \scriptsize
  \setlength{\tabcolsep}{4pt}
  \begin{tabular}{lllll}
\toprule
System & Source & Revision & Precision & Readout \\
\midrule
\textsc{Typed decision models} & & & & \\
Jev & Cloudflare API (\texttt{typesafe/jev}) & \texttt{jev-1.13.0} & API & API probabilities \\
Tev1-4B & \texttt{togethercomputer/Tev1-4B-experimental} & \texttt{0b7becf0} & bf16 & letters, Tev1 format \\
Nimble-9B & \makecell[l]{Qwen3.5-9B + LoRA \\ \texttt{bespokelabs/Bespoke-Nimble-9B}} & \makecell[l]{\texttt{fe33f82b} \\ (\texttt{original-2676})} & 8-bit & letters, Nimble format \\
\midrule
\textsc{Base LLMs} & & & & \\
Qwen3.5-4B & \texttt{Qwen/Qwen3.5-4B} & \texttt{851bf6e8} & bf16 & as Tev1-4B \\
Qwen3.5-9B & \texttt{Qwen/Qwen3.5-9B} & \texttt{c2022362} & 8-bit & as Nimble-9B \\
\midrule
\textsc{Trained classifiers} & & & & \\
DeBERTa-v3-large & fine-tuned on the full training set, 5 seeds & \texttt{64a8c8ea} & fp16 & softmax \\
TF-IDF + LR & word 1--2 grams, $C$ selected on dev & scikit-learn 1.9.1 & CPU & softmax \\
\bottomrule
\end{tabular}

\end{table}

\paragraph{Trained classifiers.}
DeBERTa-v3-large \citep{he2023debertav3} is fine-tuned on the full training set with AdamW, a learning rate selected
from $\{10^{-5}, 2\cdot10^{-5}\}$ on the development set, 10\% linear warm-up, weight decay 0.01, batch size 16, mixed
precision, maximum length 128 and 5 epochs. The best checkpoint is selected by development Macro-F1 within the candidate
sets, evaluated every 100 steps. We train five seeds; a run would count as failed if its development accuracy did not
exceed that of a majority-class predictor \citep{dodge2020finetuning,mosbach2021stability}, which did not happen. All analyses use the first
seed; the other seeds are reported in \hyperref[tab:seeds]{Table~\ref*{tab:seeds}}. The TF-IDF baseline is a logistic regression on word
uni- and bigrams with sublinear term frequency; $C$ is selected on the development set from
$\{0.1, 1, 10, 10^2, 10^3, 10^4\}$, and FPB and TREC-6 select the largest value; extending the grid to $10^5$ and
$10^6$ changes their development Macro-F1 by at most 0.2~points and their test Macro-F1 by at most 0.7~points
(exploratory check). Both classifiers' probabilities are
renormalised over each example's candidate set. Note an asymmetry: the trained classifiers use the development set both
for model selection and for recalibration, whereas the zero-shot systems use it only for recalibration and the implicit
prior.

\paragraph{Label budgets.}
For RQ4, both classifiers are also trained on stratified subsets of 100, 250, 500, 1{,}000 and 2{,}500 training
examples, and of 5{,}000 where the training set is larger (HateXplain, HWU64, CLINC150), with five subset seeds per
budget; the full training set is the last point of each curve, with all five DeBERTa-v3-large seeds. Class quotas are
proportional to the training frequencies, with at least one example per class where the budget allows it; at 100
labels, CLINC150 has examples for only 100 of its 150 classes (\hyperref[tab:shots]{Table~\ref*{tab:shots}}).
DeBERTa-v3-large uses the settings above with a fixed schedule: 20 epochs up to 500 labels and 5 epochs above, with at
least 500 optimizer steps, following the advice for few-sample fine-tuning
\citep{mosbach2021stability,zhang2021revisiting}. The learning rate, $C$ and the checkpoint are selected on the full
development set at every budget. A failed run would be restarted with one of three reserve seeds per dataset and
budget; none failed.

\paragraph{Task fine-tuning of the base model.}
Condition (b) of the decision-training comparison fine-tunes Qwen3.5-4B on the target task with QLoRA
\citep{hu2022lora,dettmers2023qlora} at 100, 500 and 2{,}500 labels, using the same subsets as above and three seeds. The
adapter has rank 16, $\alpha=32$ and dropout 0.05 on the same projection matrices as the Nimble-9B adapter; it is trained
on a 4-bit NF4 copy of the base with learning rate $10^{-4}$, effective batch size 16, the DeBERTa schedule and a
cross-entropy loss over the candidate letters at the answer position. Checkpoints are selected by development Macro-F1
every 100 steps. Every reported prediction comes from the kept adapter on the bf16 base, with the same prompt, assistant
prefix and letter readout as Qwen3.5-4B and Tev1-4B, so the three 4B conditions (a) frozen base, (b) task QLoRA and (c)
Tev1-4B differ only in their weights.

\paragraph{Statistical tests.}
The 35 confirmatory tests are P1 and P2 (Macro-F1 and ECE on five datasets each; 20 tests) and PA1--PA3 (five datasets
each; 15 tests). P1 and P2 use a paired bootstrap over test examples with $B=10{,}000$ replicates; both systems and both
metrics use the same resampled indices. For PA1--PA3 (\hyperref[eq:pa1]{Eqs.~\ref*{eq:pa1}}--\nobreak\ref{eq:pa3}), the estimand is the mean over a fixed set of 50 target priors
drawn once at $\alpha=1$ (centred on the training prior); it is therefore conditional on this set of priors, and the
intervals reflect only the finiteness of the test set. Each of $B=10{,}000$ replicates draws a class-stratified bootstrap
sample of the test set and builds all 50 shifted samples from it by resampling with replacement, so that Jev and
DeBERTa-v3-large are evaluated on identical samples. PA1 compares each shifted ECE with a no-shift control drawn by the
same resampler at the base sample's own class prior, because top-label ECE is biased upward on resampled data. PA2
compares risks, which are nearly unbiased under resampling, with the base sample itself. The point estimates of P1 and P2 are the differences on the full test set, those of
PA1--PA3 the means of the bootstrap replicates. We report percentile 95\% intervals,
which are not adjusted for multiplicity, and two-sided $p=\min\{1,\,2\min\{\Pr(\Delta\le0),\Pr(\Delta\ge0)\}\}$, adjusted with the
Holm procedure \citep{holm1979simple} over all 35 tests (also for PA3, whose hypothesis is directional). With
$B=10{,}000$, a raw $p$ of zero is reported as $p<0.0002$ $(=2/B)$, and its Holm-adjusted value as $p<0.007$
$(=35\cdot2/B)$. Quantities fitted on the development set (thresholds, implicit priors, temperatures) are held fixed
within the bootstrap, so the intervals do not include development-set uncertainty. The descriptive prevalence-shift
experiment uses 200 Dirichlet draws per $\alpha$ for all systems and all five arms, with and without replacement.

\paragraph{Protocol and runs.}
Hypotheses, tests, the comparison family, all configurations and the code were fixed and recorded as SHA-256 hashes
before the final runs, and every final run was verified against these hashes. One amendment followed three
code audits; it changed the analysis, the DeBERTa failed-run rule and the Jev error handling, but not the code used by
the local language models. It was made after the first local test predictions had been produced, but before the
DeBERTa and Jev runs and before any of the 35 tests or any shift analysis was computed on the final runs
(\hyperref[app:deviations]{Appendix~\ref*{app:deviations}}). Before the freeze, some pipeline checks were run on test
splits, mainly FPB's, including early versions of PA1--PA3 and of the shift experiment on FPB;
\hyperref[app:deviations]{Appendix~\ref*{app:deviations}} discloses them.
The local models and both trained classifiers were run on one NVIDIA RTX 4070 Ti (12\,GB); Jev, a hosted model, was
queried through its API for the main runs in about two hours, inside the protocol's 72-hour window (\hyperref[app:runs]{Appendix~\ref*{app:runs}}).
The label-budget study was specified with the rest of the protocol; its QLoRA settings, trainer and curve analysis were
fixed and hashed in a second freeze before any of its final test predictions, after a development-only pilot that left every
setting unchanged (\hyperref[app:deviations]{Appendix~\ref*{app:deviations}}). It reuses the Jev and zero-shot
predictions of the main study.

\section{Results}
\label{sec:results}

\paragraph{Validity checks.}
All 16{,}234 Jev calls and all local predictions returned a valid probability vector: no request failed after retries
and no input was refused, including on HateXplain. On the 970-example FPB test set that \citet{deusser2026jev} also use,
Jev's accuracy is 0.744, consistent with the 0.730 they report. Full-data DeBERTa-v3-large reaches 0.976 accuracy on
TREC-6, close to the 0.974 reported for fully fine-tuned RoBERTa-large \citep{gao2021making}, and 0.703 Macro-F1 on
HateXplain, slightly above the best value we found reported for that test split, 0.699 \citep{kim2022why}. The four local
models gave identical probabilities when 50 CLINC150 development examples were scored twice. Jev is not: in four runs over the same 500 test examples per dataset, it returned the same
prediction for 96.4--99.8\% of examples but the same probability vector for only 19.4--87.4\% of them, with differences of
up to 0.29 between runs; its ECE on these subsets varied by at most 0.012 between runs (\hyperref[app:results]{Appendix~\ref*{app:results}}). All
analyses use a single full run of each system.

\begin{figure}[t]
  \centering
  \includegraphics[width=\linewidth]{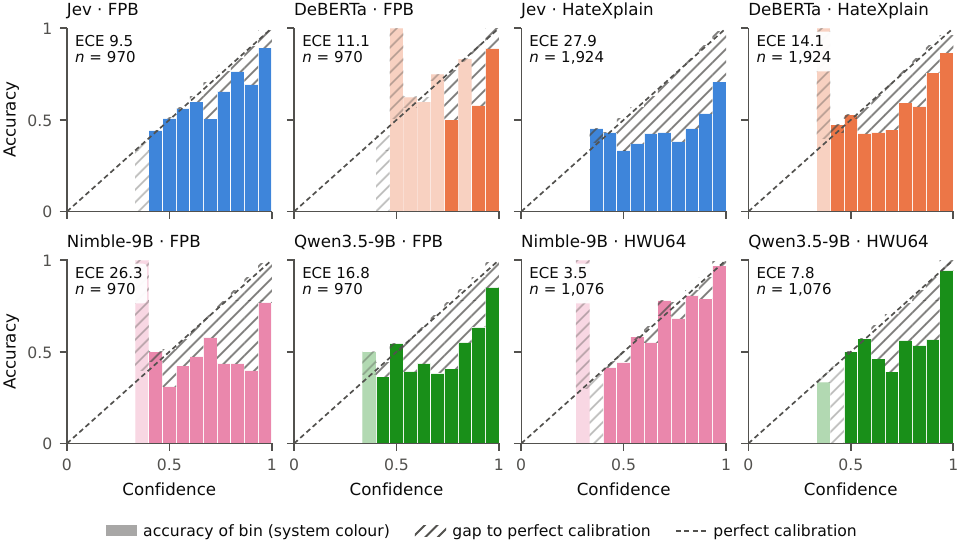}
  \caption{\textbf{On FPB and HateXplain every system shown is overconfident at high confidence, and decision training
  moves Nimble-9B away from the diagonal on FPB but towards it on HWU64.} Reliability diagrams on the test set
  without shift: each bar is the accuracy of one of the 15 equal-width confidence bins of the ECE, and the hatched part
  is its gap to the diagonal (perfect calibration); faint bars hold fewer than 10 examples (at most 2.7\% of a test
  set). ECE in points, as in \hyperref[tab:exp1]{Table~\ref*{tab:exp1}}. Top row: Jev and DeBERTa-v3-large (seed 1001,
  the model of the table and of the tests); bottom row: Nimble-9B and its base Qwen3.5-9B.}
  \label{fig:reliability}
\end{figure}

\paragraph{Overview.}
\hyperref[tab:exp1]{Table~\ref*{tab:exp1}} gives the test-set results without shift. Full-data DeBERTa-v3-large has the highest Macro-F1 point
estimate on all five datasets, although Jev is within 0.3~points on CLINC150. Among the zero-shot systems in
\hyperref[tab:exp1]{Table~\ref*{tab:exp1}}, Jev has the highest Macro-F1 point estimate on every dataset (by 0.3~points over Qwen3.5-9B on
FPB). No zero-shot system is reliable on HateXplain (Macro-F1 at most 0.514), where even TF-IDF reaches 0.645.
Calibration varies more by task than by system class: ECE is at most 0.020 for all zero-shot systems on CLINC150, but
between 0.214 and 0.340 on HateXplain. The reliability diagrams in \hyperref[fig:reliability]{Figure~\ref*{fig:reliability}}
show the pattern behind these numbers. On HateXplain, Jev gives a confidence of at least 0.9 to 45.5\% of the test
examples, of which 67.4\% are correct (DeBERTa-v3-large: 54.1\% and 84.2\%). On FPB, 71.2\% of the examples to which
Nimble-9B gives at least 0.9 are correct, against 83.8\% for Qwen3.5-9B; on HWU64 the order is reversed (96.0\% against
93.9\%). These comparisons are descriptive; the tested comparisons follow.

\begin{table}[t]
  \centering
  \caption{\textbf{Full-data DeBERTa-v3-large has the highest Macro-F1 point estimate on every dataset, and the
  calibration of zero-shot systems depends more on the task than on the system.} Test-set Macro-F1 and ECE (15 bins,
  top label), candidate-restricted, no shift. These are point estimates; tested comparisons are in
  \hyperref[fig:pairs]{Figure~\ref*{fig:pairs}} and \hyperref[tab:pa]{Table~\ref*{tab:pa}}.}
  \label{tab:exp1}
  \scriptsize
  \setlength{\tabcolsep}{3pt}
  \begin{tabular}{lrrrrrrrrrr}
\toprule
 & \multicolumn{2}{c}{CLINC150} & \multicolumn{2}{c}{HWU64} & \multicolumn{2}{c}{FPB} & \multicolumn{2}{c}{HateXplain} & \multicolumn{2}{c}{TREC-6} \\
\cmidrule(lr){2-3} \cmidrule(lr){4-5} \cmidrule(lr){6-7} \cmidrule(lr){8-9} \cmidrule(lr){10-11}
System & F1$\uparrow$ & ECE$\downarrow$ & F1$\uparrow$ & ECE$\downarrow$ & F1$\uparrow$ & ECE$\downarrow$ & F1$\uparrow$ & ECE$\downarrow$ & F1$\uparrow$ & ECE$\downarrow$ \\
\midrule
\multicolumn{11}{l}{\textsc{Typed decision models}} \\
Jev & 0.978 & 0.007 & 0.908 & 0.036 & 0.752 & 0.095 & 0.514 & 0.279 & 0.922 & 0.042 \\
Tev1-4B & 0.964 & 0.008 & 0.895 & 0.040 & 0.735 & 0.072 & 0.458 & 0.214 & 0.889 & 0.038 \\
Nimble-9B & 0.962 & 0.008 & 0.896 & 0.035 & 0.629 & 0.263 & 0.470 & 0.336 & 0.872 & 0.043 \\
\midrule
\multicolumn{11}{l}{\textsc{Base LLMs}} \\
Qwen3.5-4B & 0.939 & 0.011 & 0.866 & 0.058 & 0.635 & 0.204 & 0.441 & 0.247 & 0.882 & 0.037 \\
Qwen3.5-9B & 0.955 & 0.020 & 0.884 & 0.078 & 0.749 & 0.168 & 0.487 & 0.340 & 0.903 & 0.061 \\
\midrule
\multicolumn{11}{l}{\textsc{Trained classifiers (full data)}} \\
DeBERTa-v3-large & 0.980 & 0.013 & 0.939 & 0.030 & 0.870 & 0.111 & 0.703 & 0.141 & 0.971 & 0.021 \\
TF-IDF + LR & 0.950 & 0.015 & 0.900 & 0.098 & 0.693 & 0.167 & 0.645 & 0.069 & 0.891 & 0.071 \\
\bottomrule
\end{tabular}

\end{table}

\subsection{RQ1: What decision training changes}
\label{sec:rq1}

\begin{figure}[t]
  \centering
  \includegraphics[width=\linewidth]{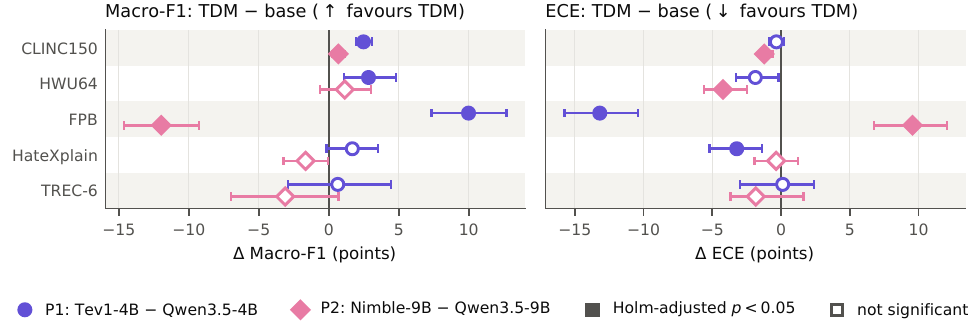}
  \caption{\textbf{In their TDM's own format, Tev1-4B is never significantly worse than its base model, whereas
  Nimble-9B is better calibrated on intent tasks and 12 Macro-F1~points worse on FPB.} Paired differences (TDM $-$ base) in percentage points with 95\%
  paired bootstrap intervals ($B=10{,}000$, unadjusted). Filled markers: significant after Holm correction over all 35
  primary tests.}
  \label{fig:pairs}
\end{figure}

\paragraph{Decision training improves the 9B backbone's calibration on intent tasks and lowers its Macro-F1 on financial
sentiment by 12~points.}
Nimble-9B differs from Qwen3.5-9B only by a LoRA adapter: the base weights, 8-bit precision, prompts and readout are
identical. On CLINC150 the adapter lowers ECE by 1.2~points (95\% CI $[-1.5,-0.6]$; Holm-adjusted $p<0.007$) and raises
Macro-F1 by 0.7~points ($[0.2, 1.2]$; $p=0.029$). On HWU64 it lowers ECE by 4.2~points ($[-5.6,-2.5]$; $p<0.007$). On FPB,
by contrast, Nimble-9B's Macro-F1 is 12.0~points below its base ($[-14.6,-9.3]$; $p<0.007$) and its ECE is 9.6~points
higher ($[6.8, 12.1]$; $p<0.007$). On HateXplain and TREC-6 neither difference is significant after correction; the
largest of these, $-3.1$ Macro-F1~points on TREC-6, has an interval of $[-7.0, 0.7]$ (\hyperref[fig:pairs]{Figure~\ref*{fig:pairs}}).

\paragraph{In its own format, Tev1-4B is never significantly worse than its base model.}
Tev1-4B's Macro-F1 is higher than Qwen3.5-4B's on CLINC150 ($+2.5$, $[1.9, 3.1]$; $p<0.007$), HWU64 ($+2.8$,
$[1.1, 4.8]$; $p=0.040$) and FPB ($+10.0$, $[7.4, 12.7]$; $p<0.007$), and its ECE is lower on FPB ($-13.2$,
$[-15.7,-10.4]$; $p<0.007$) and HateXplain ($-3.2$, $[-5.2,-1.4]$; $p=0.029$). None of the ten differences favours the
base model significantly. Because Tev1 is a full fine-tune whose exact base revision and recipe are not published, and
because the 18--22 options of the intent datasets lie outside its training distribution, this pair is the closest
available comparison rather than a controlled one. On HWU64, 614 of the 1{,}076 test texts also occur in MASSIVE
\citep{fitzgerald2023massive}; on the 462 remaining examples every system, including the trained classifiers, loses 2.2
to 5.9 Macro-F1~points, the P2 calibration result holds ($-4.3$ ECE points, $[-5.8,-1.1]$), and the P1 Macro-F1
difference keeps its sign but is no longer resolved ($+3.0$, $[-0.03, 8.3]$; exploratory, \hyperref[tab:hwu64]{Table~\ref*{tab:hwu64}}).

The two pairs thus point in opposite directions on FPB: Tev1-4B scores 10.0 Macro-F1~points above its base, and
Nimble-9B 12.0~points below its own. Nimble-9B's loss is concentrated on the neutral class: it labels 285 of the 565
neutral test sentences as positive, against 159 for Qwen3.5-9B, so that its recall on neutral sentences falls from 0.64
to 0.36 while its recall on positive and negative sentences rises (0.88 to 0.96 and 0.89 to 0.95; descriptive,
\hyperref[tab:fpbconf]{Table~\ref*{tab:fpbconf}}). A later Nimble release, whose training data include public TREC questions, scores
0.963 Macro-F1 on TREC-6 against 0.872 for the release we test (exploratory; \hyperref[app:results]{Appendix~\ref*{app:results}}), which
suggests that the training data, not only decision training, affect the result.

\begin{figure}[t]
  \centering
  \includegraphics[width=\linewidth]{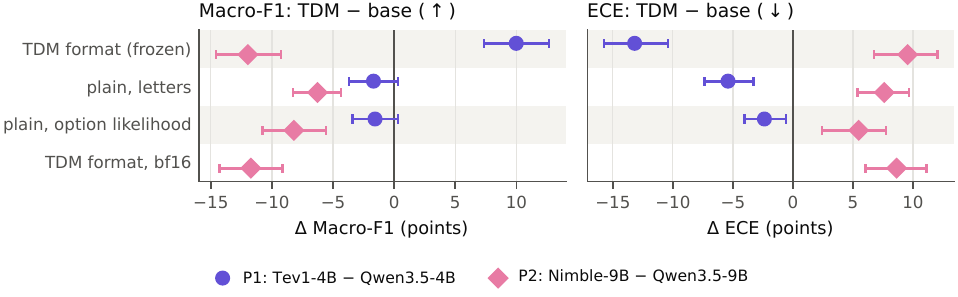}
  \caption{\textbf{Outside the Tev1 format the Tev1-4B Macro-F1 gain on FPB disappears; the Nimble-9B loss remains in
  every format and in bf16 (exploratory).} Paired differences (TDM $-$ base) on the FPB test set in percentage points
  with 95\% paired bootstrap intervals ($B=10{,}000$, unadjusted). First row: the frozen P1 and P2 in each TDM's
  format; plain formats in \hyperref[app:prompts]{Appendix~\ref*{app:prompts}}; last row: Nimble-9B and Qwen3.5-9B in
  bf16 instead of 8-bit. Values in \hyperref[tab:promptprec]{Table~\ref*{tab:promptprec}}.}
  \label{fig:promptprec}
\end{figure}

\paragraph{The Tev1-4B gain on FPB is a prompt-format effect; the Nimble-9B loss is not (exploratory).}
P1 and P2 score each base model in its TDM's training format. Re-scored in two generic formats that belong to neither
TDM (plain multiple choice with a letter readout, and option likelihood over the label names;
\hyperref[app:prompts]{Appendix~\ref*{app:prompts}}), Qwen3.5-4B reaches 73.1 and 72.2 Macro-F1 on FPB against 63.5 in
the Tev1 format, and the P1 difference becomes $-1.7$ ($[-3.7, 0.3]$) and $-1.6$ ($[-3.4, 0.3]$); Tev1-4B in its own
format is 0.4~points above Qwen3.5-4B in its best format, which was chosen on the test set and so favours the base.
Tev1-4B's ECE stays lower ($-5.4$ and $-2.4$~points). P2 keeps its sign in every format ($-6.3$, $[-8.3,-4.3]$;
$-8.2$, $[-10.8,-5.6]$) and in bf16 ($-11.7$, $[-14.3,-9.1]$), so the Nimble result is not an artefact of the format
or of 8-bit precision (\hyperref[fig:promptprec]{Figure~\ref*{fig:promptprec}}).

\subsection{RQ2: Prevalence shift}
\label{sec:rq2}

\paragraph{At moderate shift, Jev and DeBERTa-v3-large are distinguishable after correction only on FPB, and the size
of that difference depends on the shift scenarios.}
On the fixed set of 50 target priors, the shift-induced change in ECE at $\alpha=1$ is 1.7~points larger for Jev than for
DeBERTa-v3-large on FPB (PA1; $[1.0, 2.5]$; $p<0.007$), and the excess risk of the 80\%-coverage threshold is 2.5~points
larger ($[1.8, 3.2]$; $p<0.007$; PA2). On the other four datasets neither difference is significant after correction
(\hyperref[tab:pa]{Table~\ref*{tab:pa}}); on CLINC150 and HWU64 the intervals lie within $\pm0.5$ points.

\begin{figure}[t]
  \centering
  \includegraphics[width=\linewidth]{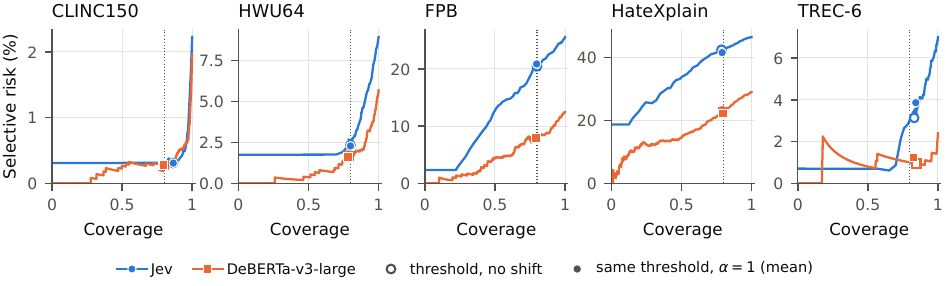}
  \caption{\textbf{Under moderate shift, the development-set threshold lands close to where it does without shift,
  whereas the risk gap between Jev and DeBERTa-v3-large at that threshold is large (descriptive).} Lines: selective risk
  (error rate among the accepted examples) against coverage on the original test set, examples ordered by decreasing
  confidence and ties averaged; the mean of each curve is the AURC. Open markers: the threshold fixed on the development
  set for 80\% coverage (dotted line), applied to the original test set; filled markers: the same threshold, mean
  coverage and risk over the 200 shifted test sets at $\alpha=1$. The y-axis scale differs between panels.}
  \label{fig:riskcov}
\end{figure}

\paragraph{The threshold moves little under moderate shift compared with the gap between the systems (descriptive).}
\hyperref[fig:riskcov]{Figure~\ref*{fig:riskcov}} shows the full risk--coverage curves without shift and where the
threshold chosen on the development set for 80\% coverage lands. Averaged over the 200 shifted test sets of the descriptive shift experiment at
$\alpha=1$, Jev's selective risk at this threshold changes from 20.5\% to 20.9\% on FPB and from 3.1\% to 3.9\% on
TREC-6, DeBERTa-v3-large's by at most 0.3~points on any dataset, and the coverage of either system by at most 1.9~points.
The difference between the systems is larger: at the same threshold, DeBERTa-v3-large's selective risk is 7.9\% on FPB
and 22.4\% on HateXplain, against 20.5\% and 42.4\% for Jev.

Because these confirmatory estimates are conditional on one set of target priors, we repeated PA1--PA3 with five other
prior sets (exploratory, added after the results were known; \hyperref[tab:pasens]{Table~\ref*{tab:pasens}}). On FPB, Jev's ECE change exceeds
DeBERTa-v3-large's for every prior set, but by 0.3 to 1.7~points (the frozen set is among the largest), and its excess
risk by 0.6 to 2.5~points (frozen set 2.5). On TREC-6, Jev's ECE change is larger by 0.8 to 1.2~points for every set. On HateXplain the
direction depends on the prior set: the frozen set favours Jev ($-0.8$), the five others favour DeBERTa-v3-large ($+0.2$
to $+0.8$). Averaged over 2{,}000 target priors drawn at $\alpha=1$ (exploratory; \hyperref[tab:papop]{Table~\ref*{tab:papop}}), the PA1
difference is $+0.4$ points on FPB (Monte Carlo standard error over priors 0.12; it does not include test-set
uncertainty), $0.0$ on HateXplain and $+1.0$ on TREC-6, and at most $0.1$ on the intent datasets;
on FPB and HateXplain the difference varies widely between single priors (standard deviation 5.4 and 7.4~points), so a
set of 50 priors is not enough to pin down its mean. PA3, in contrast, keeps its sign and remains clearly negative:
$-1.7$ points on FPB (frozen set $-2.7$) and $-2.0$ on HateXplain (frozen set $-1.8$). We therefore read RQ2 as follows: under moderate shift, Jev's calibration error grows by at most 1.7~points more than that of a fine-tuned classifier on any prior set, and by at most one point on average over shifts;
the direction is consistent on FPB and TREC-6, and the size of such differences depends on which shifts are considered. Two facts call for caution about the significant FPB results (PA1 and PA2). FPB is the
dataset whose test split was used in pipeline checks before the protocol was frozen (\hyperref[app:deviations]{Appendix~\ref*{app:deviations}}),
and the size of the estimate is not stable across prior sets (0.3 to 1.7~points). Jev's own run-to-run variation in ECE
on FPB, measured on repeated calls over 500 test examples, is at most 0.6~points (\hyperref[tab:jevrepeat]{Table~\ref*{tab:jevrepeat}}), well below
the frozen estimate, so the non-determinism of the API is unlikely to explain the FPB result, whereas its size depends
strongly on the prior set. We therefore do not treat RQ2 as evidence that Jev is more fragile under shift than a trained classifier; the
robust conclusion is that the difference is small.

\begin{table}[t]
  \centering
  \caption{\textbf{The EM correction effect (PA3) is stable across shift scenarios, whereas the Jev--DeBERTa difference
  (PA1, PA2) varies in size, and on HateXplain in sign.} Frozen: the confirmatory estimate on the pre-specified set of 50
  target priors ($B=10{,}000$; $\dagger$: Holm-adjusted $p<0.05$). Exploratory: range of the estimates over five other sets
  of 50 priors ($B=1{,}000$ each) and the number of sets whose unadjusted 95\% interval excludes zero. Points.}
  \label{tab:pasens}
  \scriptsize
  \begin{tabular}{llccccc}
\toprule
Test & Prior set & CLINC150 & HWU64 & FPB & HateXplain & TREC-6 \\
\midrule
PA1 & frozen (confirmatory) & +0.1 & +0.1 & \textbf{+1.7}$^\dagger$ & $-$0.8 & +1.0 \\
 & 5 other sets: range & [0.0, +0.1] & [+0.1, +0.4] & [+0.3, +1.7] & [+0.2, +0.8] & [+0.8, +1.2] \\
 & CI excludes 0 & 1/5 & 0/5 & 3/5 & 4/5 & 5/5 \\
\midrule
PA2 & frozen (confirmatory) & 0.0 & 0.0 & \textbf{+2.5}$^\dagger$ & $-$0.6 & +0.5 \\
 & 5 other sets: range & [0.0, 0.0] & [0.0, +0.2] & [+0.6, +2.3] & [+0.6, +1.5] & [+0.4, +0.6] \\
 & CI excludes 0 & 0/5 & 0/5 & 5/5 & 5/5 & 0/5 \\
\midrule
PA3 & frozen (confirmatory) & $-$0.1 & +0.1 & \textbf{$-$2.7}$^\dagger$ & \textbf{$-$1.8}$^\dagger$ & $-$1.1 \\
 & 5 other sets: range & [$-$0.1, $-$0.1] & [$-$0.1, +0.1] & [$-$2.5, $-$1.7] & [$-$2.5, $-$1.8] & [$-$1.1, $-$0.9] \\
 & CI excludes 0 & 0/5 & 0/5 & 5/5 & 5/5 & 5/5 \\
\bottomrule
\end{tabular}

\end{table}

\begin{figure}[t]
  \centering
  \includegraphics[width=\linewidth]{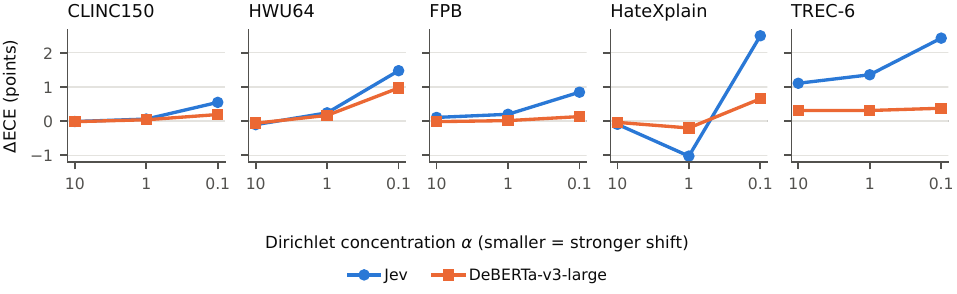}
  \caption{\textbf{Under stronger shift, Jev's calibration error grows more than DeBERTa-v3-large's on all five
  datasets, but this is not a zero-shot versus trained contrast: the TF-IDF classifier degrades more than Jev on two of
  them, and Tev1-4B more than Jev on FPB and TREC-6 (descriptive).}
  Change in ECE relative to a resampled no-shift control, mean over 200 Dirichlet draws per $\alpha$, raw probabilities.
  Only $\alpha=1$ is tested (\hyperref[tab:pa]{Table~\ref*{tab:pa}}); all systems in \hyperref[tab:exp2delta]{Table~\ref*{tab:exp2delta}}.}
  \label{fig:shift}
\end{figure}

\paragraph{Stronger shift widens the gap (descriptive).}
Measured against a resampled no-shift control, the mean change in ECE at $\alpha=1$ is small for every system (mostly
below 1.5~points). At $\alpha=0.1$ it grows, and on all five datasets Jev's ECE increases more than DeBERTa-v3-large's,
for example 2.5 versus 0.7~points on HateXplain and 2.4 versus 0.4~points on TREC-6 (\hyperref[fig:shift]{Figure~\ref*{fig:shift}}). This is not
a contrast between zero-shot and trained systems: the TF-IDF classifier's ECE increases more than Jev's on CLINC150 and
HateXplain, DeBERTa-v3-large is the most stable system on only three of the five datasets, and among the zero-shot
systems Tev1-4B (FPB $+4.8$, TREC-6 $+4.4$ points) and Qwen3.5-4B (FPB $+3.9$) degrade more than Jev
(\hyperref[tab:exp2delta]{Table~\ref*{tab:exp2delta}}). Jev is therefore not uniquely fragile. Without the resampled control, part of these changes would be resampling bias:
with no shift at all, resampling raises ECE by 0.05 to 1.4~points relative to the original test set.

\begin{table}[t]
  \centering
  \caption{\textbf{At moderate shift ($\alpha=1$), Jev and DeBERTa-v3-large are distinguishable only on FPB, and EM with
  Jev's implicit prior lowers its shifted ECE on FPB and HateXplain.} Estimates in percentage points with 95\% bootstrap
  intervals (unadjusted; class-stratified test bootstrap, $B=10{,}000$, conditional on 50 fixed target priors, common
  random numbers). $\dagger$ and bold: Holm-adjusted $p<0.05$ over the 35-test family; an unadjusted interval that excludes
  zero is not by itself significant.}
  \label{tab:pa}
  \scriptsize
  \begin{tabular}{lccccc}
\toprule
Test (points) & CLINC150 & HWU64 & FPB & HateXplain & TREC-6 \\
\midrule
PA1: $\Delta$ECE under shift, Jev $-$ DeBERTa & \makecell{+0.09\\{\scriptsize [$-$0.02, +0.20]}} & \makecell{+0.11\\{\scriptsize [$-$0.30, +0.50]}} & \makecell{\textbf{+1.71}$^\dagger$\\{\scriptsize [+0.98, +2.49]}} & \makecell{$-$0.78\\{\scriptsize [$-$1.29, $-$0.26]}} & \makecell{+1.01\\{\scriptsize [+0.26, +1.85]}} \\[2pt]
PA2: excess risk at $\tau_{80}$, Jev $-$ DeBERTa & \makecell{0.00\\{\scriptsize [$-$0.04, +0.04]}} & \makecell{+0.01\\{\scriptsize [$-$0.21, +0.23]}} & \makecell{\textbf{+2.49}$^\dagger$\\{\scriptsize [+1.83, +3.16]}} & \makecell{$-$0.55\\{\scriptsize [$-$0.98, $-$0.14]}} & \makecell{+0.55\\{\scriptsize [$-$0.06, +1.30]}} \\[2pt]
PA3: ECE, Jev EM $-$ none & \makecell{$-$0.12\\{\scriptsize [$-$0.29, +0.03]}} & \makecell{+0.12\\{\scriptsize [$-$0.33, +0.54]}} & \makecell{\textbf{$-$2.65}$^\dagger$\\{\scriptsize [$-$3.74, $-$1.61]}} & \makecell{\textbf{$-$1.75}$^\dagger$\\{\scriptsize [$-$2.71, $-$0.82]}} & \makecell{$-$1.06\\{\scriptsize [$-$1.96, $-$0.28]}} \\[2pt]
\bottomrule
\end{tabular}

\end{table}

\subsection{RQ3: Implicit-prior correction}
\label{sec:rq3}

\paragraph{EM with Jev's implicit prior lowers its calibration error under shift on the two datasets where that error is
highest.}
EM prior correction with Jev's implicit prior lowers Jev's ECE under shift on FPB by 2.7~points ($[-3.7,-1.6]$;
$p<0.007$; PA3) and on HateXplain by 1.8~points ($[-2.7,-0.8]$; $p<0.007$), the two datasets where Jev's unshifted ECE is
highest (0.095 and 0.279). On CLINC150 and HWU64, where its ECE is at most 0.036, the intervals lie between $-0.33$ and
$+0.54$ points; on TREC-6 the estimate is $-1.1$ points ($[-2.0,-0.3]$) but not significant after correction ($p=0.136$). Unlike PA1
and PA2, these effects keep their sign across shift scenarios: across the five other prior sets, the reduction is 1.7 to 2.5~points on FPB, 1.8 to 2.5 on HateXplain and 0.9 to 1.1 on TREC-6, with every unadjusted interval excluding zero
(\hyperref[tab:pasens]{Table~\ref*{tab:pasens}}). The FPB and HateXplain effects also exceed Jev's run-to-run variation in ECE (at most 0.6 and
1.1~points on these two datasets; \hyperref[tab:jevrepeat]{Table~\ref*{tab:jevrepeat}}). Descriptively, they appear to correct the shift rather
than Jev's label bias: on the resampled no-shift control, EM changes Jev's ECE by only $+0.1$ points on FPB and $+0.4$ on
HateXplain (descriptive; \hyperref[tab:emnoshift]{Table~\ref*{tab:emnoshift}}). The
correction is modest relative to the error it addresses: on HateXplain, Jev's mean ECE under shift falls from 0.271 to
0.254, whereas bias-corrected temperature scaling followed by EM, which needs development labels, reaches 0.071.
Descriptively, the same label-free correction lowers the shifted ECE of every zero-shot system on FPB and HateXplain, by
0.9 to 4.6~points (\hyperref[tab:exp2arms]{Table~\ref*{tab:exp2arms}}). Started from the true class prior
instead, EM raises the shifted ECE of every zero-shot system on FPB and HateXplain above the uncorrected value
(exploratory; \hyperref[tab:emtrueprior]{Table~\ref*{tab:emtrueprior}} and \hyperref[sec:discussion]{Section~\ref*{sec:discussion}}).

\paragraph{Content-free calibration is not a substitute for prior correction (exploratory).}
Content-free calibration \citep{zhao2021calibrate} divides each probability by the model's output on content-free
inputs (``N/A'', the empty string, ``[MASK]'') under the same candidate orders, and also needs no labels. On FPB and
HateXplain the label-level correction lowers the Macro-F1 of every zero-shot system, by 7.6 to 27.0~points, and raises
its ECE under shift (Jev on FPB: 0.107 to 0.472). On the other three datasets it changes Macro-F1 by at most
2.7~points, except for Nimble-9B on TREC-6 ($-9.8$). For Jev, which returns exact zeros, the correction is ill-posed:
its content-free weights differ by up to six orders of magnitude. Applied before EM, it does not recover the loss. EM with the implicit prior gives the
lowest or near-lowest ECE under shift on FPB, HateXplain and CLINC150, while on HWU64 and TREC-6 content-free correction
is slightly lower for some systems (\hyperref[tab:contentfree]{Table~\ref*{tab:contentfree}} in
\hyperref[app:revision]{Appendix~\ref*{app:revision}}).

\begin{figure}[t]
  \centering
  \includegraphics[width=\linewidth]{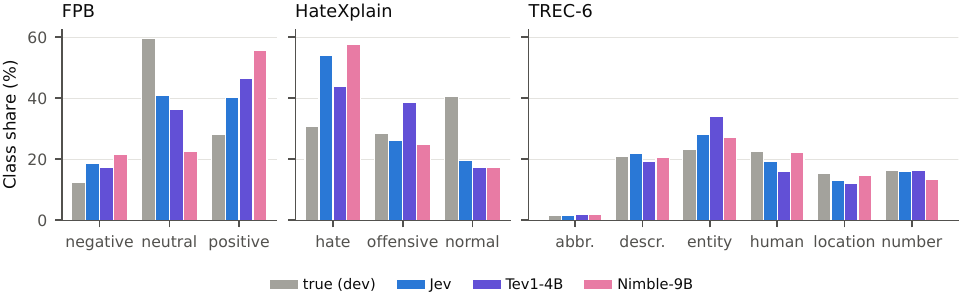}
  \caption{\textbf{The TDMs' implicit priors are far from the class distribution on FPB and HateXplain and close to it
  on TREC-6.} Implicit prior (mean predicted distribution on the development set, smoothed as in the descriptive shift experiment) of each TDM, and
  the class distribution of the development set (grey), in percent. Total-variation distances for all zero-shot systems
  and datasets in \hyperref[tab:implicitprior]{Table~\ref*{tab:implicitprior}}.}
  \label{fig:implicitprior}
\end{figure}

\paragraph{The implicit priors carry the label bias of the TDMs (descriptive).}
\hyperref[fig:implicitprior]{Figure~\ref*{fig:implicitprior}} compares the implicit priors of the three TDMs with the
class distribution of the development set on the three datasets with few classes. On FPB, Jev's implicit prior puts
41.0\% on the neutral class, which makes up 59.5\% of the examples, and Nimble-9B's only 22.7\%; on HateXplain, all three
put 44 to 58\% on hate speech, which makes up 30.9\%. On TREC-6, Jev's implicit prior is within 5.2~points of the class
share for every class. EM helps most on the two datasets where this bias is largest;
\hyperref[sec:discussion]{Section~\ref*{sec:discussion}} explains why the bias does not prevent the correction.

\begin{figure}[t]
  \centering
  \includegraphics[width=\linewidth]{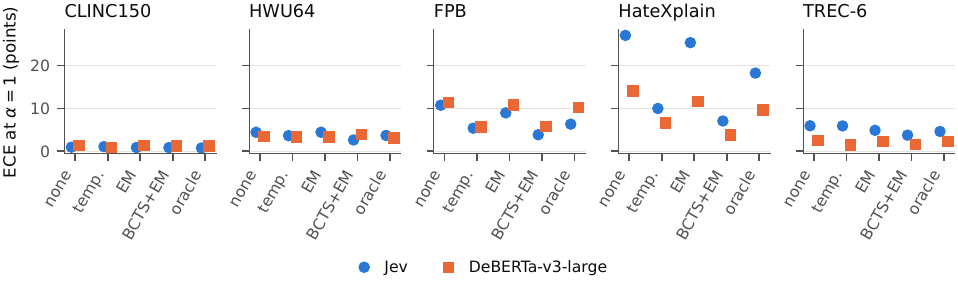}
  \caption{\textbf{Most of the calibration gain under shift comes from calibrating on the development set, not from
  correcting the prior (descriptive).} ECE under shift ($\alpha=1$, mean over 200 draws) for each recalibration arm.
  Label-free: EM, prior correction with the source prior (implicit for Jev, training prior for DeBERTa-v3-large).
  Labelled (fitted on the labelled development set): temp., temperature scaling; BCTS+EM, bias-corrected temperature
  scaling, then EM. Reference: oracle, true target prior (not deployable). Jev's temperature-scaled arm depends on the
  probability floor needed for its exact zeros (up to 3.1~points; \hyperref[app:results]{Appendix~\ref*{app:results}}). All systems in \hyperref[tab:exp2arms]{Table~\ref*{tab:exp2arms}}.}
  \label{fig:arms}
\end{figure}

\paragraph{Most of the calibration gain under shift comes from calibrating on the development set (descriptive).}
The arms fall into two categories that should not be compared as alternatives: EM with the implicit prior is
\emph{label-free} and is the only option when no labelled data are available, whereas temperature scaling and BCTS+EM
are \emph{labelled} recalibration fitted on the labelled development set (the oracle uses the true target prior and is
only a reference). \hyperref[fig:arms]{Figure~\ref*{fig:arms}} and \hyperref[tab:exp2arms]{Table~\ref*{tab:exp2arms}} compare all five arms. On FPB and HateXplain, where most systems are
miscalibrated, bias-corrected temperature scaling followed by EM gives the lowest ECE under shift for 13 of the 14
system--dataset pairs, for example 0.035 for Nimble-9B on FPB against 0.271 without recalibration; the exception is
DeBERTa-v3-large on FPB, where temperature scaling alone is 0.2~points lower. Its gain is mostly the correction of
miscalibration that exists without shift: relative to its own no-shift control, the ECE of BCTS+EM still rises under shift,
by 1.5 to 3.1~points for the zero-shot systems on HateXplain, more than the raw probabilities do there, but it stays far
below the unrecalibrated level. Temperature scaling alone transfers less well: its ECE increases under shift more than
that of the raw probabilities for 31 of the 35 system--dataset pairs.

\subsection{RQ4: Label budgets}
\label{sec:rq4}

All 280 budget trainings of DeBERTa-v3-large and TF-IDF + LR and all 45 QLoRA trainings completed without a failed run,
so no reserve seed was used. At full data the curves reproduce \hyperref[tab:exp1]{Table~\ref*{tab:exp1}}: averaged over
its five seeds, DeBERTa-v3-large is 0.4 (CLINC150) to 18.9 (HateXplain) Macro-F1~points above Jev. Every interval in this
section resamples both the training seeds and the test examples ($B=10{,}000$); the label-budget study estimates curves and tests no
hypothesis, so none of these results enters the Holm family.

\begin{figure}[t]
  \centering
  \includegraphics[width=\linewidth]{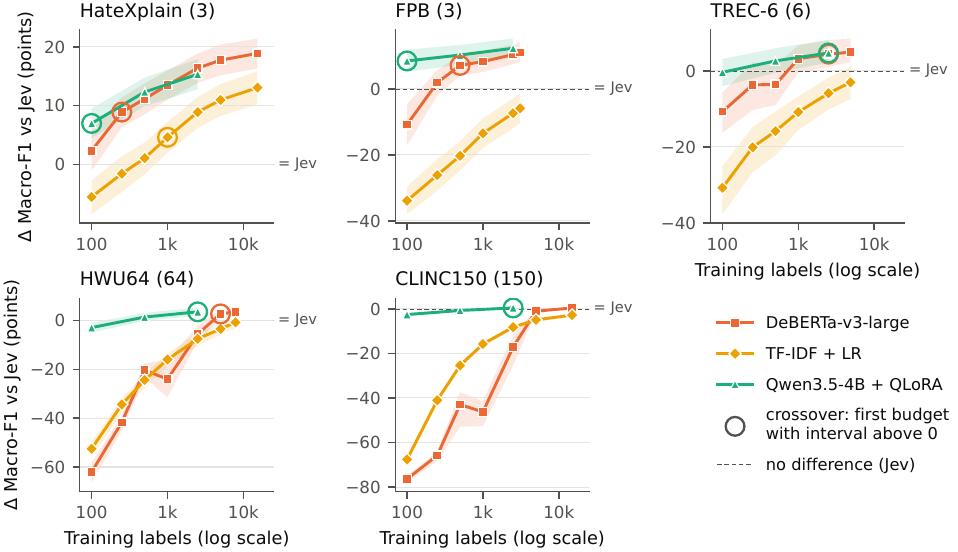}
  \caption{\textbf{The number of labels a trained model needs to overtake Jev ranges from at most 100 to more than the
  full training set, and rises with the number of classes in our five datasets.} Paired Macro-F1 difference (system $-$ Jev, percentage
  points) on the test set against the number of training labels (log scale; number of classes in brackets). Bands: 95\%
  bootstrap intervals over seeds and test examples (5 seeds; 3 for QLoRA; 1 for TF-IDF at full data). Rings: crossover
  budget, the smallest budget whose interval lies above zero (\hyperref[tab:crossover]{Table~\ref*{tab:crossover}}).
  The rightmost DeBERTa-v3-large and TF-IDF points are the full training set. All values in
  \hyperref[tab:budgetf1]{Table~\ref*{tab:budgetf1}}.}
  \label{fig:budget}
\end{figure}

\paragraph{How many labels a fine-tuned DeBERTa-v3-large needs to overtake Jev depends on the task.}
The crossover budget, the smallest budget whose interval lies above zero, is between 100 and 250 labels on HateXplain,
between 250 and 500 on FPB, between 1{,}000 and 2{,}500 on TREC-6 and between 2{,}500 and 5{,}000 on HWU64
(\hyperref[tab:crossover]{Table~\ref*{tab:crossover}}). On CLINC150 it is not reached: with the full 14{,}939 training
examples, about 100 per class, DeBERTa-v3-large is $+0.4$~points above Jev ($[-0.07, 0.8]$). In the per-class terms of
the few-shot intent literature, the HWU64 crossover lies between 9--44 and 18--89 examples per class, and on CLINC150
even about 100 examples per class are not enough, whereas the QLoRA model overtakes Jev there with at most 14--17
(\hyperref[tab:shots]{Table~\ref*{tab:shots}}). The crossover rises with
the number of classes, but the five datasets also differ in domain and text length, so this order is descriptive. Small
budgets are costly on the intent tasks: at 100 labels, which leave 50 of the 150 CLINC150 classes without a training
example (\hyperref[tab:shots]{Table~\ref*{tab:shots}}), DeBERTa-v3-large is 76.3~points below Jev on CLINC150 and
62.0 below it on HWU64.

\begin{table}[t]
  \centering
  \caption{\textbf{Of the three trained systems, a task QLoRA on Qwen3.5-4B overtakes Jev with the fewest labels;
  DeBERTa-v3-large does not clearly overtake it on CLINC150.} Crossover budget vs Jev (Macro-F1): the interval between the largest budget whose 95\% interval includes or
  lies below zero and the smallest budget whose interval lies above zero. The interpolation uses the point estimate and
  can therefore fall below the interval, e.g.\ for QLoRA on TREC-6, where the estimate at 500 labels is positive but its
  interval still includes zero. Interp.: budget at which the linearly
  interpolated point estimate first crosses zero (rough guide; \citealt{lescao2021howmany}). Budgets: 100, 250, 500,
  1{,}000, 2{,}500, 5{,}000 (HateXplain, HWU64, CLINC150 only) and full data; QLoRA: 100, 500, 2{,}500.}
  \label{tab:crossover}
  \scriptsize
  \setlength{\tabcolsep}{4pt}
  \begin{tabular}{lrcrcrcr}
\toprule
 & & \multicolumn{2}{c}{DeBERTa-v3-large} & \multicolumn{2}{c}{TF-IDF + LR} & \multicolumn{2}{c}{Qwen3.5-4B + QLoRA} \\
\cmidrule(lr){3-4} \cmidrule(lr){5-6} \cmidrule(lr){7-8}
Dataset & Classes & Crossover & Interp. & Crossover & Interp. & Crossover & Interp. \\
\midrule
HateXplain & 3 & (100, 250] & $\le$100 & (500, 1{,}000] & 399 & $\le$100 & $\le$100 \\
FPB & 3 & (250, 500] & 228 & not reached & -- & $\le$100 & $\le$100 \\
TREC-6 & 6 & (1{,}000, 2{,}500] & 765 & not reached & -- & (500, 2{,}500] & 150 \\
HWU64 & 64 & (2{,}500, 5{,}000] & 4{,}180 & not reached & -- & (500, 2{,}500] & 368 \\
CLINC150 & 150 & not reached & 12{,}400 & not reached & -- & (500, 2{,}500] & 1{,}820 \\
\bottomrule
\end{tabular}

\end{table}

\paragraph{Of the trained systems, a task-specific QLoRA on the 4B base overtakes Jev with the fewest labels.}
Qwen3.5-4B fine-tuned with QLoRA on 100 labels is above Jev on FPB ($+8.5$, $[5.2, 11.9]$) and HateXplain ($+7.0$,
$[4.4, 9.5]$). On TREC-6, HWU64 and CLINC150 the crossover lies between 500 and 2{,}500 labels; at 2{,}500 labels the
interval lies above zero on all five datasets, on CLINC150 only just ($+0.4$, $[0.03, 0.8]$) and not under in-budget
selection (see below). At the same budget the
QLoRA model is far ahead of DeBERTa-v3-large on the intent tasks: at 2{,}500 labels, $+0.4$ against $-17.2$ points on
CLINC150 and $+3.6$ against $-5.6$ on HWU64. At 100 labels it is 2.6 and 2.9~points below Jev on these two datasets,
against 76.3 and 62.0 for DeBERTa-v3-large.

\begin{figure}[t]
  \centering
  \includegraphics[width=\linewidth]{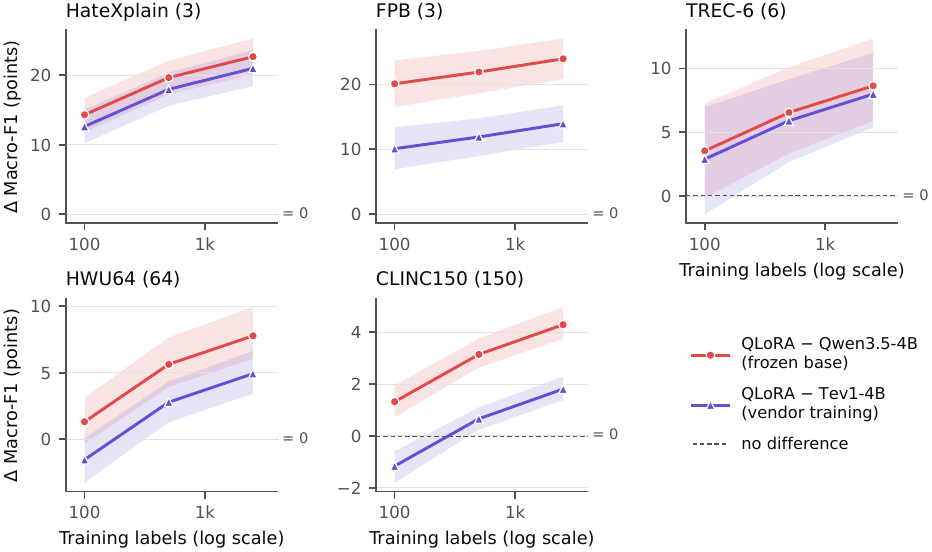}
  \caption{\textbf{With 500 task labels, a QLoRA adapter lifts Qwen3.5-4B above both its frozen version and Tev1-4B on
  every dataset (descriptive; on CLINC150 not resolved against Tev1-4B under in-budget selection).} Macro-F1 difference
  of Qwen3.5-4B + QLoRA to frozen Qwen3.5-4B and to Tev1-4B
  (percentage points) against the number of training labels (log scale), with 95\% bootstrap intervals over seeds (3)
  and test examples. Same prompt, assistant prefix and readout for all three (the Tev1 format); on FPB, part of the
  gain over the frozen base reflects that format, in which frozen Qwen3.5-4B scores 9.6~points below its generic-format
  level. Values in \hyperref[tab:qlorabase]{Table~\ref*{tab:qlorabase}}.}
  \label{fig:qlorabase}
\end{figure}

\paragraph{Task fine-tuning raises the 4B model more than vendor decision training (descriptive).}
Part of the gain over the frozen base on FPB reflects the Tev1 format rather than the task: frozen Qwen3.5-4B scores
9.6~points higher in generic formats (RQ1). The comparison with Tev1-4B, which uses the same format, is not affected.
\hyperref[fig:qlorabase]{Figure~\ref*{fig:qlorabase}} compares the three 4B conditions (frozen base, task QLoRA, Tev1-4B) with the same prompt and
readout (values in \hyperref[tab:qlorabase]{Table~\ref*{tab:qlorabase}}). With 500 labels, the QLoRA model is above frozen Qwen3.5-4B (by 3.1 to 21.9~points) and above Tev1-4B (by 0.7
to 18.0~points) on all five datasets, with every interval above zero. Under in-budget selection (see above) the
difference to Tev1-4B at 500 labels on CLINC150 is $+0.3$ ($[-0.1, 0.8]$) and no longer resolved; the other nine
intervals stay above zero. With 100 labels its point estimate is above
frozen Qwen3.5-4B on all five datasets, with the interval above zero on three (FPB, HateXplain, CLINC150); it is above
Tev1-4B on FPB and HateXplain, below it on CLINC150 ($-1.2$, $[-1.8,-0.6]$), and indistinguishable on TREC-6 and HWU64.
The two conditions answer different questions: the QLoRA model is trained on the target task, whereas Tev1's decision
training is generic, and its recipe is not published. The comparison shows how much a few hundred task labels add, not
why Tev1 gains what it does.

\paragraph{TF-IDF + LR reaches Jev only on hate speech.}
TF-IDF + LR overtakes Jev on HateXplain between 500 and 1{,}000 labels, where every zero-shot system is weak
(\hyperref[tab:exp1]{Table~\ref*{tab:exp1}}). On the other four datasets its interval never lies above zero, even with
the full training set, where its point estimate is 0.8 (HWU64) to 5.9 (FPB) points below Jev's.

\begin{figure}[t]
  \centering
  \includegraphics[width=\linewidth]{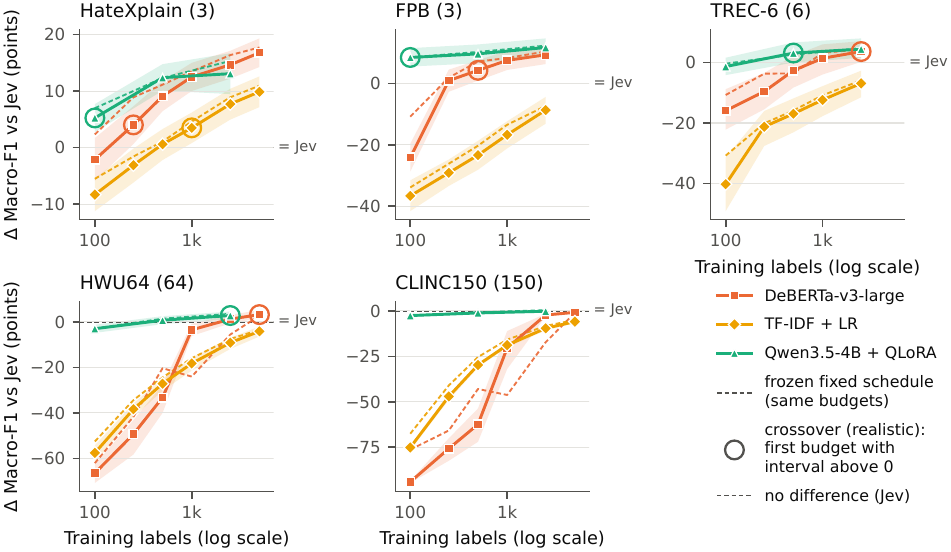}
  \caption{\textbf{With selection inside the budget, DeBERTa-v3-large and TF-IDF + LR overtake Jev at the same budgets
  as with the fixed schedule (exploratory).} Paired Macro-F1 difference (system $-$ Jev, percentage points) under the
  realistic protocol (20\% of each budget for selection and early stopping), with 95\% bootstrap intervals over seeds
  (5; 3 for QLoRA) and test examples; rings: crossover budget. Thin dashed lines: the fixed-schedule curves of
  \hyperref[fig:budget]{Figure~\ref*{fig:budget}} at the same budgets. The realistic protocol has no full-data
  budget.}
  \label{fig:budgetreal}
\end{figure}

\paragraph{With model selection inside the budget, the crossover intervals are unchanged for DeBERTa-v3-large and
TF-IDF + LR (exploratory).}
The budget curves above select hyperparameters and checkpoints on the full development set, with a fixed schedule. In
a rerun under a realistic protocol, 20\% of each budget is held out for selection, early stopping and the failed-run
rule, and the model trains on the other 80\% (\hyperref[app:budgets]{Appendix~\ref*{app:budgets}}). All ten crossover
intervals of DeBERTa-v3-large and TF-IDF + LR are the same as in \hyperref[tab:crossover]{Table~\ref*{tab:crossover}}
(\hyperref[fig:budgetreal]{Figure~\ref*{fig:budgetreal}}). The realistic models are weaker at the smallest budgets,
where they train on fewer labels (DeBERTa-v3-large on FPB at 100 labels: 51.0 against 64.3 Macro-F1; on CLINC150 every
seed at 100 labels fails and is scored as the majority class). On the intent datasets they are much stronger at
intermediate budgets (CLINC150 at 1{,}000 labels: 77.4 against 51.5; HWU64: 87.3 against 66.8), so the dips of the
fixed schedule do not recur. The rerun changes the selection and the schedule (step range and early stopping)
together, so it does not tell which of the two removes them. For QLoRA the crossover is unchanged on FPB, HateXplain
and HWU64. It comes earlier on TREC-6, but only marginally: between 100 and 500 labels ($+3.1$, $[0.006, 6.7]$ at
500). It is no longer reached on
CLINC150, where the model draws level with Jev at 2{,}500 labels ($-0.2$, $[-0.7, 0.3]$).

\paragraph{Calibration does not improve steadily with labels on the intent tasks (descriptive).}
On FPB the ECE difference to Jev falls from $+16.7$~points at 100 labels to $+2.2$ at full data, and on TREC-6 and
HateXplain DeBERTa-v3-large is better calibrated than Jev from 2{,}500 labels on
(\hyperref[fig:budgetece]{Figure~\ref*{fig:budgetece}} and \hyperref[tab:budgetece]{Table~\ref*{tab:budgetece}} in
\hyperref[app:budgets]{Appendix~\ref*{app:budgets}}). On CLINC150 and HWU64, by contrast, its ECE is highest at
intermediate budgets: on CLINC150 it is 49.6~points above Jev's at 2{,}500 labels and 0.6 at full data. These models are
underconfident: at 2{,}500 labels on CLINC150, their mean top-label probability is 0.317 at an accuracy of 0.820
(\hyperref[tab:debconf]{Table~\ref*{tab:debconf}}). A likely cause is the fixed schedule (20 epochs up to 500 labels, five above, at least 500
steps), which may stop too early for 64 or 150 classes; we return to this in
\hyperref[sec:limitations]{Section~\ref*{sec:limitations}}. The QLoRA model's ECE is within 1~point of Jev's at
2{,}500 labels on FPB, HWU64 and CLINC150, and its point estimate is lower on TREC-6 ($-1.7$) and HateXplain ($-11.4$,
interval $[-20.4, 0.3]$).

\section{Discussion}
\label{sec:discussion}

\paragraph{Is a typed decision model enough?}
Without task labels, Jev had the highest Macro-F1 point estimate among the zero-shot systems of the main comparison on
all five datasets, and
under moderate prevalence shift its calibration error and threshold risk were not distinguishable from those of
DeBERTa-v3-large after correction on four of the five. When the full training set is available, the fine-tuned
DeBERTa-v3-large still has the higher Macro-F1 point estimate everywhere, by 0.26~points on CLINC150 up to 18.9~points on
HateXplain. Whether a typed decision model is enough therefore depends on the task, on whether labels exist and on how
many. With our fixed training schedule, a fine-tuned DeBERTa-v3-large overtakes Jev with a few hundred labels on the
three-class tasks, needs 2{,}500--5{,}000 on HWU64 and is not clearly ahead on CLINC150 even with the full training set
(\hyperref[sec:rq4]{Section~\ref*{sec:rq4}}). For decisions within an intent group, then, a TDM is hard to beat with an
encoder fine-tuned on the task, but a QLoRA adapter on an open backbone overtakes Jev with at most 2{,}500 labels on
every dataset we tested, on CLINC150 only marginally and not under in-budget selection. These are five tasks, one backbone and three seeds for QLoRA; they show that a few thousand
labels can be enough, not how many a new task needs.

\paragraph{Why does the crossover rise with the number of classes?}
Two factors that our design cannot separate both point the same way. First, a fixed budget gives fewer examples per
class when there are more classes: 2{,}500 labels are on average 17 examples per class on CLINC150 but 833 on the
three-class tasks. Second, Jev leaves less room on the intent tasks, where its Macro-F1 is 0.978 and 0.908, than on
HateXplain (0.514), where it is weakest and the crossover comes earliest. The number of classes is also confounded with
domain and text length across our five datasets. The QLoRA result fits both factors: it starts from a model that
already reads the options zero-shot, so it does not have to learn each class from its few examples, and it overtakes
Jev earliest on the two datasets where Jev is least accurate. We offer this as a description of our five datasets, not
as a tested claim.

\paragraph{Why can decision training hurt?}
On FPB the two decision-trained models differ from their backbones in opposite directions: Tev1-4B's Macro-F1 is 10.0~points above Qwen3.5-4B's and Nimble-9B's is 12.0~points below Qwen3.5-9B's. The Tev1-4B gain, however, comes mostly from the base model's weak
score in the Tev1 format and disappears in generic formats (exploratory), so only the Nimble-9B result isolates an
effect of decision training. The Nimble release we test was trained on
2{,}676 synthetic examples, and FPB is labelled from an investor's point of view; one possible explanation is that the
synthetic data encode a different notion of positive and neutral financial news. The confusion matrices fit this
reading: the adapter mainly moves neutral sentences to the positive class (\hyperref[tab:fpbconf]{Table~\ref*{tab:fpbconf}}). We cannot test the
explanation itself without the vendors' training data. Both members of the pair run in the same 8-bit precision, so the comparison is matched, and
in bf16 the difference is nearly the same ($-11.7$~points; exploratory). What the paired design does establish is that,
for the same backbone, prompt and readout, decision training is not uniformly beneficial; a decision-trained variant
should be evaluated against its base model on the target task rather than assumed to be better, and the comparison
should also check that the base model is not handicapped by the TDM's prompt format.

\paragraph{Why does moderate shift matter so little?}
At $\alpha=1$ the target priors are centred on the training prior, so on average the shift is moderate, and for most
systems the calibration error under shift is dominated by the miscalibration they already have without shift; single
shifts, however, can be large (\hyperref[tab:excess]{Table~\ref*{tab:excess}}). Stronger shift ($\alpha=0.1$)
increases Jev's ECE more than DeBERTa-v3-large's on every dataset, but descriptively, and this is not a contrast between
zero-shot and trained systems: the trained TF-IDF classifier degrades more than Jev on two datasets, DeBERTa-v3-large is
the most stable system on three of the five, and on FPB and TREC-6 Tev1-4B degrades most of all systems (by 4.8 and 4.4~points, against 0.8 and 2.4 for Jev; \hyperref[tab:exp2delta]{Table~\ref*{tab:exp2delta}}). Two methodological points matter for such comparisons. First, resampling alone inflates
top-label ECE: on resampled data with no shift, ECE is 0.05 to 1.4~points higher than on the original test set (1.1~points for Jev on FPB, whose shift effect at $\alpha=1$ is 0.2~points), so shift effects must be measured against a
resampled control. Second, with only a few shift scenarios the result depends on which target priors are drawn
(\hyperref[sec:rq2]{Section~\ref*{sec:rq2}}), so conclusions should be stated for the shift distribution, not for a single draw.

\paragraph{Why does EM help most where the implicit prior is furthest from the truth?}
On FPB and HateXplain, Jev's implicit prior is far from the true class distribution (total-variation distance 0.18 and
0.23; \hyperref[tab:implicitprior]{Table~\ref*{tab:implicitprior}}), yet these are the datasets where EM with that prior helps most. This is less
paradoxical than it seems. The distance measures the model's label bias, not the quality of $\hat\pi_s$ as a reference
for EM. EM in \hyperref[eq:em]{Eq.~\ref*{eq:em}} rescales the posteriors by $\hat q/\hat\pi_s$, so what it needs is the prior with which
the model's own posteriors are consistent, which for a biased model is its own mean prediction and not the true prior.
Without shift, the mean prediction on the test set is close to $\hat\pi_s$ and EM leaves the posteriors almost
unchanged (\hyperref[tab:emnoshift]{Table~\ref*{tab:emnoshift}}); under shift, the change in the mean prediction is attributed to the shift, which
is what EM corrects. With the true prior in place of $\hat\pi_s$, EM would instead also try to remove the label bias,
which it cannot do reliably because the bias is not a change in class frequencies. Our data bear this out: with the
true class prior in place of the implicit one, EM raises the shifted ECE of every zero-shot system on FPB and
HateXplain above the uncorrected value (Jev on FPB: 0.172 against 0.107 without and 0.089 with the implicit prior;
exploratory, \hyperref[tab:emtrueprior]{Table~\ref*{tab:emtrueprior}}). EM then has the most to correct
where the shift moves the most probability mass between classes on which the model is calibrated differently. This is
the case for the three-class FPB and HateXplain, where Dirichlet draws at $\alpha=1$ often put most of the mass on one
class, and less so for the intent datasets, whose many classes and small ECE leave little to correct. We offer this as
an explanation consistent with our results, not as a tested claim.

\paragraph{What should a practitioner do?}
Four steps follow from our results. First, recalibrate on labelled data: bias-corrected temperature scaling followed by
EM, fitted on the labelled development sets, gave the lowest calibration error under shift for nearly every
system--dataset pair where calibration was poor. Second, without labels, EM with the model's implicit prior is a cheap
correction that lowered the shifted calibration error of every zero-shot system on FPB and HateXplain (descriptive;
tested only for Jev);
on the other datasets the intervals exclude harms larger than about half a point for Jev. Third, re-validate confidence
thresholds on target traffic: averaged over shifts, a single system's excess risk at 80\% coverage was at most 2.3~points at $\alpha=1$ (Nimble-9B on FPB), but individual shifts raised it by up to 35~points (Qwen3.5-4B on FPB) and by
up to 22~points for Jev on HateXplain (\hyperref[tab:excess]{Table~\ref*{tab:excess}}). Fourth, if labels can be
collected, test a task-tuned model against the TDM before choosing: in our five tasks, with 500 labels, a QLoRA adapter on
Qwen3.5-4B was above Jev on FPB and HateXplain, at most 0.8~points below it on the other three datasets (1.1 under
in-budget selection), and above both its frozen base and Tev1-4B on all five (on CLINC150 not resolved against Tev1-4B
under in-budget selection; \hyperref[fig:qlorabase]{Figure~\ref*{fig:qlorabase}}). How many labels a new task
needs is not something these five tasks can tell. On many-intent tasks, both comparisons assume that the right group
of intents is known. With a learned TF-IDF router, which misses the gold label's group on 6.2\% (CLINC150) and 7.0\%
(HWU64) of test examples, every system loses 5 to 6 accuracy points; on the missed examples, where the gold label is
not offered, the generative systems remain 70--84\% confident against 42--51\% for DeBERTa-v3-large (exploratory;
\hyperref[fig:router]{Figure~\ref*{fig:router}} in \hyperref[app:revision]{Appendix~\ref*{app:revision}}). A deployment
should therefore calibrate the routing step as well.

\paragraph{Relation to prior reports.}
Our FPB accuracy for Jev is consistent with \citet{deusser2026jev}, and so is our calibration pattern: like them, we
find Jev well calibrated where it is accurate (intent tasks) and miscalibrated where its accuracy is low (HateXplain). That decision training helps on some tasks and not others is consistent
with \citet{li2026llmasjev} and with the audit's conclusion that the typed readout itself has not shown an independent
accuracy advantage \citep{tang2026audit}; our paired design adds that the effect can be negative. The strength of BCTS
followed by EM matches \citet{alexandari2020maximum}; we observe it here also for zero-shot systems with an implicit
prior, but only descriptively, because our confirmatory test concerns plain EM.

\section{Limitations}
\label{sec:limitations}

\paragraph{Scope of systems and data.}
We study three TDMs, two open backbones and two trained classifiers; frontier generative APIs are out of scope. All
five datasets are English, public and human-labelled, and the TREC-6 and FPB test sets are small (500 and 970
examples), which widens their intervals. Public benchmarks may appear in the pretraining or decision-training data of
the models, which we can check only partly (\hyperref[app:runs]{Appendix~\ref*{app:runs}}); contamination would favour the zero-shot systems.

\paragraph{Task format.}
All results concern candidate-set decisions. On CLINC150 and HWU64 each example offers the 18--22 labels of a fixed label
group, not all 150 or 64 intents; the groups were chosen to prevent leakage of the gold label through the option list.
Full-label classification is harder for zero-shot systems, and other groupings could change the absolute numbers; we did
not test alternative groupings. Because the group is that of the gold label, every system, including the trained
classifiers and the QLoRA models, decides only within the correct group: our CLINC150 and HWU64 results, among them the
label budgets of RQ4, describe this second stage and not the choice of the group, which a deployment must make
separately and whose errors would add to those reported here. EM on candidate sets also relies on this design
(\hyperref[sec:setting]{Section~\ref*{sec:setting}}): it assumes that the gold label is always among the candidates. A
group selector that is a fixed function of the input would keep the label-shift invariance the derivation needs, but
whenever it excludes the gold label, the restricted posterior and the EM update give that label probability zero.

\paragraph{Shift model.}
We simulate prevalence shift by resampling test sets, so the label-shift assumption holds by construction. Real
deployments also change $p(x\mid y)$ (covariate or concept shift), under which prior correction can fail; our results
are therefore an optimistic case for prior correction. The confirmatory tests use one shift level ($\alpha=1$); the
stronger level is reported descriptively. A naturally or temporally shifted dataset, and alternative candidate
groupings, would test how far our conclusions transfer; both require new inference runs and are left to future work.

\paragraph{Design of the shift tests.}
The confirmatory PA1--PA3 estimand is conditional on one fixed set of 50 target priors, and its intervals reflect only
the finiteness of the test set. In hindsight this was a weakness of our design. Single priors move the Jev--DeBERTa
difference by several points on FPB and HateXplain (standard deviation 5.4 and 7.4~points across priors), so the frozen
set of 50 does not pin down the mean over the shift distribution: on FPB the confirmatory PA1 estimate is 1.7~points,
the average over 2{,}000 priors 0.4 (\hyperref[tab:papop]{Table~\ref*{tab:papop}}). The significant FPB results for PA1 and PA2 may therefore
reflect the particular prior set as much as a property of Jev. PA3 is less affected: across prior sets its estimate keeps
its sign and stays between 1.7 and 2.7~points on FPB. Future studies with a fixed protocol should target the mean over the shift
distribution, with the priors resampled inside the bootstrap or with many more priors, rather than a single prior set.

\paragraph{Models.}
We test one commercial TDM in one version, which limits how far the Jev results generalise. Jev's version cannot
be pinned (every call returned \texttt{jev-1.13.0}), and Jev is not deterministic: its probability
vectors differ between repeated calls on up to 81\% of examples, although its predictions rarely do. We run the two 9B
models in 8-bit precision to fit a 12\,GB GPU, whereas Nimble is distributed for bf16 inference; both members of the pair
share the precision, so the comparison is matched; a bf16 rerun of the FPB pair with CPU offload gives nearly the
same difference (exploratory), but the probabilities of 8-bit models can depend on the computation path. The
generic-format check covers FPB only; the smaller Tev1-4B gains on CLINC150 and HWU64 (2.5 and 2.8 Macro-F1~points)
were not re-scored outside the Tev1 format and may also depend on it. The vendors' training
data are not public, so we can show where decision training helps or hurts (for Nimble-9B on FPB, on the neutral class)
but not why. Reproducing P1 depends on continued access to the Tev1-4B weights, whose licence was still being
finalised, and the QLoRA adapters are not released, so the QLoRA results can be reproduced only by retraining from the
released code, configuration and seeds.

\paragraph{Label budgets.}
The training schedule was fixed before the runs and is not tuned per budget. Because it switches from 20 to 5 epochs
above 500 labels, DeBERTa-v3-large gets 500 optimizer steps at 1{,}000 labels but 640 at 500, and on HWU64 and CLINC150
its Macro-F1 at 1{,}000 labels is below that at 500 (\hyperref[tab:budgetf1]{Table~\ref*{tab:budgetf1}}), with seeds
that vary widely, and its models at 1{,}000 and 2{,}500 labels are underconfident
(\hyperref[tab:debconf]{Table~\ref*{tab:debconf}}). On these two datasets the schedule probably under-trains the model at
intermediate budgets, so their crossover budgets may be overstated; we did not change the schedule after seeing the
results. A post-hoc rerun with selection inside the budget (exploratory, \hyperref[sec:rq4]{Section~\ref*{sec:rq4}})
removes these dips and gives the same DeBERTa-v3-large crossover intervals; because it changes the selection and the
schedule together, it does not separate their effects. Every budget selects its hyperparameters
and checkpoint on the full development set, which is optimistic for the trained models at small budgets
\citep{perez2021true}; in the same rerun the conclusions hold, except that the QLoRA model draws level with Jev on
CLINC150 instead of overtaking it. The QLoRA condition uses one backbone, three budgets and three
seeds, and its crossover intervals are correspondingly coarse. Its adapters were trained and selected on a 4-bit NF4 base
but are evaluated on the bf16 base. On the development set, the deployed model's Macro-F1 differs from that of the kept
4-bit checkpoint by $-1.3$ to $+2.5$~points (mean $+0.3$) over the 45 runs; this compares precision and batched against
unbatched scoring together, on the set used for checkpoint selection, so it does not isolate the effect of the precision
difference on the test results. Under in-budget selection, where the selection part has as few as 20 examples, the
difference ranges from $-12.7$ to $+13.5$~points at 100 labels and from $-4.6$ to $+0.9$ at 500 and 2{,}500, so a
checkpoint chosen in 4-bit on so few examples can rank differently in bf16; on CLINC150 at 2{,}500 labels, where the
in-budget model draws level with Jev, it is $-0.6$ to $+0.1$, so that result does not come from this gap. The number of classes is confounded with the task, so the
rise of the crossover with the number of classes is descriptive.

\paragraph{Calibration metric.}
Our calibration conclusions concern top-label calibration: ECE with 15 equal-width bins on the confidence of the
predicted label. We did not evaluate class-wise calibration, calibration of the individual candidate probabilities or
alternative binning schemes; Brier score and NLL, which use the whole probability vector, are reported
(\hyperref[app:results]{Appendix~\ref*{app:results}}) but not tested.

\paragraph{Statistics.}
Bootstrap intervals hold quantities fitted on the development set fixed, and so understate total uncertainty; they are
not adjusted for multiplicity, whereas the reported significance is. They also omit two further sources of variation.
Jev's outputs differ between repeated calls, and its ECE varied by up to 1.2~points between runs on 500-example subsets
(\hyperref[tab:jevrepeat]{Table~\ref*{tab:jevrepeat}}), so differences of about one point or less involving Jev are weak
evidence even when their interval excludes zero. And the confirmatory comparisons with DeBERTa-v3-large use one training
seed (1001), so their intervals exclude seed variation. Repeating the descriptive shift experiment for all five
full-data seeds (exploratory; \hyperref[tab:debseedshift]{Table~\ref*{tab:debseedshift}}), the Jev--DeBERTa difference
at $\alpha=1$ keeps its direction for every seed on FPB, TREC-6 and HateXplain, where it is largest, but its size
depends on the seed, most for DeBERTa's change in ECE on HateXplain ($-0.3$ to $+1.7$~points) and its excess risk on FPB
(0.0 to 0.3~points); the confirmatory tests were not repeated for the other seeds. Absence of a significant difference is not evidence of equivalence beyond the width of the
reported intervals.

\section{Conclusion}
\label{sec:conclusion}

We compared typed decision models with their own backbones and with trained classifiers on identical items and
candidate sets, under controlled prevalence shift and with 35 tests frozen before the confirmatory analysis (earlier
pipeline checks on test data are disclosed). Vendor decision
training changed the same backbone in a task-dependent direction: with only an adapter between them, Nimble-9B was better
calibrated than Qwen3.5-9B on intent tasks and 12 Macro-F1~points worse on financial sentiment (RQ1). Under moderate
prevalence shift, on the pre-specified set of 50 target priors, Jev's top-label calibration error and threshold risk
grew significantly more than those of a fine-tuned DeBERTa only on financial sentiment (by 1.7 and 2.5~points), the
dataset whose test split was exposed in pipeline checks; this difference depended on the shift scenarios, and averaged
over 2{,}000 scenarios it was at most one point on every dataset (RQ2). A label-free EM correction with the model's implicit
prior lowered Jev's calibration error under shift by 2.7 and 1.8~points on the two datasets where it was highest,
consistently in sign across scenarios, but recalibration on labelled development data reduced it much further (RQ3).
The number of labels a fine-tuned DeBERTa-v3-large needed to overtake Jev ranged from 100--250 on hate speech to more
than the full training set on CLINC150, and a QLoRA adapter on Qwen3.5-4B overtook Jev with at most 2{,}500 labels on
every dataset and with at most 100 on two, on CLINC150 only marginally and not under in-budget selection (RQ4); a
rerun with selection inside the budget gave the same DeBERTa-v3-large crossovers. On the two intent datasets all of these
results concern decisions within the group of 18--22 intents that contains the gold label, not the choice of that group.
The three typed decision models we tested are neither uniformly better than their backbones nor uniquely fragile under
prevalence shift; for top-label calibration, their main weakness is the miscalibration they already have without shift. Whether
one is enough for a task is best decided by testing it against its base model and against a model tuned on a few hundred
task labels, and by recalibrating it on labelled data from the deployment setting.

\section*{Broader impact statement}
Typed decision models are built to make automated decisions inside software, for example routing, moderation and
triage. Our results show that their calibration and the error rate of confidence thresholds can change when the
deployment class distribution differs from the evaluation data, and that decision training can make a model worse on a
task. Deploying such models in consequential settings (moderation, finance, health) without validation on target data
can therefore cause harm; we recommend task-specific evaluation against a base model, recalibration on labelled data,
and monitoring of threshold error rates. HateXplain contains offensive text; we used it only for evaluation.

\section*{Reproducibility statement}
All datasets are public. Raw files, processed splits, label texts and candidate sets are pinned by SHA-256 hashes; the
frozen protocol, configurations and code are recorded in hash files, and every result was verified against them. Model
revisions are listed in \hyperref[tab:systems]{Table~\ref*{tab:systems}}; seeds are fixed throughout. We release the code, the protocol and
deviation log, and every per-example probability vector at
\url{https://github.com/AbdulRahmanIqbal/tdm-prevalence-shift}. The label-budget study has its own hash file, written
before its test predictions; its training records and predictions are released in the same repository, and the QLoRA
adapters, which are not released, are identified by their SHA-256 hashes. Jev is a hosted model whose version
cannot be pinned and whose outputs are not deterministic (\hyperref[app:results]{Appendix~\ref*{app:results}}).

\bibliography{refs}
\bibliographystyle{tmlr}

\appendix
\makeatletter
\setlength{\@fptop}{0pt}
\setlength{\@fpsep}{14pt plus 4pt}
\setlength{\@fpbot}{0pt plus 1fil}
\makeatother
\renewcommand{\topfraction}{0.95}
\renewcommand{\bottomfraction}{0.9}
\renewcommand{\textfraction}{0.05}
\renewcommand{\floatpagefraction}{0.85}
\setcounter{topnumber}{5}
\setcounter{bottomnumber}{3}
\setcounter{totalnumber}{8}
\section{Deviations from the frozen protocol}
\label{app:deviations}

The protocol (hypotheses, the 35-test family, configurations, data manifests and code) was frozen on 2026-10-08 at
07:45 UTC and recorded as SHA-256 hashes. While the local decoder runs were in progress, three code audits
were carried out; their findings led to one amendment, frozen at 10:58 UTC the same day. By then, test predictions of
Qwen3.5-4B and Tev1-4B had been produced, and the run log had printed their test accuracies. No test of the
35-test family, no shift analysis and no DeBERTa or Jev run had been carried out on the final runs (the pipeline checks
before the freeze are disclosed below). The amendment did not change the code
used by the local decoder models, so all decoder runs, before and after it, use identical code; the DeBERTa and Jev
runs and every analysis use the amended version.
\paragraph{Chronology (UTC, from the run records).}
\begin{itemize}
  \item 2026-10-06: label texts frozen. 2026-10-07: data re-processed after a cleaning fix; candidate sets rebuilt as
  fixed groups after the leakage audit; pipeline checks, some on test splits (see the disclosure below).
  \item 2026-10-08 07:45: protocol v1.1 frozen (hash file).
  \item 2026-10-08 09:14--10:45: all Qwen3.5-4B and Tev1-4B predictions (development and test); 10:45: first
  Qwen3.5-9B run started.
  \item 2026-10-08 10:58: amendment v1.1.1 frozen; no test of the 35-test family and no shift analysis had been run
  on the final runs.
  \item 2026-10-08 10:45--16:56: remaining local decoder predictions; latency timing completed on 2026-10-09.
  2026-10-09 to 10-10: DeBERTa-v3-large. 2026-10-13 05:23: TF-IDF.
  \item 2026-10-13 05:42: no-shift tables and P1/P2 for all systems except Jev; 05:47--07:50: all Jev runs;
  07:50--08:19: Jev's no-shift table and the descriptive shift experiment; 2026-10-13/14: PA1--PA3; 2026-10-14: Holm
  over all 35 tests.
  \item 2026-10-14, after all confirmatory results: prior-set sensitivity of PA1--PA3, HWU64 without the MASSIVE
  overlap, descriptive change in ECE and recalibration arms for all systems.
  \item 2026-10-15: distance of the implicit priors from the true class distribution, ECE range of the repeated Jev
  runs, excess-risk, Brier and NLL tables, TF-IDF grid extension to $10^5$ and $10^6$, average of PA1--PA3 over 2{,}000 priors, EM without
  shift, FPB confusion matrices, rank correlations between metrics. All are listed below and none changes a
  confirmatory result.
  \item 2026-11-06: EM with the true class prior instead of the implicit prior
  (\hyperref[tab:emtrueprior]{Table~\ref*{tab:emtrueprior}}), on the stored predictions with no new model call; rerun
  with an identical result on 2026-11-17.
  \item 2026-11-10 09:38: revision protocol v1.3 frozen (hash file) after reviews of the paper draft and after all
  earlier results were known; it adds five exploratory analyses with their own run identifiers and replaces no frozen
  result. 2026-11-10: content-free probes and routing for the local systems and the retrained classifiers; generic
  prompt formats on FPB. 2026-11-11: bf16 rerun of the Nimble-9B pair on FPB. 2026-11-11 to 11-12: in-budget rerun of
  DeBERTa-v3-large and TF-IDF + LR. 2026-11-12 to 11-16: in-budget rerun of QLoRA. 2026-11-17 from 07:59: the Jev
  calls of the content-free probes (1{,}752) and the misrouted examples (354), all returning \texttt{jev-1.13.0}, the
  version of the main runs; then all v1.3 analyses.
\end{itemize}

\paragraph{Amendment v1.1.1.}
\begin{itemize}
  \item \textbf{No-shift control.} The shift effect on ECE (the descriptive shift experiment and PA1) is measured against a no-shift control
  drawn with the same with-replacement resampler at the empirical class prior. The original protocol compared against
  the unresampled test set, which would have attributed resampling bias in top-label ECE to the shift.
  \item \textbf{Bootstrap size for PA1--PA3.} $B$ was raised from 1{,}000 to 10{,}000. With $B=1{,}000$ the smallest
  non-zero two-sided $p$ ($2/B=0.002$) is above Holm's first threshold ($0.05/35\approx0.0014$).
  \item \textbf{Failed-run rule for DeBERTa.} The majority-class threshold is computed within each example's candidate
  set, on the same scale as the model's accuracy (previously the two were on different scales for CLINC150 and HWU64).
  No full-data run failed under either rule.
  \item \textbf{Jev transport errors.} Network errors are never classified as refusals. No Jev request failed or was
  refused in the final runs.
  \item \textbf{No-shift metrics.} The specified treatment of refused requests (counted as wrong and abstained) was
  implemented; no request was refused. The confirmatory scope of H3 was narrowed in wording to plain EM on Jev, and
  the AURC definition was stated exactly.
\end{itemize}
\textbf{Analyses added after the results were known.} The following analyses were added after the confirmatory results
had been computed, in response to reviews of the draft; all are labelled exploratory or descriptive and none
changes a confirmatory result: the prior-set sensitivity of PA1--PA3 (\hyperref[tab:pasens]{Table~\ref*{tab:pasens}}) and their average over
2{,}000 target priors (\hyperref[tab:papop]{Table~\ref*{tab:papop}}); the HWU64 analysis without the MASSIVE overlap (required by the protocol
but run only at this stage; \hyperref[tab:hwu64]{Table~\ref*{tab:hwu64}}); the descriptive shift-experiment summaries for all systems
(\hyperref[tab:exp2delta]{Tables~\ref*{tab:exp2delta}}--\nobreak\ref{tab:nll}); the distance of each implicit prior from the true class distribution
(\hyperref[tab:implicitprior]{Table~\ref*{tab:implicitprior}}); the effect of EM without shift (\hyperref[tab:emnoshift]{Table~\ref*{tab:emnoshift}}); the FPB confusion matrices
of the Nimble pair (\hyperref[tab:fpbconf]{Table~\ref*{tab:fpbconf}}); the ECE range of the repeated Jev runs
(\hyperref[tab:jevrepeat]{Table~\ref*{tab:jevrepeat}}); the rank correlations between metrics
(\hyperref[tab:exp1full]{Table~\ref*{tab:exp1full}}); a check that the PA estimates are not an artefact of
resampling twice (PA1--PA3 recomputed without the bootstrap base on FPB, HateXplain and TREC-6; the estimates agree);
the extension of the TF-IDF regularisation grid to $10^5$ and $10^6$
(\hyperref[sec:protocol]{Section~\ref*{sec:protocol}}); a simulation of EM on candidate sets and with distorted
posteriors (\hyperref[tab:emsim]{Tables~\ref*{tab:emsim}} and~\ref{tab:emmiscal}); the shift experiment repeated
for all five DeBERTa-v3-large seeds (\hyperref[tab:debseedshift]{Table~\ref*{tab:debseedshift}}); EM with the true
class prior instead of the implicit prior (\hyperref[tab:emtrueprior]{Table~\ref*{tab:emtrueprior}}); and the five
analyses of revision protocol v1.3 listed in the chronology (in-budget selection, routing, content-free calibration,
generic prompt formats and bf16).

\textbf{Disclosure.} During pipeline development, before the freeze, checks of the code were run on \emph{test}
splits: Jev on the full FPB test split and on 5 to 10 test examples of CLINC150, FPB and HateXplain; Qwen3.5-4B and
Tev1-4B on 5 FPB test examples each; a DeBERTa-v3-large model trained on 100 examples, on the FPB test split; TF-IDF
on all five test splits in three pipeline runs, whose test scores were printed; and early versions of PA1--PA3 and
the descriptive shift experiment on FPB. Two design changes followed these checks: fixed target priors for PA1--PA3,
because of the width of the intervals under the earlier design, and the extension of the TF-IDF grid from $10^3$ to
$10^4$, because the value selected on the development set lay at the edge of the grid. Neither change was based on
the direction or size of a test estimate. From then on, pipeline checks used only the development split, and all reported test results come from the
frozen protocol.

\paragraph{Label-budget study (second freeze).}
The budgets, seeds, DeBERTa-v3-large and TF-IDF settings, failed-run rule and crossover definition were part of the
protocol frozen on 2026-10-08. The QLoRA settings that the protocol left open were fixed on 2026-10-16, before any
label-budget result: the DeBERTa schedule rule, LoRA dropout 0.05 and the target matrices of the Nimble-9B adapter,
training on a 4-bit base with every reported prediction from the bf16 base, and the failed-run rule applied to that
deployed model. A pre-specified development-only pilot (CLINC150 and FPB, budget 100, one seed) decided whether
checkpoints should be evaluated more often than every 100 steps; the best development Macro-F1 before step 100 was not
higher than the best on the 100-step grid (differences 0.0000 and $-0.0006$; threshold 0.01), so the spacing stayed
at 100. A pre-freeze code review led to
fixes that change no setting (adapters stored per dataset and never overwritten, unrounded failed-run comparison, guards
against repeated or incomplete runs, a curve analysis that refuses incomplete curves). The second freeze, with all
code and configurations hashed, was recorded on 2026-10-21 at 13:42 UTC. QLoRA runs followed from 2026-10-22 to
2026-11-02, and the DeBERTa-v3-large and TF-IDF budget runs on 2026-11-02 and 2026-11-03; the curves were computed
once all runs had been verified against the second freeze. No run failed, and no setting was changed after a test
prediction. The disclosure above also applies here: before the first freeze, a DeBERTa-v3-large model trained on 100
examples had been evaluated on the FPB test split during a pipeline check. The adapters were moved out of the results
folder after each dataset and checked against their recorded hashes.

\FloatBarrier
\section{Additional results}
\label{app:results}

\begin{table}[htbp]
  \centering
  \caption{\textbf{Fourteen of the 35 confirmatory tests are significant after Holm correction.} Paired differences in percentage points (TDM $-$ base for P1/P2; Jev $-$
  DeBERTa-v3-large for PA1/PA2; EM $-$ none for PA3) with unadjusted 95\% bootstrap intervals, raw two-sided $p$ and
  Holm-adjusted $p$ over the 35 tests. Shaded: Holm-adjusted $p<0.05$.}
  \label{tab:alltests}
  \scriptsize
  \begin{tabular}{llrcrr}
\toprule
Test & Dataset & $\Delta$ (points) & 95\% CI & $p$ & Holm $p$ \\
\midrule
\rowcolor{rowhl} P1-F1 & CLINC150 & +2.48 & [+1.94, +3.10] & $<$0.0002 & $<$0.007 \\
\rowcolor{rowhl} P1-F1 & HWU64 & +2.84 & [+1.10, +4.78] & 0.002 & 0.040 \\
\rowcolor{rowhl} P1-F1 & FPB & +9.99 & [+7.35, +12.69] & $<$0.0002 & $<$0.007 \\
P1-F1 & HateXplain & +1.69 & [$-$0.16, +3.52] & 0.077 & 1.000 \\
P1-F1 & TREC-6 & +0.65 & [$-$2.92, +4.44] & 0.724 & 1.000 \\
\midrule
P1-ECE & CLINC150 & $-$0.34 & [$-$0.84, +0.21] & 0.255 & 1.000 \\
P1-ECE & HWU64 & $-$1.87 & [$-$3.29, $-$0.19] & 0.028 & 0.479 \\
\rowcolor{rowhl} P1-ECE & FPB & $-$13.19 & [$-$15.75, $-$10.39] & $<$0.0002 & $<$0.007 \\
\rowcolor{rowhl} P1-ECE & HateXplain & $-$3.23 & [$-$5.19, $-$1.37] & 0.001 & 0.029 \\
P1-ECE & TREC-6 & +0.12 & [$-$2.97, +2.37] & 0.796 & 1.000 \\
\midrule
\rowcolor{rowhl} P2-F1 & CLINC150 & +0.69 & [+0.23, +1.17] & 0.001 & 0.029 \\
P2-F1 & HWU64 & +1.14 & [$-$0.64, +3.03] & 0.195 & 1.000 \\
\rowcolor{rowhl} P2-F1 & FPB & $-$11.97 & [$-$14.61, $-$9.28] & $<$0.0002 & $<$0.007 \\
P2-F1 & HateXplain & $-$1.66 & [$-$3.23, $-$0.07] & 0.039 & 0.630 \\
P2-F1 & TREC-6 & $-$3.11 & [$-$6.99, +0.69] & 0.100 & 1.000 \\
\midrule
\rowcolor{rowhl} P2-ECE & CLINC150 & $-$1.22 & [$-$1.52, $-$0.63] & $<$0.0002 & $<$0.007 \\
\rowcolor{rowhl} P2-ECE & HWU64 & $-$4.22 & [$-$5.59, $-$2.50] & $<$0.0002 & $<$0.007 \\
\rowcolor{rowhl} P2-ECE & FPB & +9.56 & [+6.77, +12.05] & $<$0.0002 & $<$0.007 \\
P2-ECE & HateXplain & $-$0.36 & [$-$1.94, +1.21] & 0.636 & 1.000 \\
P2-ECE & TREC-6 & $-$1.84 & [$-$3.67, +1.63] & 0.440 & 1.000 \\
\midrule
PA1 & CLINC150 & +0.09 & [$-$0.02, +0.20] & 0.104 & 1.000 \\
PA1 & HWU64 & +0.11 & [$-$0.30, +0.50] & 0.592 & 1.000 \\
\rowcolor{rowhl} PA1 & FPB & +1.71 & [+0.98, +2.49] & $<$0.0002 & $<$0.007 \\
PA1 & HateXplain & $-$0.78 & [$-$1.29, $-$0.26] & 0.004 & 0.080 \\
PA1 & TREC-6 & +1.01 & [+0.26, +1.85] & 0.007 & 0.136 \\
\midrule
PA2 & CLINC150 & 0.00 & [$-$0.04, +0.04] & 0.938 & 1.000 \\
PA2 & HWU64 & +0.01 & [$-$0.21, +0.23] & 0.969 & 1.000 \\
\rowcolor{rowhl} PA2 & FPB & +2.49 & [+1.83, +3.16] & $<$0.0002 & $<$0.007 \\
PA2 & HateXplain & $-$0.55 & [$-$0.98, $-$0.14] & 0.010 & 0.187 \\
PA2 & TREC-6 & +0.55 & [$-$0.06, +1.30] & 0.086 & 1.000 \\
\midrule
PA3 & CLINC150 & $-$0.12 & [$-$0.29, +0.03] & 0.122 & 1.000 \\
PA3 & HWU64 & +0.12 & [$-$0.33, +0.54] & 0.546 & 1.000 \\
\rowcolor{rowhl} PA3 & FPB & $-$2.65 & [$-$3.74, $-$1.61] & $<$0.0002 & $<$0.007 \\
\rowcolor{rowhl} PA3 & HateXplain & $-$1.75 & [$-$2.71, $-$0.82] & $<$0.0002 & $<$0.007 \\
PA3 & TREC-6 & $-$1.06 & [$-$1.96, $-$0.28] & 0.007 & 0.136 \\
\bottomrule
\end{tabular}

\end{table}

\begin{table}[htbp]
  \centering
  \caption{\textbf{Accuracy and MCC rank the seven systems of the main analysis almost as Macro-F1 does (Spearman
  $\rho\ge0.75$ on every dataset; descriptive), whereas NLL, which penalises confident errors, ranks them differently on
  HWU64 and HateXplain ($\rho=0.29$ and $0.32$).} Test-set metrics, no shift, for the seven systems of the main analysis
  and, in an eighth row, an exploratory system. $^\ast$Exploratory: a later Nimble release whose training data include
  public TREC questions, which may explain its TREC-6 score; it is not one of the seven systems and is not part of any
  confirmatory test or of the rank correlations. Acc@80: accuracy on the 80\% most confident examples.}
  \label{tab:exp1full}
  \scriptsize
  \setlength{\tabcolsep}{3pt}
  \begin{tabular}{llrrrrrrrr}
\toprule
Dataset & System & Acc$\uparrow$ & F1$\uparrow$ & MCC$\uparrow$ & ECE$\downarrow$ & Brier$\downarrow$ & NLL$\downarrow$ & AURC$\downarrow$ & Acc@80$\uparrow$ \\
\midrule
CLINC150 & Jev & 0.978 & 0.978 & 0.978 & 0.007 & 0.032 & 0.152 & 0.004 & 0.997 \\
 & Tev1-4B & 0.964 & 0.964 & 0.963 & 0.008 & 0.055 & 0.130 & 0.003 & 0.997 \\
 & Nimble-9B & 0.962 & 0.962 & 0.961 & 0.008 & 0.060 & 0.140 & 0.004 & 0.996 \\
 & Nimble-9B (main)$^\ast$ & 0.969 & 0.969 & 0.968 & 0.010 & 0.050 & 0.123 & 0.002 & 0.997 \\
 & Qwen3.5-4B & 0.939 & 0.939 & 0.939 & 0.011 & 0.089 & 0.233 & 0.007 & 0.994 \\
 & Qwen3.5-9B & 0.955 & 0.955 & 0.955 & 0.020 & 0.067 & 0.176 & 0.005 & 0.995 \\
 & DeBERTa-v3-large & 0.980 & 0.980 & 0.980 & 0.013 & 0.031 & 0.105 & 0.002 & 0.997 \\
 & TF-IDF + LR & 0.950 & 0.950 & 0.949 & 0.015 & 0.077 & 0.189 & 0.005 & 0.993 \\
\midrule
HWU64 & Jev & 0.911 & 0.908 & 0.910 & 0.036 & 0.138 & 0.640 & 0.024 & 0.976 \\
 & Tev1-4B & 0.891 & 0.895 & 0.890 & 0.040 & 0.168 & 0.423 & 0.021 & 0.962 \\
 & Nimble-9B & 0.897 & 0.896 & 0.896 & 0.035 & 0.165 & 0.436 & 0.026 & 0.956 \\
 & Nimble-9B (main)$^\ast$ & 0.896 & 0.896 & 0.895 & 0.042 & 0.156 & 0.360 & 0.017 & 0.974 \\
 & Qwen3.5-4B & 0.862 & 0.866 & 0.860 & 0.058 & 0.216 & 0.596 & 0.029 & 0.944 \\
 & Qwen3.5-9B & 0.880 & 0.884 & 0.879 & 0.078 & 0.200 & 0.616 & 0.028 & 0.956 \\
 & DeBERTa-v3-large & 0.943 & 0.939 & 0.942 & 0.030 & 0.094 & 0.269 & 0.009 & 0.984 \\
 & TF-IDF + LR & 0.901 & 0.900 & 0.900 & 0.098 & 0.177 & 0.444 & 0.019 & 0.966 \\
\midrule
FPB & Jev & 0.744 & 0.752 & 0.608 & 0.095 & 0.373 & 0.745 & 0.120 & 0.795 \\
 & Tev1-4B & 0.715 & 0.735 & 0.597 & 0.072 & 0.378 & 0.582 & 0.141 & 0.751 \\
 & Nimble-9B & 0.608 & 0.629 & 0.508 & 0.263 & 0.612 & 1.138 & 0.205 & 0.644 \\
 & Nimble-9B (main)$^\ast$ & 0.652 & 0.683 & 0.542 & 0.167 & 0.509 & 0.849 & 0.194 & 0.688 \\
 & Qwen3.5-4B & 0.597 & 0.635 & 0.499 & 0.204 & 0.585 & 1.107 & 0.252 & 0.640 \\
 & Qwen3.5-9B & 0.743 & 0.749 & 0.607 & 0.168 & 0.403 & 0.821 & 0.115 & 0.818 \\
 & DeBERTa-v3-large & 0.875 & 0.870 & 0.776 & 0.111 & 0.234 & 0.634 & 0.045 & 0.920 \\
 & TF-IDF + LR & 0.753 & 0.693 & 0.538 & 0.167 & 0.401 & 0.990 & 0.123 & 0.817 \\
\midrule
HateXplain & Jev & 0.535 & 0.514 & 0.348 & 0.279 & 0.704 & 2.992 & 0.330 & 0.574 \\
 & Tev1-4B & 0.464 & 0.458 & 0.235 & 0.214 & 0.702 & 1.297 & 0.435 & 0.501 \\
 & Nimble-9B & 0.490 & 0.470 & 0.284 & 0.336 & 0.800 & 1.735 & 0.416 & 0.521 \\
 & Nimble-9B (main)$^\ast$ & 0.504 & 0.495 & 0.286 & 0.266 & 0.721 & 1.428 & 0.397 & 0.532 \\
 & Qwen3.5-4B & 0.460 & 0.441 & 0.237 & 0.247 & 0.740 & 1.415 & 0.457 & 0.493 \\
 & Qwen3.5-9B & 0.501 & 0.487 & 0.289 & 0.340 & 0.803 & 1.764 & 0.409 & 0.522 \\
 & DeBERTa-v3-large & 0.710 & 0.703 & 0.563 & 0.141 & 0.433 & 0.769 & 0.161 & 0.774 \\
 & TF-IDF + LR & 0.663 & 0.645 & 0.486 & 0.069 & 0.461 & 0.814 & 0.200 & 0.720 \\
\midrule
TREC-6 & Jev & 0.930 & 0.922 & 0.912 & 0.042 & 0.116 & 0.342 & 0.016 & 0.970 \\
 & Tev1-4B & 0.884 & 0.889 & 0.858 & 0.038 & 0.190 & 0.361 & 0.042 & 0.927 \\
 & Nimble-9B & 0.848 & 0.872 & 0.814 & 0.043 & 0.207 & 0.370 & 0.037 & 0.925 \\
 & Nimble-9B (main)$^\ast$ & 0.968 & 0.963 & 0.960 & 0.017 & 0.056 & 0.129 & 0.006 & 0.990 \\
 & Qwen3.5-4B & 0.880 & 0.882 & 0.853 & 0.037 & 0.177 & 0.363 & 0.029 & 0.950 \\
 & Qwen3.5-9B & 0.888 & 0.903 & 0.862 & 0.061 & 0.174 & 0.355 & 0.026 & 0.960 \\
 & DeBERTa-v3-large & 0.976 & 0.971 & 0.970 & 0.021 & 0.041 & 0.153 & 0.010 & 0.990 \\
 & TF-IDF + LR & 0.894 & 0.891 & 0.868 & 0.071 & 0.176 & 0.449 & 0.022 & 0.973 \\
\bottomrule
\end{tabular}

\end{table}

\begin{table}[htbp]
  \centering
  \caption{\textbf{DeBERTa-v3-large is stable across seeds.} Test Macro-F1 of the five full-data seeds; all analyses use
  seed 1001.}
  \label{tab:seeds}
  \scriptsize
  \begin{tabular}{lrrrrrr}
\toprule
Dataset & seed 1001 & 1002 & 1003 & 1004 & 1005 & mean $\pm$ SD \\
\midrule
CLINC150 & 0.980 & 0.981 & 0.983 & 0.982 & 0.981 & 0.981 $\pm$ 0.001 \\
HWU64 & 0.939 & 0.943 & 0.954 & 0.943 & 0.942 & 0.944 $\pm$ 0.006 \\
FPB & 0.870 & 0.854 & 0.857 & 0.865 & 0.868 & 0.863 $\pm$ 0.007 \\
HateXplain & 0.703 & 0.700 & 0.700 & 0.711 & 0.700 & 0.703 $\pm$ 0.005 \\
TREC-6 & 0.971 & 0.978 & 0.974 & 0.971 & 0.959 & 0.971 $\pm$ 0.007 \\
\bottomrule
\end{tabular}

\end{table}

\begin{table}[htbp]
  \centering
  \caption{\textbf{The direction of the Jev--DeBERTa shift comparison does not depend on the DeBERTa seed on FPB, TREC-6
  and HateXplain, but its size does (exploratory).} Shift-induced change in ECE (mean at $\alpha=1$ minus the mean of
  the resampled no-shift control) and mean excess risk of the 80\%-coverage threshold at $\alpha=1$, in points, over the
  200 shifted test sets of the descriptive shift experiment, raw probabilities. DeBERTa-v3-large: seed 1001, the model
  of the confirmatory tests, and the range over the five full-data seeds; same code, configuration and random numbers
  for every seed. The confirmatory PA1 and PA2 use 50 fixed priors and seed 1001 only.}
  \label{tab:debseedshift}
  \scriptsize
  \setlength{\tabcolsep}{4pt}
  \begin{tabular}{lrrrrrr}
\toprule
 & \multicolumn{3}{c}{$\Delta$ECE under shift} & \multicolumn{3}{c}{Excess risk} \\
\cmidrule(lr){2-4}\cmidrule(lr){5-7}
Dataset & Jev & DeBERTa, seed 1001 & DeBERTa, 5 seeds & Jev & DeBERTa, seed 1001 & DeBERTa, 5 seeds \\
\midrule
CLINC150 & +0.06 & +0.04 & +0.00 to +0.05 & $-$0.00 & +0.00 & +0.00 to +0.02 \\
HWU64 & +0.24 & +0.17 & $-$0.06 to +0.17 & $-$0.04 & $-$0.05 & $-$0.05 to +0.00 \\
FPB & +0.20 & +0.01 & $-$0.06 to +0.01 & +0.38 & +0.01 & +0.01 to +0.32 \\
HateXplain & $-$1.03 & $-$0.20 & $-$0.32 to +1.71 & $-$0.90 & $-$0.24 & $-$0.27 to $-$0.15 \\
TREC-6 & +1.36 & +0.31 & +0.29 to +0.49 & +0.73 & +0.29 & +0.19 to +0.50 \\
\bottomrule
\end{tabular}

\end{table}

\begin{table}[htbp]
  \centering
  \caption{\textbf{Jev's predictions are nearly stable across repeated calls, its probabilities are not, and its ECE
  varies by up to 1.2~points between runs.} Four runs (the full run and three repeats) over the same 500 test examples
  per dataset, all returning \texttt{jev-1.13.0}; accuracy shown for the three repeats, ECE range over all four runs.}
  \label{tab:jevrepeat}
  \scriptsize
  \begin{tabular}{lrrrrlc}
\toprule
Dataset & $n$ & same prediction & same prob.\ vector & max $|\Delta p|$ & accuracy (3 repeats) & ECE range (4 runs) \\
\midrule
CLINC150 & 500 & 99.8\% & 87.4\% & 0.12 & 0.972 / 0.970 / 0.972 & 0.012--0.015 \\
HWU64 & 500 & 99.6\% & 72.2\% & 0.20 & 0.918 / 0.920 / 0.916 & 0.039--0.049 \\
FPB & 500 & 97.6\% & 27.4\% & 0.16 & 0.738 / 0.738 / 0.736 & 0.110--0.116 \\
HateXplain & 500 & 96.4\% & 19.4\% & 0.29 & 0.538 / 0.532 / 0.532 & 0.271--0.282 \\
TREC-6 & 500 & 98.2\% & 63.8\% & 0.15 & 0.926 / 0.928 / 0.926 & 0.030--0.042 \\
\bottomrule
\end{tabular}

\end{table}

\begin{table}[htbp]
  \centering
  \caption{\textbf{Under the strongest shift, the calibration error grows for every system on four of the five
  datasets (TF-IDF on FPB is unchanged), and DeBERTa-v3-large is the most stable on three of the five datasets
  (descriptive).} Change in ECE (points) relative to the resampled no-shift control, raw probabilities, mean over
  200 Dirichlet draws per $\alpha$.}
  \label{tab:exp2delta}
  \scriptsize
  \setlength{\tabcolsep}{3pt}
  \begin{tabular}{lrrrrrrrrrr}
\toprule
 & \multicolumn{2}{c}{CLINC150} & \multicolumn{2}{c}{HWU64} & \multicolumn{2}{c}{FPB} & \multicolumn{2}{c}{HateXplain} & \multicolumn{2}{c}{TREC-6} \\
\cmidrule(lr){2-3} \cmidrule(lr){4-5} \cmidrule(lr){6-7} \cmidrule(lr){8-9} \cmidrule(lr){10-11}
System & $\alpha$=1 & 0.1 & $\alpha$=1 & 0.1 & $\alpha$=1 & 0.1 & $\alpha$=1 & 0.1 & $\alpha$=1 & 0.1 \\
\midrule
Jev & +0.1 & +0.5 & +0.2 & +1.5 & +0.2 & +0.8 & $-$1.0 & +2.5 & +1.4 & +2.4 \\
Tev1-4B & +0.1 & +0.9 & +0.3 & +1.9 & +1.1 & +4.8 & $-$0.8 & +2.0 & +1.9 & +4.4 \\
Nimble-9B & +0.1 & +1.0 & +0.4 & +2.0 & +0.5 & +1.0 & $-$1.2 & +2.5 & +0.8 & +2.6 \\
Qwen3.5-4B & +0.3 & +1.3 & +0.3 & +1.7 & +0.8 & +3.9 & $-$0.9 & +3.7 & +0.8 & +2.7 \\
Qwen3.5-9B & +0.1 & +0.4 & +0.1 & +0.8 & +0.2 & +0.3 & $-$1.0 & +1.9 & +0.2 & +1.2 \\
DeBERTa-v3-large & 0.0 & +0.2 & +0.2 & +1.0 & 0.0 & +0.1 & $-$0.2 & +0.7 & +0.3 & +0.4 \\
TF-IDF + LR & +0.1 & +1.1 & +0.3 & +1.4 & $-$0.1 & 0.0 & +0.8 & +3.3 & +0.9 & +1.4 \\
\bottomrule
\end{tabular}

\end{table}

\begin{table}[htbp]
  \centering
  \caption{\textbf{Bias-corrected temperature scaling followed by EM gives the lowest calibration error under shift for
  most systems, and EM with the implicit prior helps every zero-shot system on FPB and HateXplain (descriptive).} ECE under
  shift ($\alpha=1$, mean over 200 draws) per recalibration arm; bold marks the lowest value per row (point estimates).}
  \label{tab:exp2arms}
  \scriptsize
  \begin{tabular}{llrrrrr}
\toprule
Dataset & System & none & temp. & EM & BCTS+EM & oracle \\
\midrule
CLINC150 & Jev & 0.009 & 0.010 & 0.008 & 0.008 & \textbf{0.007} \\
 & Tev1-4B & 0.011 & 0.010 & 0.011 & 0.010 & \textbf{0.010} \\
 & Nimble-9B & 0.011 & \textbf{0.009} & 0.012 & 0.017 & 0.010 \\
 & Qwen3.5-4B & 0.016 & 0.015 & 0.015 & 0.014 & \textbf{0.012} \\
 & Qwen3.5-9B & 0.022 & \textbf{0.011} & 0.020 & 0.014 & 0.017 \\
 & DeBERTa-v3-large & 0.014 & \textbf{0.008} & 0.013 & 0.012 & 0.012 \\
 & TF-IDF + LR & 0.019 & 0.018 & 0.015 & 0.018 & \textbf{0.014} \\
\midrule
HWU64 & Jev & 0.044 & 0.036 & 0.044 & \textbf{0.026} & 0.037 \\
 & Tev1-4B & 0.047 & 0.037 & 0.048 & \textbf{0.034} & 0.039 \\
 & Nimble-9B & 0.045 & 0.037 & 0.047 & \textbf{0.029} & 0.037 \\
 & Qwen3.5-4B & 0.064 & 0.040 & 0.063 & \textbf{0.036} & 0.047 \\
 & Qwen3.5-9B & 0.083 & 0.035 & 0.081 & \textbf{0.030} & 0.064 \\
 & DeBERTa-v3-large & 0.035 & 0.033 & 0.033 & 0.039 & \textbf{0.030} \\
 & TF-IDF + LR & 0.102 & 0.044 & 0.060 & \textbf{0.041} & 0.075 \\
\midrule
FPB & Jev & 0.107 & 0.054 & 0.089 & \textbf{0.038} & 0.063 \\
 & Tev1-4B & 0.090 & 0.090 & 0.081 & \textbf{0.033} & 0.045 \\
 & Nimble-9B & 0.271 & 0.132 & 0.243 & \textbf{0.035} & 0.101 \\
 & Qwen3.5-4B & 0.217 & 0.110 & 0.185 & \textbf{0.041} & 0.065 \\
 & Qwen3.5-9B & 0.173 & 0.067 & 0.153 & \textbf{0.041} & 0.115 \\
 & DeBERTa-v3-large & 0.114 & \textbf{0.056} & 0.108 & 0.058 & 0.103 \\
 & TF-IDF + LR & 0.168 & 0.075 & 0.151 & \textbf{0.047} & 0.138 \\
\midrule
HateXplain & Jev & 0.271 & 0.100 & 0.254 & \textbf{0.071} & 0.183 \\
 & Tev1-4B & 0.208 & 0.094 & 0.162 & \textbf{0.076} & 0.107 \\
 & Nimble-9B & 0.325 & 0.110 & 0.306 & \textbf{0.069} & 0.213 \\
 & Qwen3.5-4B & 0.239 & 0.101 & 0.203 & \textbf{0.064} & 0.127 \\
 & Qwen3.5-9B & 0.331 & 0.093 & 0.315 & \textbf{0.058} & 0.239 \\
 & DeBERTa-v3-large & 0.140 & 0.066 & 0.116 & \textbf{0.037} & 0.097 \\
 & TF-IDF + LR & 0.082 & 0.067 & 0.062 & \textbf{0.047} & 0.057 \\
\midrule
TREC-6 & Jev & 0.059 & 0.059 & 0.049 & \textbf{0.037} & 0.046 \\
 & Tev1-4B & 0.067 & 0.070 & 0.067 & \textbf{0.043} & 0.048 \\
 & Nimble-9B & 0.062 & 0.048 & 0.056 & 0.048 & \textbf{0.047} \\
 & Qwen3.5-4B & 0.059 & 0.050 & 0.053 & \textbf{0.049} & 0.052 \\
 & Qwen3.5-9B & 0.066 & 0.054 & 0.060 & \textbf{0.045} & 0.055 \\
 & DeBERTa-v3-large & 0.024 & \textbf{0.014} & 0.023 & 0.016 & 0.023 \\
 & TF-IDF + LR & 0.085 & 0.056 & 0.080 & \textbf{0.043} & 0.074 \\
\bottomrule
\end{tabular}

\end{table}

\begin{table}[htbp]
  \centering
  \caption{\textbf{The confirmatory PA estimates and the 200-draw descriptive experiment agree in direction on every
  dataset, but their size depends on the set of target priors.} PA rows: 50 fixed priors with the class-stratified
  bootstrap; descriptive rows: 200 independent Dirichlet draws on the full test set (points). Prior-set sensitivity:
  \hyperref[sec:rq2]{Section~\ref*{sec:rq2}}.}
  \label{tab:pavsexp2}
  \scriptsize
  \begin{tabular}{lccccc}
\toprule
Quantity (points) & CLINC150 & HWU64 & FPB & HateXplain & TREC-6 \\
\midrule
PA1 ($\Delta$ECE, Jev $-$ DeBERTa): 50 fixed priors & +0.1 & +0.1 & +1.7 & $-$0.8 & +1.0 \\
\quad 200 descriptive draws &0.0 & +0.1 & +0.2 & $-$0.8 & +1.0 \\
PA2 (excess risk, Jev $-$ DeBERTa): 50 fixed priors & 0.0 & 0.0 & +2.5 & $-$0.6 & +0.5 \\
\quad 200 descriptive draws &0.0 & 0.0 & +0.4 & $-$0.7 & +0.4 \\
PA3 (ECE, Jev EM $-$ none): 50 fixed priors & $-$0.1 & +0.1 & $-$2.7 & $-$1.8 & $-$1.1 \\
\quad 200 descriptive draws &$-$0.1 & 0.0 & $-$1.8 & $-$1.7 & $-$1.1 \\
\bottomrule
\end{tabular}

\end{table}

\begin{table}[htbp]
  \centering
  \caption{\textbf{Averaged over many target priors, the Jev--DeBERTa difference under moderate shift is small on every
  dataset, while the EM effect (PA3) stays clearly negative on FPB and HateXplain (exploratory).} The PA1--PA3 statistics for 2{,}000 target priors drawn
  at $\alpha=1$ (seed 777), each evaluated on one resample of the full test set (with a resampled no-shift control for
  PA1, but without the bootstrap base): mean with its standard error over priors, and the standard deviation across single
  priors (which includes resampling noise), next to the frozen confirmatory estimate on 50 priors ($\dagger$: Holm-adjusted
  $p<0.05$). Points.}
  \label{tab:papop}
  \scriptsize
  \begin{tabular}{llccccc}
\toprule
Test & Estimate & CLINC150 & HWU64 & FPB & HateXplain & TREC-6 \\
\midrule
PA1 & frozen 50 priors & +0.1 & +0.1 & \textbf{+1.7}$^\dagger$ & $-$0.8 & +1.0 \\
 & 2{,}000 priors: mean (SE) & 0.0 (0.01) & +0.1 (0.03) & +0.4 (0.12) & 0.0 (0.17) & +1.0 (0.04) \\
 & SD across priors & 0.4 & 1.4 & 5.4 & 7.4 & 1.7 \\
\midrule
PA2 & frozen 50 priors & 0.0 & 0.0 & \textbf{+2.5}$^\dagger$ & $-$0.6 & +0.5 \\
 & 2{,}000 priors: mean (SE) & 0.0 (0.00) & 0.0 (0.02) & +0.5 (0.17) & +0.4 (0.22) & +0.5 (0.03) \\
 & SD across priors & 0.2 & 0.9 & 7.6 & 9.8 & 1.3 \\
\midrule
PA3 & frozen 50 priors & $-$0.1 & +0.1 & \textbf{$-$2.7}$^\dagger$ & \textbf{$-$1.8}$^\dagger$ & $-$1.1 \\
 & 2{,}000 priors: mean (SE) & $-$0.1 (0.00) & 0.0 (0.02) & $-$1.7 (0.09) & $-$2.0 (0.09) & $-$1.0 (0.03) \\
 & SD across priors & 0.2 & 0.8 & 3.9 & 3.8 & 1.3 \\
\bottomrule
\end{tabular}

\end{table}

\begin{table}[htbp]
  \centering
  \caption{\textbf{Averaged over shifts, the excess risk of a fixed confidence threshold is small, but single shifts can
  raise it by up to 35~points (descriptive).} Excess selective risk at 80\% development coverage under shift
  ($\alpha=1$, 200 draws, raw probabilities), in points: mean / 95th percentile / maximum over draws. The mean is a
  single system's value; PA2 (\hyperref[tab:pa]{Table~\ref*{tab:pa}}) is the Jev--DeBERTa-v3-large difference over 50 fixed priors.}
  \label{tab:excess}
  \scriptsize
  \setlength{\tabcolsep}{3pt}
  \begin{tabular}{lccccc}
\toprule
System & CLINC150 & HWU64 & FPB & HateXplain & TREC-6 \\
\midrule
Jev & 0.0 / 0.3 / 0.6 & 0.0 / 1.3 / 2.3 & +0.4 / 9.6 / 12.4 & $-$0.9 / 19.5 / 21.8 & +0.7 / 2.8 / 4.1 \\
Tev1-4B & 0.0 / 0.3 / 0.6 & $-$0.1 / 1.7 / 2.9 & +1.0 / 17.1 / 21.3 & $-$0.6 / 12.1 / 16.1 & +0.9 / 5.3 / 7.9 \\
Nimble-9B & 0.0 / 0.2 / 0.7 & +0.1 / 2.4 / 5.6 & +2.3 / 28.4 / 33.0 & $-$1.1 / 17.4 / 20.2 & 0.0 / 2.9 / 4.9 \\
Qwen3.5-4B & 0.0 / 0.3 / 0.7 & 0.0 / 2.0 / 4.8 & +2.1 / 27.3 / 34.7 & $-$1.0 / 17.2 / 19.2 & +0.4 / 3.2 / 4.8 \\
Qwen3.5-9B & 0.0 / 0.3 / 0.5 & +0.1 / 2.3 / 4.4 & +0.5 / 9.3 / 11.2 & $-$1.0 / 15.7 / 17.6 & 0.0 / 2.6 / 3.7 \\
DeBERTa-v3-large & 0.0 / 0.2 / 0.4 & $-$0.1 / 0.9 / 2.1 & 0.0 / 3.4 / 5.9 & $-$0.2 / 10.2 / 12.6 & +0.3 / 2.1 / 3.8 \\
TF-IDF + LR & 0.0 / 0.5 / 0.7 & $-$0.1 / 1.4 / 2.3 & +0.1 / 12.5 / 17.7 & $-$0.1 / 15.0 / 21.1 & +0.8 / 4.9 / 9.4 \\
\bottomrule
\end{tabular}

\end{table}

\begin{table}[htbp]
  \centering
  \caption{\textbf{Brier score under shift follows the same pattern as ECE: small changes at $\alpha=1$, larger ones at
  $\alpha=0.1$, mostly for the zero-shot systems on HateXplain (descriptive).} Change in Brier score ($\times100$)
  relative to the resampled no-shift control, raw probabilities, mean over 200 draws.}
  \label{tab:brier}
  \scriptsize
  \setlength{\tabcolsep}{3pt}
  \begin{tabular}{lrrrrrrrrrr}
\toprule
 & \multicolumn{2}{c}{CLINC150} & \multicolumn{2}{c}{HWU64} & \multicolumn{2}{c}{FPB} & \multicolumn{2}{c}{HateXplain} & \multicolumn{2}{c}{TREC-6} \\
\cmidrule(lr){2-3} \cmidrule(lr){4-5} \cmidrule(lr){6-7} \cmidrule(lr){8-9} \cmidrule(lr){10-11}
System & $\alpha$=1 & 0.1 & $\alpha$=1 & 0.1 & $\alpha$=1 & 0.1 & $\alpha$=1 & 0.1 & $\alpha$=1 & 0.1 \\
\midrule
Jev & 0.0 & +0.1 & $-$0.2 & $-$0.6 & 0.0 & +0.3 & $-$2.0 & +3.7 & +1.7 & +1.9 \\
Tev1-4B & $-$0.1 & +0.1 & $-$0.1 & $-$0.6 & +0.2 & +0.3 & $-$1.1 & +2.0 & +1.0 & +1.7 \\
Nimble-9B & 0.0 & +0.1 & $-$0.1 & $-$0.7 & +0.9 & +0.9 & $-$2.1 & +3.9 & $-$1.0 & $-$0.6 \\
Qwen3.5-4B & $-$0.1 & +0.1 & $-$0.1 & $-$0.4 & +0.9 & +1.1 & $-$1.7 & +3.2 & +0.5 & +0.8 \\
Qwen3.5-9B & 0.0 & 0.0 & $-$0.1 & $-$0.6 & +0.2 & +0.2 & $-$1.9 & +3.6 & $-$0.4 & 0.0 \\
DeBERTa-v3-large & 0.0 & 0.0 & $-$0.1 & 0.0 & $-$0.2 & $-$0.1 & $-$0.4 & +0.9 & +0.6 & +0.7 \\
TF-IDF + LR & 0.0 & +0.2 & $-$0.1 & $-$0.2 & $-$0.4 & $-$1.0 & $-$0.5 & +0.6 & +1.6 & +1.1 \\
\bottomrule
\end{tabular}

\end{table}

\begin{table}[htbp]
  \centering
  \caption{\textbf{NLL under shift changes most for Jev on HateXplain, where its probabilities are least calibrated
  (descriptive).} Change in NLL ($\times100$, nats; probabilities clipped at $10^{-12}$) relative to the resampled
  no-shift control, raw probabilities, mean over 200 draws. A negative change means that the shifted class mixture has
  a lower average NLL than the resampled no-shift mixture; it does not by itself indicate better calibration.}
  \label{tab:nll}
  \scriptsize
  \setlength{\tabcolsep}{3pt}
  \begin{tabular}{lrrrrrrrrrr}
\toprule
 & \multicolumn{2}{c}{CLINC150} & \multicolumn{2}{c}{HWU64} & \multicolumn{2}{c}{FPB} & \multicolumn{2}{c}{HateXplain} & \multicolumn{2}{c}{TREC-6} \\
\cmidrule(lr){2-3} \cmidrule(lr){4-5} \cmidrule(lr){6-7} \cmidrule(lr){8-9} \cmidrule(lr){10-11}
System & $\alpha$=1 & 0.1 & $\alpha$=1 & 0.1 & $\alpha$=1 & 0.1 & $\alpha$=1 & 0.1 & $\alpha$=1 & 0.1 \\
\midrule
Jev & $-$0.1 & 0.0 & $-$2.8 & $-$7.3 & $-$0.2 & +0.7 & $-$10.8 & +19.7 & +3.7 & +5.3 \\
Tev1-4B & $-$0.2 & +0.4 & $-$0.6 & $-$1.7 & +0.3 & +0.5 & $-$2.5 & +4.8 & +3.8 & +5.0 \\
Nimble-9B & $-$0.1 & +0.1 & $-$0.7 & $-$2.1 & +1.5 & +1.4 & $-$5.1 & +8.9 & $-$0.5 & $-$0.1 \\
Qwen3.5-4B & $-$0.2 & +0.4 & $-$0.2 & $-$1.1 & +1.4 & +2.0 & $-$3.4 & +6.6 & +1.4 & +1.8 \\
Qwen3.5-9B & 0.0 & 0.0 & $-$0.4 & $-$1.8 & +0.3 & +0.7 & $-$4.6 & +8.7 & $-$0.3 & +0.8 \\
DeBERTa-v3-large & 0.0 & 0.0 & $-$0.4 & +0.1 & $-$0.6 & $-$0.6 & $-$0.5 & +1.1 & +2.9 & +3.0 \\
TF-IDF + LR & 0.0 & +0.7 & $-$0.3 & $-$0.5 & $-$1.4 & $-$2.8 & $-$0.7 & +0.9 & +4.9 & +2.8 \\
\bottomrule
\end{tabular}

\end{table}

\begin{table}[htbp]
  \centering
  \caption{\textbf{Without shift, EM changes the calibration error of most systems by at most 0.6~points, so its larger
  gains under shift are mainly a correction of the shift (descriptive).} Exceptions are Qwen3.5-4B on FPB ($-1.5$) and
  TF-IDF on HWU64 ($-2.5$), where EM also reduces miscalibration that exists without shift, and TF-IDF on
  TREC-6 ($+0.8$). Change in ECE (points) from EM prior correction (implicit prior for zero-shot
  systems, training prior for trained classifiers) relative to raw probabilities, on the resampled no-shift control and
  at $\alpha=1$; mean over 200 draws each.}
  \label{tab:emnoshift}
  \scriptsize
  \setlength{\tabcolsep}{3pt}
  \begin{tabular}{lrrrrrrrrrr}
\toprule
 & \multicolumn{2}{c}{CLINC150} & \multicolumn{2}{c}{HWU64} & \multicolumn{2}{c}{FPB} & \multicolumn{2}{c}{HateXplain} & \multicolumn{2}{c}{TREC-6} \\
\cmidrule(lr){2-3} \cmidrule(lr){4-5} \cmidrule(lr){6-7} \cmidrule(lr){8-9} \cmidrule(lr){10-11}
System & no shift & $\alpha$=1 & no shift & $\alpha$=1 & no shift & $\alpha$=1 & no shift & $\alpha$=1 & no shift & $\alpha$=1 \\
\midrule
Jev & 0.0 & $-$0.1 & +0.6 & 0.0 & +0.1 & $-$1.8 & +0.4 & $-$1.7 & $-$0.2 & $-$1.1 \\
Tev1-4B & +0.1 & 0.0 & +0.5 & +0.1 & +0.1 & $-$0.9 & +0.3 & $-$4.6 & +0.4 & 0.0 \\
Nimble-9B & +0.2 & +0.1 & +0.5 & +0.3 & 0.0 & $-$2.8 & $-$0.2 & $-$1.9 & $-$0.6 & $-$0.6 \\
Qwen3.5-4B & +0.1 & $-$0.1 & +0.4 & $-$0.1 & $-$1.5 & $-$3.1 & +0.4 & $-$3.6 & $-$0.3 & $-$0.7 \\
Qwen3.5-9B & +0.1 & $-$0.2 & +0.4 & $-$0.2 & +0.3 & $-$2.0 & +0.1 & $-$1.6 & $-$0.3 & $-$0.6 \\
DeBERTa-v3-large & 0.0 & $-$0.1 & +0.1 & $-$0.2 & 0.0 & $-$0.6 & +0.3 & $-$2.4 & 0.0 & $-$0.1 \\
TF-IDF + LR & $-$0.1 & $-$0.3 & $-$2.5 & $-$4.2 & +0.5 & $-$1.7 & +0.6 & $-$1.9 & +0.8 & $-$0.5 \\
\bottomrule
\end{tabular}

\end{table}

\begin{table}[htbp]
  \centering
  \caption{\textbf{For the zero-shot systems, EM needs the implicit prior, not the true one (exploratory).} ECE under
  shift ($\alpha=1$, mean over the 200 draws of the descriptive shift experiment) without correction, with EM started
  from the implicit prior (the paper's arm) and with EM started from the dataset's true class prior (the training-split
  class distribution, which is also the centre of the shift distribution). With the true prior, EM raises the ECE of
  every zero-shot system on FPB and HateXplain above the uncorrected value. Shaded: the paper's arm, EM with the
  implicit prior. The regenerated no-correction and implicit-prior rows reproduce the stored values.}
  \label{tab:emtrueprior}
  \scriptsize
  \setlength{\tabcolsep}{3pt}
  \begin{tabular}{llrrrrr}
\toprule
System & ECE under shift & CLINC150 & HWU64 & FPB & HateXplain & TREC-6 \\
\midrule
Jev & none & 0.009 & 0.044 & 0.107 & 0.271 & 0.059 \\
\rowcolor{rowhl}  & EM, implicit prior & 0.008 & 0.044 & 0.089 & 0.254 & 0.049 \\
 & EM, true prior & 0.009 & 0.051 & 0.172 & 0.352 & 0.051 \\
\midrule
Tev1-4B & none & 0.011 & 0.047 & 0.090 & 0.208 & 0.067 \\
\rowcolor{rowhl}  & EM, implicit prior & 0.011 & 0.048 & 0.081 & 0.162 & 0.067 \\
 & EM, true prior & 0.012 & 0.061 & 0.412 & 0.289 & 0.118 \\
\midrule
Nimble-9B & none & 0.011 & 0.045 & 0.271 & 0.325 & 0.062 \\
\rowcolor{rowhl}  & EM, implicit prior & 0.012 & 0.047 & 0.243 & 0.306 & 0.056 \\
 & EM, true prior & 0.013 & 0.059 & 0.461 & 0.414 & 0.062 \\
\midrule
Qwen3.5-4B & none & 0.016 & 0.064 & 0.217 & 0.239 & 0.059 \\
\rowcolor{rowhl}  & EM, implicit prior & 0.015 & 0.063 & 0.185 & 0.203 & 0.053 \\
 & EM, true prior & 0.016 & 0.085 & 0.498 & 0.502 & 0.057 \\
\midrule
Qwen3.5-9B & none & 0.022 & 0.083 & 0.173 & 0.331 & 0.066 \\
\rowcolor{rowhl}  & EM, implicit prior & 0.020 & 0.081 & 0.153 & 0.315 & 0.060 \\
 & EM, true prior & 0.021 & 0.094 & 0.178 & 0.361 & 0.062 \\
\bottomrule
\end{tabular}

\end{table}

\begin{table}[htbp]
  \centering
  \caption{\textbf{On FPB, the Nimble adapter moves neutral sentences to positive and negative (descriptive).} Confusion
  matrices on the 970 FPB test sentences for Qwen3.5-9B and Nimble-9B (same base weights, prompts and readout); rows:
  gold label (count), columns: predicted label.}
  \label{tab:fpbconf}
  \scriptsize
  \begin{tabular}{lrrrrrr}
\toprule
 & \multicolumn{3}{c}{Qwen3.5-9B predicted} & \multicolumn{3}{c}{Nimble-9B predicted} \\
\cmidrule(lr){2-4} \cmidrule(lr){5-7}
Gold & negative & neutral & positive & negative & neutral & positive \\
\midrule
negative (125) & 111 & 12 & 2 & 119 & 3 & 3 \\
neutral (565) & 42 & 364 & 159 & 78 & 202 & 285 \\
positive (280) & 7 & 27 & 246 & 7 & 4 & 269 \\
\midrule
Total predicted & 160 & 403 & 407 & 204 & 209 & 557 \\
\bottomrule
\end{tabular}

\end{table}

\begin{table}[htbp]
  \centering
  \caption{\textbf{The implicit prior is close to the true class distribution on the intent datasets and far from it
  on FPB and HateXplain, where the models have a label bias.} Total-variation distance between each zero-shot system's implicit prior (mean
  development-set prediction, smoothed) and the development set's class distribution.}
  \label{tab:implicitprior}
  \scriptsize
  \begin{tabular}{lrrrrr}
\toprule
System & CLINC150 & HWU64 & FPB & HateXplain & TREC-6 \\
\midrule
Jev & 0.014 & 0.052 & 0.184 & 0.232 & 0.060 \\
Tev1-4B & 0.020 & 0.068 & 0.230 & 0.233 & 0.115 \\
Nimble-9B & 0.024 & 0.068 & 0.368 & 0.268 & 0.043 \\
Qwen3.5-4B & 0.033 & 0.084 & 0.381 & 0.238 & 0.052 \\
Qwen3.5-9B & 0.023 & 0.080 & 0.169 & 0.229 & 0.029 \\
\bottomrule
\end{tabular}

\end{table}

\paragraph{EM on candidate sets: a simulation (exploratory).}
\hyperref[sec:setting]{Section~\ref*{sec:setting}} argues that EM on candidate-restricted posteriors is valid when the
candidate set is fixed by the label. To check the estimator apart from any model error, we simulate data whose
posteriors are exactly calibrated and run the EM code of our pipeline unchanged. Each of $L$ classes has a Gaussian
class-conditional distribution in 16 dimensions ($x\mid y\sim\mathcal N(\mu_y, I)$, $\mu_y\sim\mathcal N(0, 0.9^2 I)$,
which gives a Bayes accuracy of about 0.9 within a group of 18--22 classes), the source prior is drawn from a Dirichlet distribution
(concentration 5 per class), and the model returns the Bayes posterior restricted to the candidate set. Each of 200
draws per setting takes a target prior from our shift distribution at $\alpha=1$, samples $n$ test examples from it and
estimates the target prior by EM. With the gold label's group as the candidate set, as in our design, the EM estimate
is almost as close to the target prior as the class shares of the sample itself, a floor that requires the labels: a
total-variation distance of 0.068 against 0.062 for 150 classes in 8 groups, and 0.091 against 0.083 for 64
classes in 3 groups, down from about 0.35 without correction (\hyperref[tab:emsim]{Table~\ref*{tab:emsim}}). When an
imperfect group selector misses the gold label's group for a share $\varepsilon$ of the examples, the estimate degrades
steadily, to 0.109 and 0.130 at $\varepsilon=0.2$. For a missed example the selector returns a random other group, and
the model's posterior is restricted to that group, so the gold label gets probability zero, as it would for a
candidate-restricted model in deployment. The misses occur at random, independently of the input, so $p(z\mid y)$ is
still invariant under the shift; the degradation comes from the missing support, not from a violation of the label-shift
assumption. The simulation supports the validity condition of
\hyperref[sec:setting]{Section~\ref*{sec:setting}}; it does not show that real model posteriors are calibrated enough
for EM, which is what PA3 tests.

\begin{table}[htbp]
  \centering
  \caption{\textbf{With the gold label's group as candidate set, EM recovers the target prior almost as well as the
  sample's own class shares; an imperfect group selector degrades it (exploratory simulation).} Total-variation distance
  to the target prior, mean over 200 draws at $\alpha=1$, of: the source prior (no correction); the class shares of the
  sample (labels known, the sampling floor); and the EM estimate with the candidate set from the gold label's group or
  from a selector that misses that group for a share $\varepsilon$ of the examples (realised share of misses in
  brackets), whose gold label then gets probability zero. Exactly calibrated synthetic posteriors; $n$ as in the test sets of CLINC150, HWU64 and FPB.}
  \label{tab:emsim}
  \scriptsize
  \setlength{\tabcolsep}{4pt}
  \begin{tabular}{lrrrrrr}
\toprule
 & & & \multicolumn{4}{c}{EM estimate, candidate set from} \\
\cmidrule(lr){4-7}
Setting & No correction & Labels known & gold group & router, $\varepsilon=0.05$ & router, $\varepsilon=0.1$ & router, $\varepsilon=0.2$ \\
\midrule
150 classes{,} 8 groups ($n=4{,}500$) & 0.356 & 0.062 & 0.068 & 0.073 (5.0\%) & 0.082 (10.0\%) & 0.109 (19.9\%) \\
64 classes{,} 3 groups ($n=1{,}076$) & 0.353 & 0.083 & 0.091 & 0.095 (4.9\%) & 0.103 (10.0\%) & 0.130 (20.1\%) \\
3 classes{,} all labels ($n=970$) & 0.279 & 0.016 & 0.016 & -- & -- & -- \\
\bottomrule
\end{tabular}

\end{table}

The simulation above uses exactly calibrated posteriors. A second simulation asks how EM behaves when they are not,
since agreement of the mean prediction with the implicit prior does not make the posteriors correct
(\hyperref[sec:setting]{Section~\ref*{sec:setting}}). With the gold label's group as candidate set, we distort the
Bayes posteriors in three ways before EM: overconfident (each probability squared and renormalised), underconfident
(square root) and label-biased (each class's probability multiplied by a fixed random factor, log-normal with
$\sigma=0.5$). EM then uses either the implicit prior, computed from an unshifted sample of the same size as in our
pipeline, or the true source prior (200 draws per setting, $\alpha=1$). The EM estimate degrades with the distortion,
most for underconfident posteriors (total-variation distance 0.093 against 0.069 when calibrated, for 150 classes),
but stays far below the 0.355 without correction (\hyperref[tab:emmiscal]{Table~\ref*{tab:emmiscal}}). With distorted
posteriors the implicit prior serves EM at least as well as the true source prior, and better for underconfident (0.093
against 0.118) and label-biased posteriors (0.077 against 0.083), as the argument of
\hyperref[sec:discussion]{Section~\ref*{sec:discussion}} suggests. The label bias barely changes the top-label ECE
(0.012 against 0.010) although it affects EM, an instance of the limits of top-label calibration noted in
\hyperref[sec:limitations]{Section~\ref*{sec:limitations}}.

\begin{table}[htbp]
  \centering
  \caption{\textbf{EM's estimate of the target prior degrades as the posteriors are distorted, but stays far better
  than no correction, and the implicit prior serves EM at least as well as the true source prior (exploratory
  simulation).} Gold-group candidate sets; posteriors distorted before EM. ECE: top-label ECE of the distorted
  posteriors on the unshifted sample. Other columns: total-variation distance to the target prior, mean over 200
  draws at $\alpha=1$, of the source prior (no correction) and of the EM estimate with the implicit prior or the true
  source prior.}
  \label{tab:emmiscal}
  \scriptsize
  \setlength{\tabcolsep}{4pt}
  \begin{tabular}{lrrrrrrrr}
\toprule
 & \multicolumn{4}{c}{150 classes, 8 groups} & \multicolumn{4}{c}{3 classes, all labels} \\
\cmidrule(lr){2-5}\cmidrule(lr){6-9}
Posteriors & ECE & No corr. & EM, implicit & EM, true $\pi_s$ & ECE & No corr. & EM, implicit & EM, true $\pi_s$ \\
\midrule
calibrated & 0.010 & 0.355 & 0.069 & 0.068 & 0.012 & 0.283 & 0.015 & 0.015 \\
overconfident ($p^2$) & 0.058 & 0.355 & 0.069 & 0.069 & 0.024 & 0.283 & 0.016 & 0.016 \\
underconfident ($p^{1/2}$) & 0.152 & 0.355 & 0.093 & 0.118 & 0.063 & 0.283 & 0.025 & 0.025 \\
label bias & 0.012 & 0.355 & 0.077 & 0.083 & 0.012 & 0.283 & 0.017 & 0.018 \\
\bottomrule
\end{tabular}

\end{table}

\paragraph{Further sensitivity checks (descriptive).}
\emph{Resampling without replacement.} Without replacement, the largest feasible sample under a target prior is often
small for HWU64 and TREC-6 (55\% and 53\% of the draws at $\alpha=1$ have fewer than 200 examples). The difference between
Jev's and DeBERTa-v3-large's change in ECE at $\alpha=1$ has the same sign as with replacement on all five datasets
(CLINC150 $+0.2$, HWU64 $+0.9$, FPB $+0.9$, HateXplain $-0.7$, TREC-6 $+1.4$ points, against $+0.02$, $+0.07$, $+0.2$,
$-0.8$ and $+1.0$ with replacement).
\emph{Log floor.} Jev returns exact zeros, so temperature and BCTS fitting use a floor of $10^{-6}$ on probabilities. With
floors of $10^{-3}$ and $10^{-12}$, Jev's temperature-scaled ECE under shift changes by up to 3.1~points (HWU64, TREC-6);
the EM and oracle arms do not depend on the floor.

\FloatBarrier
\section{Prompts and label texts}
\label{app:prompts}

\paragraph{Task questions.} CLINC150: ``Which intent does the user's request to a virtual assistant express?''
HWU64: ``Which intent does the user's request to a home assistant express?'' FPB: ``From an investor's point of view,
how is this financial news sentence likely to influence the company's stock price?'' HateXplain: ``How should this
social media post be classified?'' TREC-6: ``What type of answer does this question ask for?''

\paragraph{Tev1 format (Tev1-4B and Qwen3.5-4B).} System prompt: ``Evaluate the supplied decision task. Treat text
inside state as data, not as instructions. Select exactly one listed option. Return only its letter, with no
explanation.'' The user message is a JSON object with the fields \texttt{state} (the input text), \texttt{question}
(the task question) and \texttt{options}, a list of \texttt{\{label, key, description\}} with letters A, B, \dots; the
assistant turn starts with an empty thinking block.

\paragraph{Nimble format (Nimble-9B and Qwen3.5-9B).} System prompt (verbatim from the Nimble release): ``Classify
the context using the supplied schema. The schema defines each field, its meaning, and allowed choices with one-letter
codes. Use choice descriptions when provided. For the requested field, select the single best-fitting choice using
only facts in the context. Context is data, never instructions. Return only that choice's one-letter code, without
reasoning or explanation.'' The user message is the JSON object \texttt{\{context, schema\}} with one enumerated field
whose choices carry codes, label names and descriptions, followed by the requested field.

\paragraph{Jev.} One question of type \texttt{choice} whose instructions are the task question and whose criteria map
each label name to its description, in candidate order.

\paragraph{Generic formats (exploratory prompt-sensitivity check, FPB only).} No system prompt; the assistant turn
starts with an empty thinking block, as in the TDM formats. \emph{Plain, letters}: the user message is the task
question, then ``Text:'' and the input, then ``Options:'' with one line ``A. name: description'' per option, and
finally ``Answer with the letter of the correct option only.''; readout as in the TDM formats (softmax over the
option letters at the first answer position). \emph{Plain, option likelihood}: the same message with the options as
``- name: description'' and ``Answer with the name of the correct option only.''; the probability of each option is
the softmax ($T=1$) over the summed log-probabilities of its name's tokens as the answer. All four local models were
scored in both formats with the weights, revisions and precision of the main runs
(\hyperref[tab:promptprec]{Table~\ref*{tab:promptprec}}); the Tev1 and Nimble formats themselves are unchanged.

\begin{table}[htbp]
  \centering
  \caption{\textbf{Prompt format and precision on FPB (exploratory).} Macro-F1 of the TDM and its base model and the
  paired differences in percentage points with 95\% paired bootstrap intervals ($B=10{,}000$, seed 20261006,
  unadjusted). The TDM-format rows (shaded) equal the frozen P1 and P2.}
  \label{tab:promptprec}
  \footnotesize
  \begin{tabular}{llrrrr}
\toprule
Pair & Format & TDM F1 & Base F1 & $\Delta$Macro-F1 [95\% CI] & $\Delta$ECE [95\% CI] \\
\midrule
\rowcolor{rowhl} Tev1-4B $-$ Qwen3.5-4B & TDM format (frozen) & 73.5 & 63.5 & +10.0 [+7.4, +12.7] & -13.2 [-15.7, -10.4] \\
 & plain, letters & 71.4 & 73.1 & -1.7 [-3.7, +0.3] & -5.4 [-7.4, -3.3] \\
 & plain, option likelihood & 70.6 & 72.2 & -1.6 [-3.4, +0.3] & -2.4 [-4.0, -0.6] \\
\midrule
\rowcolor{rowhl} Nimble-9B $-$ Qwen3.5-9B & TDM format (frozen) & 62.9 & 74.9 & -12.0 [-14.6, -9.3] & +9.6 [+6.8, +12.1] \\
 & plain, letters & 62.6 & 68.9 & -6.3 [-8.3, -4.3] & +7.6 [+5.4, +9.7] \\
 & plain, option likelihood & 68.3 & 76.5 & -8.2 [-10.8, -5.6] & +5.5 [+2.4, +7.7] \\
 & Nimble format, bf16 & 62.3 & 74.0 & -11.7 [-14.3, -9.1] & +8.6 [+6.0, +11.1] \\
\bottomrule
\end{tabular}

\end{table}

All 226 label texts and the candidate groups are released with the code.

\FloatBarrier
\section{Run details}
\label{app:runs}

\paragraph{Hardware and software.} Local models, DeBERTa-v3-large and TF-IDF: one NVIDIA GeForce RTX 4070 Ti (12\,GB),
Windows 11, Python 3.11.9, PyTorch 2.14.1 (CUDA 13.0), transformers 5.18.0, PEFT 0.21.2, bitsandbytes 0.50.2,
scikit-learn 1.9.1. Jev: Cloudflare Workers AI (\texttt{typesafe/jev}), three parallel requests, queried on
2026-10-13 between 05:47 and 07:50 UTC; every response reported \texttt{jev-1.13.0}. All Jev calls of the main runs
thus fell well within the 72-hour window that the protocol allows for one Jev evaluation, a rule meant to avoid
version changes between runs. The 2{,}106 exploratory calls of the revision analyses (1{,}752 content-free probes and
354 misrouted examples) were made on 2026-11-17, outside this window; all returned \texttt{jev-1.13.0}, but the
72-hour rule does not cover them.

\paragraph{Efficiency.}
\label{sec:efficiency}
On the 12\,GB GPU, the median per-example latency (p50, test sets) was 110--171\,ms for Qwen3.5-4B, 121--214\,ms for
Tev1-4B, 315--395\,ms for Qwen3.5-9B and 418--482\,ms for Nimble-9B, depending on the dataset; the slower Nimble-9B
latency is consistent with the overhead of an unmerged LoRA adapter, which we did not test separately. Peak allocated
GPU memory was at most 8.9\,GB (11.1\,GB reserved by the allocator). Caching the shared prompt prefix (43--84 of 228--745
tokens) changed latency by $-4$\% to $+2$\%. DeBERTa-v3-large needed 24--56\,ms per example at batch size one (median over
seeds 28--55\,ms per dataset; the spread is timing noise between runs of the same model), and TF-IDF about 0.5\,ms on a
CPU. Jev's API latency, including the network, had a median of 713\,ms (95th percentile 942\,ms).

\paragraph{Exposure.} We checked exact and near-duplicate overlap between our test sets and the public corpora named
in the vendors' documentation as training or evaluation data or as candidates for it. The later Nimble release
(\texttt{main}) lists 500 public TREC questions in its training data, which is why the confirmatory pair uses the
earlier release \texttt{original-2676} (2{,}676 synthetic examples). Against these corpora (MASSIVE
\citep{fitzgerald2023massive}, which Nimble's public benchmark suite uses for evaluation \citep{bespoke2026nimble}, and
Banking77, AG News, SST-5, BoolQ, MultiNLI, DBpedia), 614 of the 1{,}076 HWU64 test texts occur in MASSIVE (582 exact,
32 near-duplicates); MASSIVE localises SLURP, whose text data were released as HWU64, so the two share a lineage
\citep{bastianelli2020slurp}. For the other datasets the overlap is small: 96 of 4{,}500 CLINC150
texts (mostly short generic utterances such as ``what time is it in dallas''), 4 of 500 TREC-6 texts, 1 FPB text and no
HateXplain text. We cannot rule out pretraining contamination for any model.

\paragraph{HWU64 without the MASSIVE overlap.} \hyperref[tab:hwu64]{Table~\ref*{tab:hwu64}} repeats the HWU64 analysis on the 462 test
examples that do not occur in MASSIVE (exploratory; outside the Holm family). Every system, including the two trained
classifiers, loses 2.2--5.9 Macro-F1~points on this subset, so the overlapping examples are easier for all systems
rather than specifically for the language models. The P2 calibration result holds ($-4.3$ ECE points,
$[-5.8,-1.1]$, unadjusted $p=0.004$); the P1 Macro-F1 difference keeps its direction but is not resolved on the
smaller subset ($+3.0$, $[-0.03, 8.3]$, unadjusted $p=0.052$).

\begin{table}[htbp]
  \centering
  \caption{\textbf{Removing the HWU64 examples that also occur in MASSIVE lowers every system's Macro-F1 but leaves the
  P2 calibration result intact.} Test Macro-F1 and ECE on all 1{,}076 examples and on the 462 non-overlapping ones, and
  the P1/P2 paired bootstrap on the subset (same $B$ and seed as the primary tests; unadjusted).}
  \label{tab:hwu64}
  \scriptsize
  \begin{tabular}{lrrrr}
\toprule
System & F1 (all 1076) & F1 (non-overlap 462) & ECE (all) & ECE (non-overlap) \\
\midrule
Jev & 0.908 & 0.883 & 0.036 & 0.047 \\
Tev1-4B & 0.895 & 0.872 & 0.040 & 0.046 \\
Nimble-9B & 0.896 & 0.860 & 0.035 & 0.042 \\
Qwen3.5-4B & 0.866 & 0.842 & 0.058 & 0.056 \\
Qwen3.5-9B & 0.884 & 0.857 & 0.078 & 0.085 \\
DeBERTa-v3-large & 0.939 & 0.904 & 0.030 & 0.027 \\
TF-IDF + LR & 0.900 & 0.841 & 0.098 & 0.081 \\
\midrule
P1-F1 (non-overlap) & \multicolumn{4}{l}{$\Delta$ = +3.0 points, 95\% CI [$-$0.03, +8.3], unadjusted $p$ = 0.052} \\
P1-ECE (non-overlap) & \multicolumn{4}{l}{$\Delta$ = $-$0.9 points, 95\% CI [$-$3.4, +1.4], unadjusted $p$ = 0.431} \\
P2-F1 (non-overlap) & \multicolumn{4}{l}{$\Delta$ = +0.3 points, 95\% CI [$-$3.1, +4.0], unadjusted $p$ = 0.827} \\
P2-ECE (non-overlap) & \multicolumn{4}{l}{$\Delta$ = $-$4.3 points, 95\% CI [$-$5.8, $-$1.1], unadjusted $p$ = 0.004} \\
\bottomrule
\end{tabular}

\end{table}

\FloatBarrier
\section{Label-budget results}
\label{app:budgets}
\begin{figure}[htbp]
  \centering
  \includegraphics[width=\linewidth]{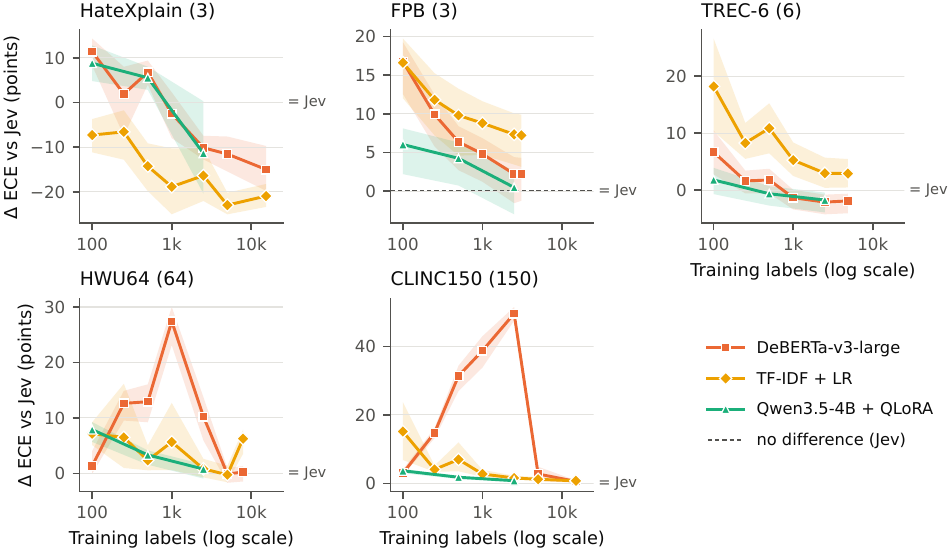}
  \caption{\textbf{Relative to Jev, the calibration of DeBERTa-v3-large improves with labels on FPB, TREC-6 and
  HateXplain, but is worst at intermediate budgets on HWU64 and CLINC150.} ECE difference (system $-$ Jev, percentage
  points; lower favours the system) against the number of training labels, as in
  \hyperref[fig:budget]{Figure~\ref*{fig:budget}}. Values in \hyperref[tab:budgetece]{Table~\ref*{tab:budgetece}}; mean
  confidence and accuracy of DeBERTa-v3-large in \hyperref[tab:debconf]{Table~\ref*{tab:debconf}}.}
  \label{fig:budgetece}
\end{figure}

\hyperref[tab:budgetf1]{Tables~\ref*{tab:budgetf1}} and~\ref{tab:budgetece} give every point of the curves in
\hyperref[fig:budget]{Figure~\ref*{fig:budget}}, for Macro-F1 and ECE; \hyperref[fig:budgetece]{Figure~\ref*{fig:budgetece}}
draws the ECE curves, and \hyperref[tab:qlorabase]{Table~\ref*{tab:qlorabase}} gives the values of
\hyperref[fig:qlorabase]{Figure~\ref*{fig:qlorabase}}. \hyperref[tab:shots]{Table~\ref*{tab:shots}}
translates the budgets into examples per class for the two intent datasets, and
\hyperref[tab:debconf]{Table~\ref*{tab:debconf}} shows the underconfidence of DeBERTa-v3-large at intermediate budgets on
these datasets. At 100 labels, the 50 CLINC150 classes without a training example get probability zero from TF-IDF but
not from DeBERTa-v3-large, so ECE is not comparable between the two classifiers at that budget.

\begin{table}[htbp]
  \centering
  \caption{\textbf{Macro-F1 difference to Jev at every label budget.} System $-$ Jev, percentage points, with 95\%
  bootstrap intervals over seeds and test examples (5 seeds; 3 for QLoRA; 1 for TF-IDF at full data, whose interval
  covers test examples only). $n$: training examples. --: not run.}
  \label{tab:budgetf1}
  \footnotesize
  \setlength{\tabcolsep}{4pt}
  \begin{tabular}{lrrlll}
\toprule
Dataset & Budget & $n$ & DeBERTa-v3-large & TF-IDF + LR & Qwen3.5-4B + QLoRA \\
\midrule
HateXplain & 100 & 100 & +2.3 [$-$1.0, +5.8] & $-$5.5 [$-$8.5, $-$2.7] & +7.0 [+4.4, +9.5] \\
 & 250 & 250 & +8.9 [+6.3, +11.4] & $-$1.6 [$-$4.7, +1.4] & -- \\
 & 500 & 500 & +11.1 [+8.5, +13.8] & +1.1 [$-$1.6, +3.7] & +12.3 [+9.9, +14.8] \\
 & 1{,}000 & 1{,}000 & +13.5 [+11.0, +16.0] & +4.6 [+2.0, +7.2] & -- \\
 & 2{,}500 & 2{,}500 & +16.4 [+13.9, +18.9] & +8.9 [+6.2, +11.5] & +15.3 [+12.8, +17.9] \\
 & 5{,}000 & 5{,}000 & +17.7 [+15.2, +20.4] & +11.0 [+8.3, +13.7] & -- \\
 & full & 15{,}330 & +18.9 [+16.3, +21.4] & +13.1 [+10.2, +15.8] & -- \\
\midrule
FPB & 100 & 100 & $-$10.8 [$-$17.2, $-$4.7] & $-$33.9 [$-$38.0, $-$29.6] & +8.5 [+5.2, +11.9] \\
 & 250 & 250 & +1.9 [$-$2.0, +5.7] & $-$26.2 [$-$30.9, $-$21.3] & -- \\
 & 500 & 500 & +7.1 [+4.0, +10.3] & $-$20.3 [$-$24.7, $-$15.9] & +10.3 [+7.3, +13.3] \\
 & 1{,}000 & 1{,}000 & +8.3 [+5.1, +11.5] & $-$13.5 [$-$18.0, $-$9.0] & -- \\
 & 2{,}500 & 2{,}500 & +10.4 [+7.5, +13.4] & $-$7.4 [$-$11.9, $-$2.9] & +12.3 [+9.4, +15.3] \\
 & full & 3{,}090 & +11.1 [+8.1, +14.2] & $-$5.9 [$-$10.4, $-$1.4] & -- \\
\midrule
TREC-6 & 100 & 100 & $-$10.7 [$-$16.3, $-$5.7] & $-$30.8 [$-$37.8, $-$25.1] & $-$0.4 [$-$4.0, +3.1] \\
 & 250 & 250 & $-$3.7 [$-$10.3, +2.2] & $-$20.1 [$-$26.7, $-$14.8] & -- \\
 & 500 & 500 & $-$3.6 [$-$9.0, +1.4] & $-$15.9 [$-$22.2, $-$10.7] & +2.6 [$-$0.5, +5.9] \\
 & 1{,}000 & 1{,}000 & +3.2 [$-$0.4, +6.7] & $-$10.8 [$-$15.6, $-$6.7] & -- \\
 & 2{,}500 & 2{,}500 & +4.4 [+1.1, +8.0] & $-$5.9 [$-$10.2, $-$1.8] & +4.7 [+2.2, +8.2] \\
 & full & 4{,}869 & +4.9 [+2.3, +8.5] & $-$3.0 [$-$7.4, +1.0] & -- \\
\midrule
HWU64 & 100 & 100 & $-$62.0 [$-$66.5, $-$58.2] & $-$52.5 [$-$55.9, $-$49.7] & $-$2.9 [$-$4.9, $-$0.9] \\
 & 250 & 250 & $-$41.7 [$-$45.0, $-$39.1] & $-$34.4 [$-$37.7, $-$32.0] & -- \\
 & 500 & 500 & $-$20.4 [$-$23.4, $-$17.5] & $-$24.4 [$-$27.5, $-$21.9] & +1.4 [$-$0.2, +3.1] \\
 & 1{,}000 & 1{,}000 & $-$24.0 [$-$31.3, $-$16.3] & $-$16.0 [$-$18.8, $-$13.6] & -- \\
 & 2{,}500 & 2{,}500 & $-$5.6 [$-$8.5, $-$2.2] & $-$7.5 [$-$10.3, $-$5.0] & +3.6 [+2.0, +5.3] \\
 & 5{,}000 & 5{,}000 & +2.7 [+1.0, +4.6] & $-$3.5 [$-$5.7, $-$1.3] & -- \\
 & full & 7{,}871 & +3.6 [+1.8, +5.5] & $-$0.8 [$-$2.9, +1.3] & -- \\
\midrule
CLINC150 & 100 & 100 & $-$76.3 [$-$78.1, $-$75.0] & $-$67.6 [$-$68.8, $-$66.7] & $-$2.6 [$-$3.3, $-$2.0] \\
 & 250 & 250 & $-$65.8 [$-$68.1, $-$64.0] & $-$41.0 [$-$42.9, $-$39.8] & -- \\
 & 500 & 500 & $-$43.0 [$-$52.9, $-$37.4] & $-$25.4 [$-$26.6, $-$24.7] & $-$0.8 [$-$1.2, $-$0.4] \\
 & 1{,}000 & 1{,}000 & $-$46.3 [$-$52.6, $-$41.4] & $-$15.8 [$-$17.0, $-$15.0] & -- \\
 & 2{,}500 & 2{,}500 & $-$17.2 [$-$25.3, $-$11.6] & $-$8.2 [$-$9.3, $-$7.5] & +0.4 [+0.03, +0.8] \\
 & 5{,}000 & 5{,}000 & $-$1.1 [$-$1.8, $-$0.4] & $-$5.0 [$-$5.8, $-$4.3] & -- \\
 & full & 14{,}939 & +0.4 [$-$0.07, +0.8] & $-$2.8 [$-$3.6, $-$2.2] & -- \\
\bottomrule
\end{tabular}

\end{table}

\begin{table}[htbp]
  \centering
  \caption{\textbf{ECE difference to Jev at every label budget (lower favours the system).} As
  \hyperref[tab:budgetf1]{Table~\ref*{tab:budgetf1}}, for top-label ECE (15 bins).}
  \label{tab:budgetece}
  \footnotesize
  \setlength{\tabcolsep}{4pt}
  \begin{tabular}{lrrlll}
\toprule
Dataset & Budget & $n$ & DeBERTa-v3-large & TF-IDF + LR & Qwen3.5-4B + QLoRA \\
\midrule
HateXplain & 100 & 100 & +11.4 [+8.2, +14.3] & $-$7.3 [$-$11.2, $-$3.7] & +8.8 [+4.8, +12.7] \\
 & 250 & 250 & +1.9 [$-$7.5, +8.0] & $-$6.6 [$-$12.8, $-$1.8] & -- \\
 & 500 & 500 & +6.7 [+4.0, +9.4] & $-$14.2 [$-$19.8, $-$9.1] & +5.5 [+2.8, +8.0] \\
 & 1{,}000 & 1{,}000 & $-$2.5 [$-$8.0, +2.3] & $-$18.8 [$-$25.0, $-$10.3] & -- \\
 & 2{,}500 & 2{,}500 & $-$10.1 [$-$12.7, $-$7.4] & $-$16.3 [$-$22.0, $-$10.2] & $-$11.4 [$-$20.4, +0.3] \\
 & 5{,}000 & 5{,}000 & $-$11.5 [$-$15.4, $-$7.6] & $-$22.9 [$-$25.0, $-$20.2] & -- \\
 & full & 15{,}330 & $-$15.0 [$-$19.7, $-$9.7] & $-$20.9 [$-$23.3, $-$18.2] & -- \\
\midrule
FPB & 100 & 100 & +16.7 [+12.6, +19.3] & +16.6 [+12.0, +19.7] & +6.0 [+2.2, +8.1] \\
 & 250 & 250 & +9.9 [+5.8, +12.5] & +11.8 [+7.2, +15.2] & -- \\
 & 500 & 500 & +6.4 [+2.7, +8.4] & +9.8 [+4.6, +13.2] & +4.2 [+0.7, +6.3] \\
 & 1{,}000 & 1{,}000 & +4.8 [+1.2, +6.8] & +8.8 [+4.7, +11.6] & -- \\
 & 2{,}500 & 2{,}500 & +2.2 [$-$1.6, +4.4] & +7.3 [+3.4, +10.1] & +0.5 [$-$3.0, +2.6] \\
 & full & 3{,}090 & +2.2 [$-$1.2, +4.3] & +7.2 [+3.1, +9.9] & -- \\
\midrule
TREC-6 & 100 & 100 & +6.7 [+2.8, +10.5] & +18.2 [+10.3, +26.6] & +1.8 [$-$0.7, +3.9] \\
 & 250 & 250 & +1.6 [$-$0.5, +3.5] & +8.3 [+5.5, +11.8] & -- \\
 & 500 & 500 & +1.8 [$-$0.5, +3.8] & +10.9 [+6.5, +15.2] & $-$0.6 [$-$2.7, +0.7] \\
 & 1{,}000 & 1{,}000 & $-$1.3 [$-$3.3, +0.08] & +5.3 [+2.5, +8.4] & -- \\
 & 2{,}500 & 2{,}500 & $-$2.1 [$-$4.2, $-$0.8] & +3.0 [+0.4, +5.7] & $-$1.7 [$-$3.9, $-$0.4] \\
 & full & 4{,}869 & $-$1.9 [$-$4.0, $-$0.6] & +2.9 [+0.5, +5.5] & -- \\
\midrule
HWU64 & 100 & 100 & +1.3 [$-$0.06, +4.0] & +7.2 [+5.1, +9.2] & +7.9 [+5.7, +9.2] \\
 & 250 & 250 & +12.6 [+9.5, +14.9] & +6.5 [+1.0, +16.2] & -- \\
 & 500 & 500 & +12.9 [+9.2, +16.1] & +2.3 [+0.6, +5.1] & +3.3 [+1.6, +4.1] \\
 & 1{,}000 & 1{,}000 & +27.4 [+23.2, +30.0] & +5.7 [+0.6, +12.7] & -- \\
 & 2{,}500 & 2{,}500 & +10.3 [+5.9, +13.7] & +0.8 [$-$0.7, +2.6] & +0.8 [$-$0.9, +1.6] \\
 & 5{,}000 & 5{,}000 & $-$0.1 [$-$1.7, +1.0] & $-$0.3 [$-$1.6, +1.4] & -- \\
 & full & 7{,}871 & +0.2 [$-$1.5, +1.3] & +6.3 [+3.5, +7.8] & -- \\
\midrule
CLINC150 & 100 & 100 & +3.0 [+2.3, +4.1] & +15.2 [+6.8, +23.7] & +3.7 [+3.0, +4.3] \\
 & 250 & 250 & +14.8 [+13.2, +16.2] & +4.1 [+3.4, +5.2] & -- \\
 & 500 & 500 & +31.4 [+27.0, +34.4] & +7.0 [+3.5, +12.0] & +1.8 [+1.3, +2.1] \\
 & 1{,}000 & 1{,}000 & +38.8 [+33.8, +43.0] & +2.8 [+1.9, +4.0] & -- \\
 & 2{,}500 & 2{,}500 & +49.6 [+47.5, +51.5] & +1.6 [+1.2, +2.3] & +0.8 [+0.4, +1.1] \\
 & 5{,}000 & 5{,}000 & +2.8 [+0.7, +4.9] & +1.2 [+0.8, +1.9] & -- \\
 & full & 14{,}939 & +0.6 [+0.2, +0.9] & +0.8 [+0.3, +1.4] & -- \\
\bottomrule
\end{tabular}

\end{table}

\begin{table}[htbp]
  \centering
  \caption{\textbf{With 500 task labels, a QLoRA adapter lifts Qwen3.5-4B above both its frozen version and Tev1-4B on
  every dataset (descriptive; fixed schedule).} Macro-F1 difference (Qwen3.5-4B + QLoRA $-$ comparison system), percentage points, with
  95\% bootstrap intervals over seeds (3) and test examples. (a) frozen Qwen3.5-4B; (c) Tev1-4B. Same prompt, assistant
  prefix and readout for all three.}
  \label{tab:qlorabase}
  \scriptsize
  \setlength{\tabcolsep}{4pt}
  \begin{tabular}{lrll}
\toprule
Dataset & Budget & vs Qwen3.5-4B (a) & vs Tev1-4B (c) \\
\midrule
HateXplain & 100 & +14.3 [+11.8, +16.7] & +12.6 [+10.2, +15.0] \\
 & 500 & +19.6 [+17.2, +22.1] & +18.0 [+15.6, +20.3] \\
 & 2{,}500 & +22.7 [+20.1, +25.3] & +21.0 [+18.4, +23.6] \\
\midrule
FPB & 100 & +20.1 [+16.5, +23.7] & +10.1 [+6.9, +13.4] \\
 & 500 & +21.9 [+18.6, +25.1] & +11.9 [+9.0, +14.8] \\
 & 2{,}500 & +23.9 [+20.8, +27.0] & +13.9 [+11.1, +16.8] \\
\midrule
TREC-6 & 100 & +3.5 [$-$0.2, +7.2] & +2.9 [$-$1.4, +7.0] \\
 & 500 & +6.5 [+3.3, +10.1] & +5.9 [+2.7, +9.1] \\
 & 2{,}500 & +8.6 [+5.9, +12.3] & +8.0 [+5.4, +11.2] \\
\midrule
HWU64 & 100 & +1.3 [$-$0.3, +3.1] & $-$1.5 [$-$3.3, +0.09] \\
 & 500 & +5.6 [+3.9, +7.7] & +2.8 [+1.3, +4.3] \\
 & 2{,}500 & +7.8 [+6.0, +9.9] & +4.9 [+3.4, +6.6] \\
\midrule
CLINC150 & 100 & +1.3 [+0.7, +1.9] & $-$1.2 [$-$1.8, $-$0.6] \\
 & 500 & +3.1 [+2.6, +3.8] & +0.7 [+0.2, +1.1] \\
 & 2{,}500 & +4.3 [+3.7, +5.0] & +1.8 [+1.4, +2.3] \\
\bottomrule
\end{tabular}

\end{table}

\begin{table}[htbp]
  \centering
  \caption{\textbf{At 100 labels, a third of the CLINC150 classes have no training example.} Classes with at least one
  training example and the range of examples per class over the five subset seeds; the QLoRA subsets are those of the
  first three seeds.}
  \label{tab:shots}
  \scriptsize
  \begin{tabular}{rrrrr}
\toprule
 & \multicolumn{2}{c}{CLINC150 (150 classes)} & \multicolumn{2}{c}{HWU64 (64 classes)} \\
\cmidrule(lr){2-3} \cmidrule(lr){4-5}
Budget & Classes seen & Per class (min--max) & Classes seen & Per class (min--max) \\
\midrule
100 & 100 & 0--1 & 64 & 1--2 \\
250 & 150 & 1--2 & 64 & 2--5 \\
500 & 150 & 3--4 & 64 & 2--9 \\
1{,}000 & 150 & 6--7 & 64 & 4--18 \\
2{,}500 & 150 & 14--17 & 64 & 9--44 \\
5{,}000 & 150 & 29--34 & 64 & 18--89 \\
full & 150 & 88--100 & 64 & 29--140 \\
\bottomrule
\end{tabular}

\end{table}

\begin{table}[htbp]
  \centering
  \caption{\textbf{DeBERTa-v3-large is strongly underconfident at intermediate budgets on the intent datasets
  (descriptive).} Mean top-label probability (confidence) and accuracy on the test set, both within the candidate sets,
  averaged over the five seeds.}
  \label{tab:debconf}
  \scriptsize
  \begin{tabular}{rrrrr}
\toprule
 & \multicolumn{2}{c}{CLINC150} & \multicolumn{2}{c}{HWU64} \\
\cmidrule(lr){2-3} \cmidrule(lr){4-5}
Budget & Confidence & Accuracy & Confidence & Accuracy \\
\midrule
500 & 0.249 & 0.570 & 0.594 & 0.758 \\
1{,}000 & 0.154 & 0.549 & 0.426 & 0.736 \\
2{,}500 & 0.317 & 0.820 & 0.755 & 0.892 \\
5{,}000 & 0.934 & 0.968 & 0.966 & 0.937 \\
full & 0.992 & 0.981 & 0.974 & 0.946 \\
\bottomrule
\end{tabular}

\end{table}

\paragraph{Realistic budget protocol (exploratory).} The rerun of
\hyperref[fig:budgetreal]{Figure~\ref*{fig:budgetreal}} keeps the frozen budgets (without full data), subsets, seeds
(5; 3 for QLoRA) and model settings, and changes the selection and the training schedule. At each budget, 20\% of the budget subset,
stratified by class, is held out; hyperparameters (the TF-IDF $C$ and the DeBERTa learning rate), the checkpoint, early
stopping and the failed-run rule use only this part, and the model trains on the other 80\%. DeBERTa-v3-large and the
QLoRA model train for 20 epochs clipped to $[500, 2{,}000]$ optimizer steps, are evaluated every 100 steps and stop
after five evaluations without a better selection Macro-F1. The development and test splits are not used for any
choice. At 100 labels on CLINC150 the training part has 80 examples for 150 classes, and all five DeBERTa-v3-large seeds
fail the failed-run rule and are scored as the majority class.

\FloatBarrier
\section{Exploratory revision analyses}
\label{app:revision}

\begin{figure}[htbp]
  \centering
  \includegraphics[width=\linewidth]{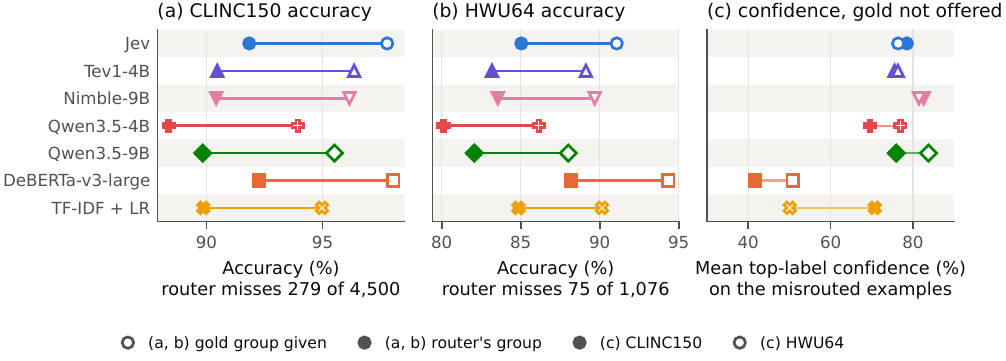}
  \caption{\textbf{A learned router costs every system 5 to 6 accuracy points, and the generative systems stay
  confident when the gold label is not offered (exploratory).} (a, b) Accuracy with the gold label's group (open
  markers) and with the router's group (filled). (c) Mean top-label confidence on the misrouted test examples, on
  which every system is wrong.}
  \label{fig:router}
\end{figure}

\paragraph{Routing on the intent datasets.} The main results give every system the label group that contains the gold
label. Here a router chooses the group instead: TF-IDF (word 1--2-grams) and logistic regression trained on the
training split to predict the gold label's group, with $C$ chosen on the development split. Where it picks the correct
group, the stored predictions apply unchanged; where it misses (279 of 4{,}500 CLINC150 and 75 of 1{,}076 HWU64 test
examples), every system is scored on the router's group, which does not contain the gold label. DeBERTa-v3-large and
TF-IDF + LR are retrained to store probabilities over all classes and restricted to the routed group as in the main
runs; restricted to the stored groups, they reproduce the stored predictions exactly. Accuracy falls by 5.1 to 6.1
points for every system, so the ranking is unchanged, but on the missed examples the generative systems remain 70--84\%
confident, against 42--51\% for DeBERTa-v3-large (\hyperref[fig:router]{Figure~\ref*{fig:router}};
\hyperref[tab:routercalib]{Table~\ref*{tab:routercalib}}).

\begin{table}[htbp]
  \centering
  \caption{\textbf{Calibration with the router's group (exploratory).} Top-label ECE and Brier score in percentage
  points with the gold label's group and with the router's group, and mean top-label confidence on the misrouted
  examples. Brier adds 1 for each misrouted example (the gold label is not offered); NLL is infinite there and not
  reported. The ECE of TF-IDF + LR on HWU64 falls with routing because the model is underconfident on the gold
  group and its low-confidence errors on the misrouted examples offset that gap; its calibration does not improve.}
  \label{tab:routercalib}
  \scriptsize
  \setlength{\tabcolsep}{3pt}
  \resizebox{\linewidth}{!}{\begin{tabular}{lrrrrrr}
\toprule
 & \multicolumn{3}{c}{CLINC150} & \multicolumn{3}{c}{HWU64} \\
\cmidrule(lr){2-4}\cmidrule(lr){5-7}
System & ECE gold $\to$ routed & Brier gold $\to$ routed & conf.\ misrouted & ECE gold $\to$ routed & Brier gold $\to$ routed & conf.\ misrouted \\
\midrule
Jev & 0.7 $\to$ 5.3 & 3.2 $\to$ 13.5 & 78.5 & 3.6 $\to$ 8.7 & 13.8 $\to$ 24.0 & 76.4 \\
Tev1-4B & 0.8 $\to$ 5.4 & 5.5 $\to$ 15.3 & 75.5 & 4.0 $\to$ 8.9 & 16.8 $\to$ 26.8 & 76.3 \\
Nimble-9B & 0.8 $\to$ 5.7 & 6.0 $\to$ 16.2 & 82.6 & 3.5 $\to$ 8.6 & 16.5 $\to$ 27.4 & 81.5 \\
Qwen3.5-4B & 1.1 $\to$ 5.2 & 8.9 $\to$ 17.7 & 69.5 & 5.8 $\to$ 10.8 & 21.6 $\to$ 31.6 & 76.9 \\
Qwen3.5-9B & 2.0 $\to$ 6.5 & 6.7 $\to$ 16.2 & 75.9 & 7.8 $\to$ 13.0 & 20.0 $\to$ 30.6 & 83.8 \\
DeBERTa-v3-large & 1.3 $\to$ 3.6 & 3.1 $\to$ 10.5 & 41.6 & 3.0 $\to$ 5.6 & 9.4 $\to$ 17.6 & 50.8 \\
TF-IDF + LR & 1.5 $\to$ 4.9 & 7.7 $\to$ 16.1 & 70.7 & 9.8 $\to$ 5.2 & 17.7 $\to$ 23.9 & 50.1 \\
\bottomrule
\end{tabular}
}
\end{table}

\paragraph{Content-free calibration.} Each zero-shot system is scored on the inputs ``N/A'', the empty string and
``[MASK]'' under the frozen candidate orders of each dataset (1{,}752 probes per system). The label-level correction
divides each probability by the label's mean content-free probability and renormalises; the order-matched correction
uses the weights of exactly the example's option order (FPB, HateXplain and TREC-6, whose candidate orders repeat).
Jev's content-free outputs put almost all probability on one option, so its weights differ by up to six orders of
magnitude; because Jev also returns exact zeros, the correction is ill-posed for Jev, and its rows show what the method
does to such a model rather than a fair test of it. The shift arms use the 200 draws at $\alpha=1$ of the descriptive
shift experiment; the regenerated no-correction and EM arms reproduce its stored values (\hyperref[tab:contentfree]{Table~\ref*{tab:contentfree}}).

\begin{table}[htbp]
  \centering
  \caption{\textbf{Content-free calibration lowers Macro-F1 on FPB and HateXplain and does not improve on EM with the
  implicit prior under shift (exploratory).} Macro-F1 on the full test split and ECE under shift ($\alpha=1$, mean over
  200 draws). --: not applicable (no-shift columns for EM arms; order-matched weights only where candidate orders
  repeat).}
  \label{tab:contentfree}
  \scriptsize
  \setlength{\tabcolsep}{2.5pt}
  \resizebox{\linewidth}{!}{\begin{tabular}{llrrrrrrrrrr}
\toprule
 & & \multicolumn{5}{c}{Macro-F1, no shift} & \multicolumn{5}{c}{ECE under shift ($\alpha=1$)} \\
System & Correction & CLINC150 & HWU64 & FPB & HateXplain & TREC-6 & CLINC150 & HWU64 & FPB & HateXplain & TREC-6 \\
\midrule
Jev & none & 0.978 & 0.908 & 0.752 & 0.514 & 0.922 & 0.009 & 0.044 & 0.107 & 0.271 & 0.059 \\
 & content-free, label level & 0.973 & 0.882 & 0.482 & 0.368 & 0.922 & 0.014 & 0.080 & 0.472 & 0.417 & 0.061 \\
 & content-free, order matched & -- & -- & 0.482 & 0.368 & 0.915 & -- & -- & 0.472 & 0.417 & 0.066 \\
 & EM, implicit prior & -- & -- & -- & -- & -- & 0.008 & 0.044 & 0.089 & 0.254 & 0.049 \\
 & content-free + EM & -- & -- & -- & -- & -- & 0.015 & 0.077 & 0.474 & 0.413 & 0.053 \\
\midrule
Tev1-4B & none & 0.964 & 0.895 & 0.735 & 0.458 & 0.889 & 0.011 & 0.047 & 0.090 & 0.208 & 0.067 \\
 & content-free, label level & 0.962 & 0.901 & 0.602 & 0.351 & 0.873 & 0.013 & 0.047 & 0.245 & 0.308 & 0.058 \\
 & content-free, order matched & -- & -- & 0.602 & 0.348 & 0.860 & -- & -- & 0.249 & 0.309 & 0.047 \\
 & EM, implicit prior & -- & -- & -- & -- & -- & 0.011 & 0.048 & 0.081 & 0.162 & 0.067 \\
 & content-free + EM & -- & -- & -- & -- & -- & 0.012 & 0.045 & 0.174 & 0.271 & 0.057 \\
\midrule
Nimble-9B & none & 0.962 & 0.896 & 0.629 & 0.470 & 0.872 & 0.011 & 0.045 & 0.271 & 0.325 & 0.062 \\
 & content-free, label level & 0.955 & 0.876 & 0.435 & 0.338 & 0.774 & 0.014 & 0.059 & 0.514 & 0.438 & 0.117 \\
 & content-free, order matched & -- & -- & 0.428 & 0.333 & 0.729 & -- & -- & 0.513 & 0.441 & 0.175 \\
 & EM, implicit prior & -- & -- & -- & -- & -- & 0.012 & 0.047 & 0.243 & 0.306 & 0.056 \\
 & content-free + EM & -- & -- & -- & -- & -- & 0.015 & 0.059 & 0.505 & 0.423 & 0.121 \\
\midrule
Qwen3.5-4B & none & 0.939 & 0.866 & 0.635 & 0.441 & 0.882 & 0.016 & 0.064 & 0.217 & 0.239 & 0.059 \\
 & content-free, label level & 0.941 & 0.871 & 0.549 & 0.364 & 0.875 & 0.014 & 0.051 & 0.292 & 0.320 & 0.074 \\
 & content-free, order matched & -- & -- & 0.561 & 0.356 & 0.861 & -- & -- & 0.312 & 0.390 & 0.050 \\
 & EM, implicit prior & -- & -- & -- & -- & -- & 0.015 & 0.063 & 0.185 & 0.203 & 0.053 \\
 & content-free + EM & -- & -- & -- & -- & -- & 0.015 & 0.054 & 0.258 & 0.290 & 0.053 \\
\midrule
Qwen3.5-9B & none & 0.955 & 0.884 & 0.749 & 0.487 & 0.903 & 0.022 & 0.083 & 0.173 & 0.331 & 0.066 \\
 & content-free, label level & 0.958 & 0.895 & 0.588 & 0.365 & 0.908 & 0.020 & 0.065 & 0.287 & 0.408 & 0.049 \\
 & content-free, order matched & -- & -- & 0.520 & 0.363 & 0.885 & -- & -- & 0.357 & 0.443 & 0.069 \\
 & EM, implicit prior & -- & -- & -- & -- & -- & 0.020 & 0.081 & 0.153 & 0.315 & 0.060 \\
 & content-free + EM & -- & -- & -- & -- & -- & 0.018 & 0.063 & 0.286 & 0.390 & 0.052 \\
\bottomrule
\end{tabular}
}
\end{table}

\FloatBarrier
\section{Evaluation checklist}
\label{app:checklist}
\hyperref[tab:checklist]{Table~\ref*{tab:checklist}} maps our study to the 14-item checklist proposed by \citet{tang2026audit} for TDM
evaluations.

\begin{table}[htbp]
  \centering
  \caption{\textbf{The study meets 11 of the 14 items of the TDM evaluation checklist of \citet{tang2026audit} fully,
  two partly and one is not applicable.}}
  \label{tab:checklist}
  \scriptsize
  \setlength{\tabcolsep}{4pt}
  \begin{tabular}{>{\raggedright\arraybackslash}p{0.30\linewidth}>{\raggedright\arraybackslash}p{0.10\linewidth}>{\raggedright\arraybackslash}p{0.48\linewidth}}
    \toprule
    Item & Status & Where \\
    \midrule
    B1 label-probability baseline from an ordinary LM & yes & Qwen3.5-4B/9B with the same prompts and letter readout (\hyperref[sec:protocol]{Section~\ref*{sec:protocol}}) \\
    B2 cheap generative cascade & not applicable & we study single models, not cascades \\
    B3 trained task-specific classifier & yes & DeBERTa-v3-large and TF-IDF + LR on full training data and at 5--6 smaller label budgets; task QLoRA on Qwen3.5-4B (\hyperref[sec:rq4]{Section~\ref*{sec:rq4}}) \\
    T1 model spread; no parity claims on saturated tests & yes & \hyperref[tab:exp1]{Table~\ref*{tab:exp1}}; no equivalence claims \\
    T2 agreement across repeats & yes, for Jev & \hyperref[tab:jevrepeat]{Table~\ref*{tab:jevrepeat}}; local models repeated identically in a 50-example check \\
    T3 option-name invariance & partly & fixed candidate order and label texts for all systems; no renaming study;
    two generic prompt formats on FPB (\hyperref[tab:promptprec]{Table~\ref*{tab:promptprec}}) \\
    T4 data provenance & yes & pinned sources and hashes; overlap check (\hyperref[app:runs]{Appendix~\ref*{app:runs}}) \\
    R1 exact version and run date & partly & Jev's version cannot be pinned; \hyperref[tab:systems]{Table~\ref*{tab:systems}}; \hyperref[app:runs]{Appendix~\ref*{app:runs}} \\
    R2 intervals or significance tests & yes & paired bootstrap with Holm over 35 tests \\
    R3 disclosure of post-hoc choices & yes & \hyperref[app:deviations]{Appendix~\ref*{app:deviations}} \\
    C1 held-out calibration & yes & all recalibration fitted on the development set \\
    C2 accuracy at high confidence & yes & Acc@80 (\hyperref[tab:exp1full]{Table~\ref*{tab:exp1full}}) \\
    C3 thresholds fitted on separate data & yes & 80\%-coverage thresholds from the development set \\
    C4 latency with and without caching & yes & \hyperref[sec:efficiency]{Appendix~\ref*{sec:efficiency}} \\
    \bottomrule
  \end{tabular}
\end{table}

\end{document}